%% file: arxiv.tex
\PassOptionsToPackage{table,dvipsnames}{xcolor}
\documentclass{article}

\usepackage[preprint,nonatbib]{neurips_2026}

\usepackage[utf8]{inputenc} 
\usepackage[T1]{fontenc}    
\usepackage{hyperref}       
\usepackage{url}            
\usepackage{booktabs}       
\usepackage{amsfonts}       
\usepackage{nicefrac}       
\usepackage[most]{tcolorbox}
\usepackage{xcolor}
\usepackage{tikz}
\usepackage{colortbl}
\usepackage{multirow}
\usepackage{graphicx}
\usepackage{amsmath,amssymb,amsthm,bbm,mathtools}
\usepackage{algorithm}
\usepackage{algorithmic}
\usepackage{wrapfig}
\usepackage{pifont}
\usepackage{makecell}
\usepackage{placeins}
\usepackage{xspace}
\usepackage{enumitem}
\usepackage{newtxtext} 
\usepackage{threeparttable}
\usepackage{arydshln}
\usepackage{hhline}
\usepackage[most]{tcolorbox}
\usepackage[table,dvipsnames]{xcolor}
\usepackage{subcaption}
\usepackage[square,numbers]{natbib}
\usepackage{contour}
\newcolumntype{g}{>{\columncolor{gray!15}}l}

\newcommand{\confoundingshap}{\textsc{ConfoundingSHAP}\xspace}

\definecolor{lightblue}{HTML}{DAE8FC}
\definecolor{darkblue}{HTML}{6C8EBF}
\definecolor{orange}{HTML}{fdae6b}
\definecolor{darkorange}{HTML}{8c2d04}

\DeclareRobustCommand\circled[1]{\tikz[baseline=(char.base)]{\node[shape=circle,draw=darkblue,minimum size=0.35cm,inner sep=0pt,fill=lightblue] (char) {\fontfamily{phv}\selectfont \scriptsize \textbf{#1}};}}

\newcommand*\circledred[1]{%
\tikz[baseline=(char.base)]{
  \node[shape=circle, draw=BrickRed!60, fill=BrickRed!10, thick, inner sep=1pt] (char) {\scriptsize\textsf{#1}};
}}

\definecolor{oursrow}{HTML}{FFF3E0}
\definecolor{numbercolor}{HTML}{91ADC8}%
\definecolor{iv}{HTML}{7F96B2}
\definecolor{effmod}{HTML}{A087AF}
\definecolor{confounder}{HTML}{A9BCA0}
\definecolor{othernodes}{HTML}{B2C8DF}
\definecolor{note}{HTML}{F29AAE}
\definecolor{circnumcolor}{HTML}{215CAF}

\newcommand{\iv}[1]{\textcolor{iv}{#1}}
\newcommand{\effmod}[1]{\textcolor{effmod}{#1}}
\newcommand{\confounder}[1]{\textcolor{confounder}{#1}}

\newcommand{\circnum}[2][circnumcolor]{%
  \tikz[baseline=(N.base)]\node[
    circle,
    fill=#1,
    text=white,
    inner sep=0.25ex,
    font=\sffamily\itshape\scriptsize
  ](N){#2};%
}

\title{ConfoundingSHAP: Quantifying confounding strength in causal inference}

\author{%
\begin{tabular}{c}
\textbf{Marie Brockschmidt}\textsuperscript{1,2,*},
\textbf{Santo M.A.R. Thies}\textsuperscript{1,2,3},
\textbf{Maresa Schröder}\textsuperscript{1,2},
\textbf{Dennis Frauen}\textsuperscript{1,2}
\\[-0.1em]
\textbf{Valentyn Melnychuk}\textsuperscript{1,2},
\textbf{Maximilian Muschalik}\textsuperscript{1,2},
\textbf{Eyke Hüllermeier}\textsuperscript{1,2,3},
\textbf{Stefan Feuerriegel}\textsuperscript{1,2}
\\[0.7em]
{\normalfont\mdseries
\textsuperscript{1}LMU Munich
\qquad
\textsuperscript{2}Munich Center for Machine Learning (MCML)}
\\[0.2em]
{\normalfont\mdseries
\textsuperscript{3}German Research Center for Artificial Intelligence (DFKI)}
\\[0.5em]
{\normalfont\mdseries
\textsuperscript{*}Corresponding author: \texttt{marie.brockschmidt@lmu.de}}
\end{tabular}
}

\begin{document}

\maketitle

\begin{abstract}
In causal inference, confounders are variables that influence both treatment decisions and outcomes. However, unlike as in randomized clinical trials, the treatment assignment mechanism in observational studies is not known, and it is thus unclear which covariates act as confounders. Here, we aim to generate insight for causal inference and answer: which of the observed covariates act as confounders? We introduce \confoundingshap, a Shapley-based method for attributing confounding strength to individual covariates. Our contributions are twofold. First, we propose a Shapley game targeted to infer the confounding strength of the covariates. Our resulting Shapley values differ from the standard applications of SHAP explanations on causal targets, such as understanding treatment effect heterogeneity, which are ill-suited for our task. Second, as our task requires evaluating the value function over many adjustment sets, we provide a scalable TabPFN-based estimation that avoids exhaustive refitting. We demonstrate the practical value across various datasets, where \confoundingshap provides informative explanations of which observed covariates drive confounding and thereby helps to provide more insight for causal inference in practice.
\end{abstract}

\section{Introduction} 
\label{sec:introduction}

\begin{wrapfigure}{r}{0.4\linewidth}
  \centering
  \vspace{-1em}
  \includegraphics[width=0.49\linewidth]{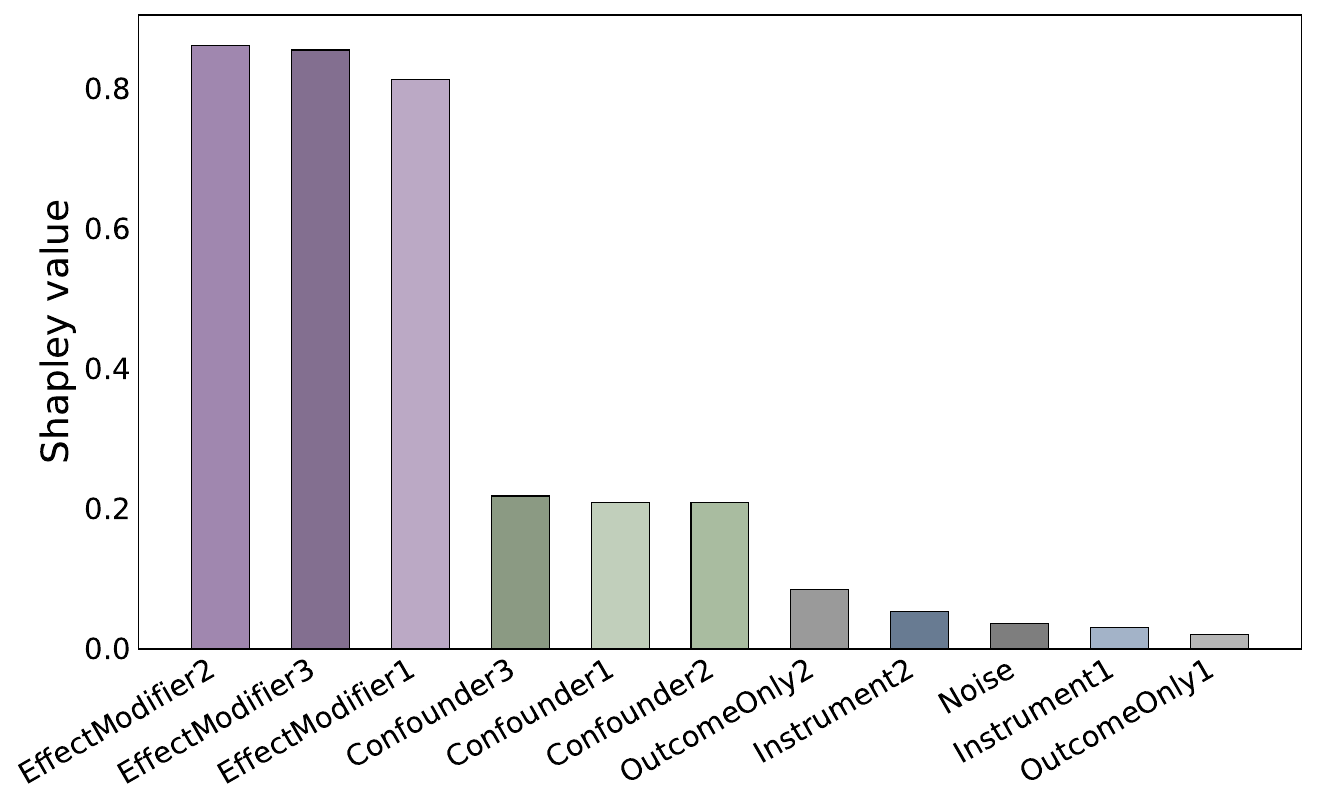}
  \hfill
  \includegraphics[width=0.49\linewidth]{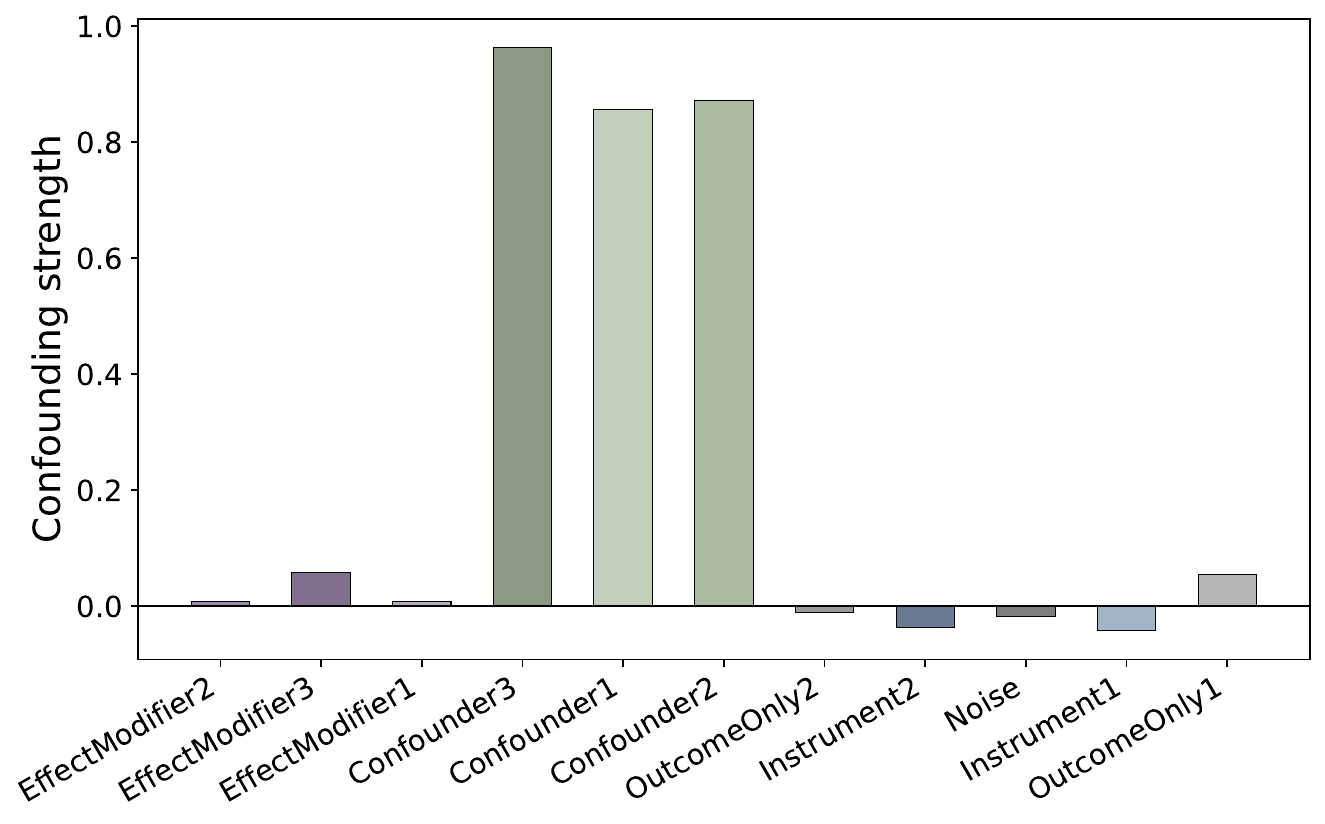}
  \caption{Shown is the difference between variable importance obtained by applying SHAP to a CATE estimator \cite{kennedy2023} vs. of confounding strength (\textit{ours}).}
  \label{fig:cancellingconfounder}
\end{wrapfigure}

\vspace{-0.3em}
In causal inference, treatment assignment is often not random, so treated and untreated groups can differ systematically in their characteristics \citep{imbens2015, pearl2009, rubin1974}. This gives rise to observed \textbf{confounding}, where variables influence both treatment assignment and the outcome, making it difficult to attribute observed differences to the treatment itself \citep{vanderweele2013}.\footnote{Throughout this paper, we mainly rely on the quantitative definition of confounding from Definition 5 in \citep{vanderweele2013}.}

\vspace{-0.3em}
To generate new insight for causal inference, we ask: \emph{\textbf{which observed covariates drive confounding strength?}}

\vspace{-0.3em}
\textbf{Example:} \emph{In oncology, targeted cancer therapies are often known to improve survival, but clinicians are more likely to prescribe these treatments to patients with more advanced disease or worse overall health, such as those with later tumor stage  \citep{hu2023}. These characteristics are also strongly related to patient outcomes and therefore act as confounders.}
\begin{figure}[htbp]
  \vspace{-3em}
  \centering
  \includegraphics[width=\linewidth]{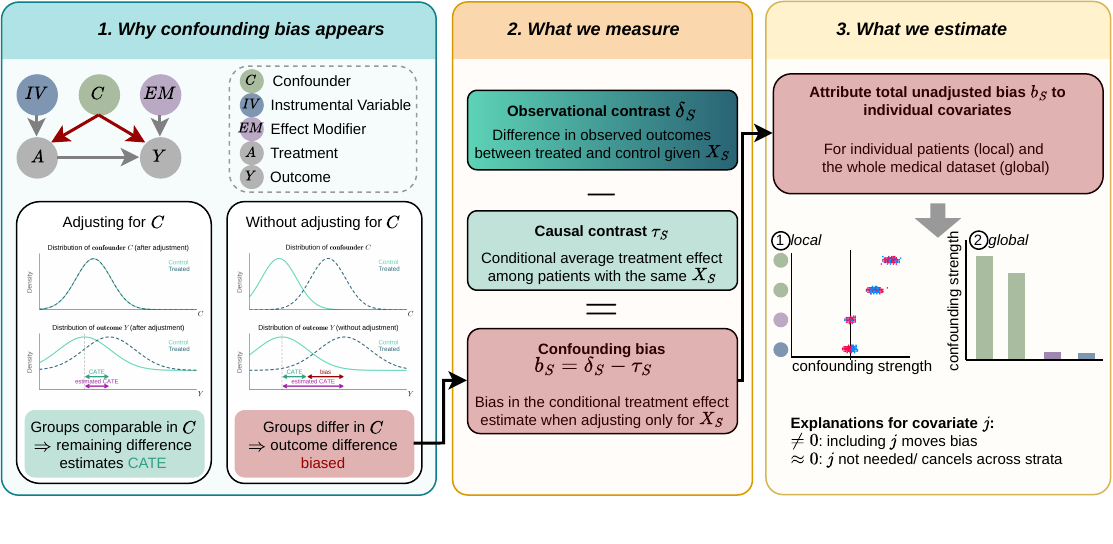}
  \vspace{-3em}
  \caption{\textbf{Overview of our \confoundingshap method.} Therein, residual confounding bias is defined as the gap between observational and causal contrasts and is attributed to individual variates locally and globally.}
  \vspace{-1.5em}
  \label{fig:overview}
\end{figure}

Identifying which variables drive confounding strength helps determine which patient characteristics are needed to make treatment groups comparable, for example, when choosing a minimal sufficient adjustment set in observational studies or conducting prospective observational studies, such as emulated target trials? \citep{chen2016a, feuerriegel2024, hemkens2018, hernan2016, textor2017}.

\vspace{-0.3em}
Several explanation methods have been developed for causal machine learning \citep{paillard2025, rehill2024}, but with \textit{different} target objects. Examples are methods for variable importance of propensity scores (to explain the treatment assignment) \citep{gutman2025}, of outcome models (to understand the outcome prediction) \citep{hu2022}, and of the conditional average treatment effect (CATE) itself (to understand treatment effect heterogeneity) \citep{svensson2025}. However, \underline{none} of these help explain confounding strength.

\vspace{-0.3em}
Explaining confounding strength is fundamentally different from explaining treatment effect heterogeneity, such as applying SHAP on CATE estimators. The reason is that explanations of the CATE target variation in treatment effects, \textbf{not} the confounding strength. Thus, CATE explanations will be dominated by \textbf{\effmod{effect modifiers}} and include \textbf{\confounder{observed confounders}}, but will \textbf{not isolate} them. We demonstrate this difference in Fig.~\ref{fig:cancellingconfounder}. In a special case, explanations of a fitted CATE estimator can even miss confounders. If a confounder shifts treated and untreated outcomes in a similar way, the contributions cancel out in the CATE, and the \textbf{\confounder{observed confounder}} does not show up in the variable importance for treatment effect heterogeneity (see Supplement~\ref{sec:cancellingconfounder}). Hence, a na{\"i}ve application of existing SHAP \citep{lundberg2017} explanations would only explain effect heterogeneity and is thus \emph{ill-suited} for our task of quantifying confounding strength.

\vspace{-0.3em}
In this paper, we propose \confoundingshap: \emph{a Shapley-based method for attributing observed confounding strength to individual covariates}. For this, we define a cooperative game directly over the functional that captures confounding strength. In our method, the value function thus quantifies, for each coalition of covariates, the residual confounding bias that remains in CATE estimation when adjustment is restricted to that coalition. \confoundingshap then decomposes the bias functional into covariate-level Shapley contributions, which yields an attribution of which observed covariates drive confounding bias and by how much. We provide a scalable computational algorithm for this task. Specifically, since our task requires evaluating the value function over many adjustment sets, we propose an efficient TabPFN-based approximation that avoids exhaustive refitting. Our method is designed for the CATE, but can also trivially yield attributions for the average treatment effect (ATE) simply by averaging the bias functional over the covariate distribution.

\vspace{-0.3em}
Our \textbf{main contributions}\footnote{Code: \url{https://anonymous.4open.science/r/ConfoundingSHAP-17D3}. Upon acceptance, we will make our code public as part of a Python package.} are the following. \textbf{(1)}~We introduce \confoundingshap: a novel Shapley-based method to measure the confounding strength at the covariate level.  \textbf{(2)}~We propose a novel scalable approximation for \confoundingshap based on TabPFN estimators, which avoids the exhaustive refitting required by the exact bias functional while preserving the interpretation of the underlying cooperative game. \textbf{(3)}~We demonstrate the practical value across various numerical experiments, where \confoundingshap provides informative explanations of which observed covariates drive confounding strength.

\section{\texorpdfstring{Related Work\protect\footnote{We provide an extended related work in Supplement~\ref{sec:extendedrelatedwork}.}}{Related Work}}
\label{sec:related_work}

\textbf{Explanation methods for variable importance.} Explanation methods are widely used for measuring variable importance with respect to a fitted model or predictive task. A first class are methods with local surrogates. A prominent example is LIME \citep{ribeiro2016}, which provides instance-level explanations. A second class are \emph{Shapley value decompositions}. A prominent example is SHAP \citep{lundberg2017}, which explains local or global model outputs by attributing contributions to individual features based on cooperative game theory at the feature level. Related approaches such as SAGE \citep{covert2020} extend this idea to define global importance for a predictive utility. For a detailed review, see Supplement~\ref{sec:extendedrelatedwork}. \emph{$\Rightarrow$ However, these methods explain predictive models rather than confounding strength.}

\begin{wraptable}{r}{0.65\textwidth}
\vspace{-0.5em}
\centering
\scriptsize
\setlength{\tabcolsep}{2.5pt}
\begin{tabular}{@{}p{0.22\linewidth}p{0.2\linewidth}p{0.34\linewidth}p{0.13\linewidth}@{}}
\toprule
Target & $\nu(S)$ & $\phi_j$ explains & Example(s) \\
\midrule
\circledred{1} Propensity score &
$\mathbb{E}[\hat{\pi}(X) \mid X_S]$ &
treatment assignment &
\citep{gutman2025, martini2022} \\

\circledred{2} Outcome model &
$\mathbb{E}[\hat\mu_a(X)\mid X_S]$ &
outcome prediction &
\citep{schrod2022} \\

\circledred{3} CATE &
$\mathbb{E}[\hat\tau(X) \mid X_S]$ &
effect heterogeneity &
\citep{benard2025,svensson2025} \\

\midrule
\cellcolor{oursrow}\textbf{Confounding bias} &
\cellcolor{oursrow}$-\mathbb{E}[b_S(X_S)]$ &
\cellcolor{oursrow}bias under different adjustments &
\cellcolor{oursrow}\textit{this paper} \\
\bottomrule
\end{tabular}
\vspace{-0.5em}
\caption{\textbf{Differences \confoundingshap vs. key related work.} Shown is the semantics of Shapley values in causal inference depending on the causal target.}
\label{tab:semantics}
\end{wraptable}

\textbf{Causal Shapley values.} A different literature stream develops causal formulations of Shapley values, for example, by incorporating causal structure into feature attribution through causal graphs, asymmetric coalitions, or intervention-based value functions \citep{frye2020, heskes2020, parafita2025, witter2026}. \emph{$\Rightarrow$ However, our work is orthogonal: our focus is \textbf{not} on causal formulations of Shapley attributions, but to use Shapley attributions to understand causal questions.}

\vspace{-0.3em}
\textbf{Explanations for causal targets.} Several methods focus on different causal targets (see Table~\ref{tab:semantics}), based on which Shapley values have \textit{different} semantics:

\vspace{-0.3em}
\textbf{\circledred{1} Variable importance for propensity scores (treatment assignment).}
Feature importance methods have been used to explain treatment assignment/ propensity score models; for example, to identify variables predictive of treatment allocation and guide propensity score analyses \citep{gutman2025,martini2022}. \textit{$\Rightarrow$ However, these explanations target who receives treatment, but \textbf{not} whether these variables induce confounding (e.g., an instrument may strongly predict treatment without directly affecting the outcome).}

\vspace{-0.3em}
\textbf{\circledred{2} Variable importance for outcome models (outcome prediction).}
In medical applications of causal machine learning, SHAP has been used to explain treatment-specific outcome models, i.e., the predicted potential outcomes under a given intervention (such as, e.g., survival or adverse event risk under alternative treatment options \citep{schrod2022}). In clinical terminology, such explanations primarily identify \textit{prognostic} variables (i.e., variables associated with outcome risk), but not necessarily \textit{predictive} variables, (i.e., variables that predict differential treatment responses) \citep{sechidis2018}.

\vspace{-0.3em}
\textbf{\circledred{3} Variable importance for effect heterogeneity (CATE).}
Several works study variable importance for CATEs to identify effect modifiers that drive variation in individualized treatment effects \citep{benard2025, hines2025, morzywolek2025, paillard2025, sechidis2025, svensson2025}. These methods are useful for understanding the structure of the estimated CATE, but they do \textbf{not} directly explain which covariates drive confounding strength. $\Rightarrow$ \emph{Variable importance for CATE heterogeneity and for confounding strength answer distinct but complementary questions.}

\vspace{-0.3em}
\textbf{Research gap:} To the best of our knowledge, there is \underline{no} Shapley-based method to explain observed confounding strength. To fill this gap, we propose a method to attribute confounding strength under different adjustment strategies to individual covariates.

\section{Setting}
\label{sec:setting}
\textbf{Preliminaries: Shapley values.}
For a cooperative game with $p$ players and value function $\nu : 2^{[p]} \to \mathbb{R}$, the \textit{Shapley value} of player $j$ is the unique allocation satisfying efficiency, symmetry, linearity, and the dummy axiom
\citep{shapley1953}:
\begin{equation}
\phi_j(\nu)
=
\sum_{S \subseteq [p]\setminus \{j\}}
\frac{1}{p \binom{p-1}{|S|}}
\left[
\nu(S \cup \{j\}) - \nu(S)
\right].
    \label{eq:shapley}
\end{equation}
The efficiency property guarantees $\sum_{j=1}^p \phi_j(\nu) = \nu([p]) - \nu(\emptyset)$, meaning that the covariate-level Shapley values add up exactly to the change in the value function from using no covariates to using the full covariate set. In our setting, this is crucial because it allows us to interpret the Shapley values as a decomposition of the total confounding reduction across individual covariates.

\textbf{Setup.} We consider the standard causal inference setting for CATE estimation \citep{rubin1974}. with population $Z = (X, A, Y) \sim \mathbb{P}$, where $X \in \mathcal{X} \subseteq \mathbb{R}^p$ are observed pre-treatment covariates, $A \in \{0,1\}$ is a binary treatment indicator, and $Y \in \mathbb{R}$ is a scalar outcome. For a subset $S \subseteq [p]$, we write $X_S \in \mathbb{R}^{|S|}$ for the corresponding subvector and $X_{-S}$ for its complement. We observe an i.i.d.\ sample $\mathcal{D} = \{(x_i, a_i, y_i)\}_{i=1}^n$ of size $n \in \mathbb{N}$.

\vspace{-0.3em}
\textbf{CATE.}
We work within the potential outcomes framework \citep{rubin1974, rubin2005}. Let $Y(0)$ and $Y(1)$ denote the \textit{potential outcomes} under control and treatment, respectively. Let $\pi(x) := \mathbb{P}(A=1 \mid X=x)$ denote the \textit{propensity score}, and let $\mu_a(x) := \mathbb{E}[Y \mid X=x,A=a]$ denote the \textit{outcome regression} in treatment arm $a \in \{0,1\}$. The \textit{conditional average treatment effect (CATE)} is given by $\tau(x) := \mathbb{E}[Y(1)-Y(0) \mid X=x] = \mu_1(x) - \mu_0(x),$ and the \textit{average treatment effect (ATE)} by $\tau := \mathbb{E}[\tau(X)]$.

\vspace{-0.3em}
Causal effects are identified under consistency, positivity, and no unmeasured confounding conditional on the full observed covariate vector $X$ \citep{rosenbaum1983a, rubin1974, vanderlaan2006}. Identification also holds, equivalently, when the full covariate set $X$ is the valid adjustment set. Importantly, a strict subset $X_S$, $S \subsetneq [p]$, may no longer be a valid adjustment set and thus introduces confounding in the CATE, which we aim to quantify.

\vspace{-0.3em}
\textbf{Objective.} Our aim is to \emph{quantify the contribution of individual covariates to confounding strength}. We first note the fact that restricting the covariate set to $X_S$, $S \subsetneq [p]$, changes which observed differences between treated and untreated groups are captured. The resulting change in the treatment effect estimate provides a signal of the confounding strength associated with the omitted covariates. Our goal is to attribute this observed confounding strength to individual covariates.

\vspace{-0.3em}
We formalize \textbf{confounding bias} \citep{imbens2015, pearl2009, rubin1974} at the level of a covariate subset $X_S$ as the difference between two quantities defined at the same information level: the \emph{observational contrast} and the \emph{causal contrast}. We define both as follows.

\vspace{-0.3em}
$\bullet$ \textbf{Observational contrast}. Let the \textit{observational contrast}\footnote{Importantly, $\delta_S(x_S)$ is an observational quantity: it compares observed outcomes between the actually treated and actually untreated units conditional on $X_S=x_S$, rather than comparing the potential outcomes $Y(1)$ and $Y(0)$ for the same target population.} be the subgroup effect wrt. $X_S$, namely the conditional difference in observed outcomes between treated and control units \emph{given the restricted adjustment set $X_S$}, i.e.,

\vspace{-1.5em}
\begin{equation}
\delta_S(x_S) 
:=
\mathbb{E}[Y \mid A = 1, X_S = x_S]
-
\mathbb{E}[Y \mid A = 0, X_S = x_S].
\label{eq:delta}
\end{equation}

\vspace{-0.3em}
This quantity captures the observed treatment-control difference within groups defined by $X_S$. If adjustment by $X_S$ is insufficient, $\delta_S(x_S)$ will reflect systematic differences between treated and control units, and therefore the observational contrast is \underline{not} equal to the causal contrast.

\vspace{-0.3em}
$\bullet$ \textbf{Causal contrast}. Let the \textit{causal contrast} be the projected CATE wrt. $X_S$. We use the full covariate set $X$ as the reference adjustment set. We then define the causal contrast at information level $X_S$ as the projection of the full-adjustment CATE onto $X_S$, i.e.,

\vspace{-1.5em}
\begin{equation}
\tau_S(x_S)
:=
\mathbb{E}[\tau(X) \mid X_S = x_S].
\label{eq:tauS}
\end{equation}
Here, $\tau(X)$ denotes the CATE when adjusting by the full covariate set $X$. Thus, $\tau_S(x_S)$ is the average fully-adjusted treatment effect but only among units sharing $X_S=x_S$.
\vspace{-0.3em}

Then, the \emph{confounding bias} can be quantified as a residual bias from confounding under adjustment by $X_S$, i.e.,

\vspace{-2em}
\begin{equation}
b_S(x_S)
:=
\delta_S(x_S) - \tau_S(x_S).
\label{eq:bias}
\end{equation}

\vspace{-0.3em}
\textbf{Interpretation of $b_S(x_S) = 0$.} The quantity $b_S(x_S)$ captures the residual bias that remains when we only include $X_S$ to the adjustment set rather than all covariates $X$. Importantly, if $X_S$ contains {all the confounders} (e.g., $X_S$ is a sufficient adjustment set or when $S=[p]$), then $b_S(x_S)$ is zero:

\vspace{-1.5em}
\begin{equation} \label{eq:interpretation}
    X_S \text{ contains all the confounders} \quad \Rightarrow \quad b_S(x_S) = 0
\end{equation}

\vspace{-0.3em}
However, the converse does not hold in general, and $b_S(x_S)=0$ could thus mean two things: (i) either $X_S$ is a sufficient adjustment set, or (ii) $X_S$ is actually insufficient, yet the confounding bias caused by other covariates outside of $X_S$ perfectly cancels out during averaging (e.g., we can have $b_{\emptyset}(x_{\emptyset}) = b_{[p]}(x) = 0$, but $b_{S}(x_{S}) > 0$ for $S \subset [p]$; see our counterexample in Supplement~\ref{app:cancel}). 

\vspace{-0.3em}
\textbf{Interpretation of $b_S(x_S) \neq 0$.} By the contraposition of Eq.~\eqref{eq:interpretation}, we know that, if $b_S(x_S) \neq 0$, some confounders are missing in $X_S$. Again, the converse is not generally true: if certain confounders are not included in $X_S$, then $b_S(x_S)$ can be both non-zero or even zero (due to perfect cancellation of confounding from another variable).

\vspace{-0.3em}
\textbf{Relation to sensitivity models.} Interestingly, our quantifier of observed confounding (i.e., residual bias $b_S$) has a connection to existing sensitivity models \citep{mcclean2026} (where the main aim is to upper-bound the strength of unobserved confounding). Specifically, our residual bias $b_S$ is the closest to the sensitivity model of \citet{ludtke2015}: it is easy to see that their sensitivity parameter upper-bounds the \emph{absolute} average of our residual bias $\lvert \mathbb{E}[b_S(X_S)] \rvert$. However, our \emph{signed} values of residual bias allow to quantify not only the magnitude but also the direction of the confounding bias.

\vspace{-0.3em}
\textbf{Difference to SHAP-based analysis of CATEs.} The above formulation highlights that our task is fundamentally \textit{different} from feature attribution for fitted CATE estimators \citep{hines2025, liu2025, morzywolek2025}. Our method does \textit{not} explain variation in $\tau(x)$, i.e., causal contrasts, but the residual confounding bias $b_S(x_S)$ induced by restricted adjustment, i.e., the difference between observational and causal contrasts. This distinction matters because cancellations have different implications: in \confoundingshap, they occur on the residual bias scale and inform the net bias from restricted adjustment, whereas, in CATE attributions, they only reflect limited contribution to treatment effect heterogeneity. For example, if a covariate shifts both potential outcomes similarly, the contributions to $\mu_1(x)$ and $\mu_0(x)$ may largely disappear in $\tau(x)=\mu_1(x)-\mu_0(x)$. Then, the causal contrast for a subset $S$ excluding all confounders may coincide with the full-covariate contrast (i.e., $\tau_S=\tau_{[p]}$) and thus yields zero CATE SHAP, even though the observational contrasts $\delta_S$ and $\delta_{[p]}$ differ (see example in Fig.~\ref{fig:cancellingconfounder}). 
A detailed discussion is in Supplement~\ref{app:nuisance-shap-relation}.

\section{\confoundingshap}
\label{sec:method}

\textbf{Overview.}
We introduce \confoundingshap, a Shapley-based method for attributing confounding strength to individual covariates. The method consists of three steps (see Fig.~\ref{fig:method}). \circnum{A} We first define the residual confounding bias functional $b_S$ that compares the \emph{observational contrast} to the corresponding \emph{causal contrast}, both with the same information level $X_S$. \circnum{B} We then construct a cooperative game for which the value function is defined directly by this bias functional, so that Shapley values attribute bias reduction to individual covariates. \circnum{C} We estimate coalition values for the value function to replace potentially expensive refitting per coalition with an in-context approach (using TabPFN as a state-of-the-art tabular foundation model).

\vspace{-0.3em}
The key idea is that the explanation target is not a fitted prediction, propensity, or CATE model. Instead, \confoundingshap defines feature importance with respect to a causal bias functional. The resulting attributions therefore quantify contribution to confounding strength in the data rather than predictive relevance, treatment assignment relevance, or treatment effect heterogeneity.

\vspace{-0.3em}
\subsection*{\circnum{A} Confounding bias under restricted adjustment}
\label{sec:bias-functional}

\vspace{-0.3em}
First, we briefly recall the residual confounding bias from Section~\ref{sec:setting}. For a covariate subset $S \subseteq [p]$, the observational contrast $\delta_S(x_S)$ compares treated and control units conditional on $X_S=x_S$, whereas the causal contrast $\tau_S(x_S)$ is the full-adjustment CATE projected onto the same subset. The difference (i.e., $b_S(x_S)=\delta_S(x_S)-\tau_S(x_S)$) measures the residual confounding bias associated with using $X_S$ as the adjustment set (see Panel~\circnum{A}  in Fig.~\ref{fig:method}). If $X_S$ is sufficient for adjustment, the two contrasts agree and $b_S(x_S)=0$; if relevant confounders are missing, residual bias may remain. The collection $\{b_S(X_S):S\subseteq[p]\}$ therefore describes how confounding bias changes across adjustment sets and provides the input for the cooperative game in step~\circnum{B}.

\begin{figure}[htbp]
  \centering
  \includegraphics[width=1.0\linewidth]{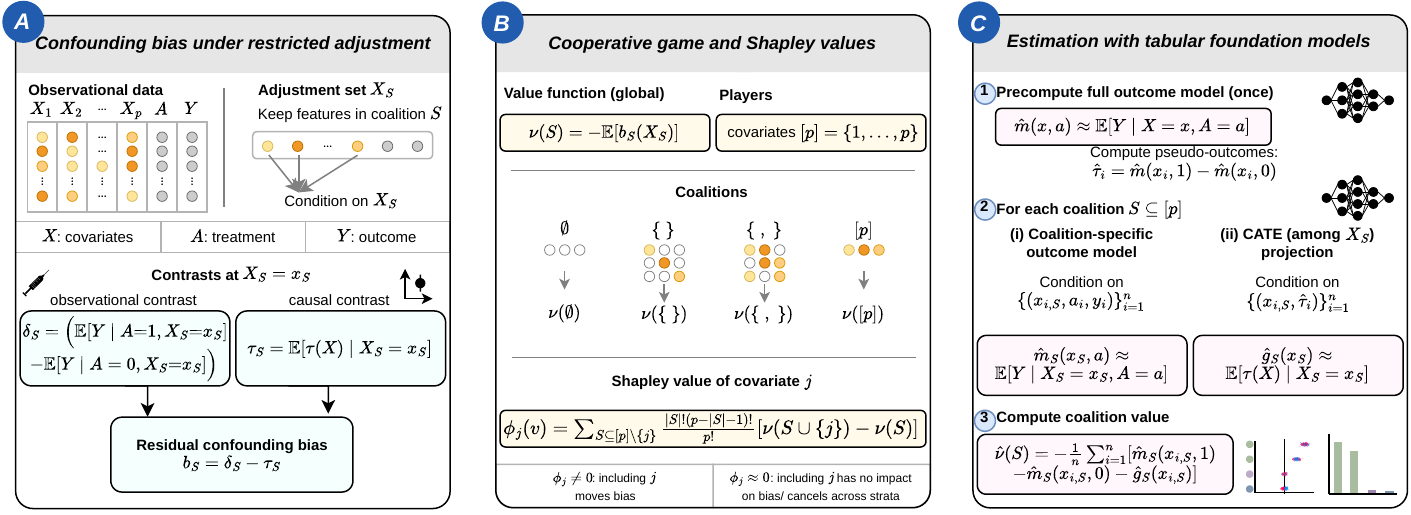}
  \caption{\textbf{Overview of \confoundingshap.} Our method defines confounding bias under restricted adjustment (\circnum{A}), introduces a Shapley game over this bias functional (\circnum{B}) and then estimates coalition values with tabular foundation models (\circnum{C}).}
  \label{fig:method}
  \vspace{-0.5em}
\end{figure}

\subsection*{\circnum{B} A cooperative game over residual confounding bias}
\label{sec:game}

\vspace{-0.5em}
We now define the cooperative game in which the players are the covariates $j \in [p]$. A coalition of players $S \subseteq [p]$ represents the set of covariates that are included in the adjustment set. The goal of the cooperative game in \confoundingshap is to attribute the residual confounding bias to individual covariates by comparing coalitions that differ only in whether a given covariate is included. We define the game as follows:

\vspace{-0.3em}
\textbf{Coalition values.}
For an individual with covariates $x$, we define the datapoint-level coalition value as

\vspace{-2em}
\begin{equation}
\nu_x(S)
:=
-b_S(x_S).
\label{eq:local-value}
\end{equation}

\vspace{-0.3em}
The corresponding population-level coalition value averages over the distribution of
$X_S$, i.e.,

\vspace{-1.3em}
\begin{equation}
\nu(S)
:=
-\mathbb{E}\left[b_S(X_S)\right].
\label{eq:global-value}
\end{equation}

\vspace{-0.3em}
\emph{Why the negative sign?}
We define $\nu(S)=-\mathbb{E}[b_S(X_S)]$ with a negative sign, so that, when comparing a smaller subset $S$ to the larger subset $S\cup\{j\}$, a positive contribution means that the confounding bias is attributed to not yet including $j$. Hence, the sign of the resulting Shapley value matches the direction of the bias attributed to that covariate.

\vspace{-0.3em}
\textbf{Covariate-level decomposition.} The efficiency property of Shapley values is the main reason for using the above construction. The property states that the covariate-level attributions sum to the difference between the full and empty conditioning sets, i.e., $\sum_{j=1}^{p} \phi_j(\nu) = \nu([p]) - \nu(\emptyset)$. Since the full conditioning set is the reference adjustment set, we have $b_{[p]}(X)=0$ and hence $\nu([p])=0$. Therefore, $\sum_{j=1}^{p} \phi_j(\nu) = 0 - \left(-\mathbb{E}[b_{\emptyset}(X_{\emptyset})]\right) = b_{\emptyset}$. Thus, the Shapley values decompose the population-level residual bias of the unconditioned comparison into covariate-level contributions. Hence, the Shapley values explain how the difference between no conditioning and full conditioning is distributed across the covariates. Analogously, local Shapley values obtained from the datapoint-level values $\nu_x$ decompose the residual bias for a specific covariate profile $x$: $\sum_{j=1}^{p} \phi_j(\nu_x) = \nu_x([p]) - \nu_x(\emptyset) = b_{\emptyset}$.

\vspace{-0.3em}
\textbf{Causal interpretation of value functions, their differences, and Shapley values.} The value functions $\nu_x(S)=-b_S(x_S)$ and $\nu(S)=-\mathbb{E}[b_S(X_S)]$ measure residual confounding bias relative to the full covariate set. The local value $\nu_x(S)$ describes, for a specific covariate profile $x$, how much signed residual bias remains when the treatment-control comparison is made only within groups defined by $X_S$, or equivalently when the covariates outside \(S\) are withheld from the adjustment set. At the same time, the global value $\nu(S)$ is the corresponding population average and quantifies the residual confounding bias wrt. ATE. Hence, the differences of global values (i.e., $\nu(S\cup\{j\})-\nu(S)$) measure how the \emph{signed residual bias changes when covariate $j$ is added to the adjustment set}. Finally, the Shapley value $\phi_j$ averages these marginal differences over all possible adjustment contexts $S$, and therefore \emph{quantifies the average contribution of covariate $j$ to removing signed residual confounding bias}.

\vspace{-0.3em}
\textbf{Cancellations and Shapley values.} As previously discussed, $b_S(x_S) = 0$ could mean both (i) the absence and (ii) a perfect cancellation of confounders outside of $X_S$. Yet, it does \emph{not} by itself imply that all covariates outside $S$ have zero Shapley values. A zero Shapley value $\phi_j=0$ can be observed when (a) the marginal contributions of covariate $j$ cancel on average across the Shapley aggregation, or when (b) $j$ is a null player for the bias game. Hence, a covariate can have $\phi_j=0$ \emph{even if it changes residual bias for some coalitions}, as long as positive and negative marginal contributions cancel. Conversely, $\phi_j\neq 0$ indicates that covariate $j$ has a systematic signed contribution to changing residual confounding bias across adjustment contexts. Thus, cancellations can lead to $b_S(x_S)=0$ and can also make individual Shapley values vanish, but these are distinct forms of cancellation. 

\vspace{-0.3em}
\textbf{Why are cancellations not a problem?} In practice, our method is still useful because nonzero $\phi_j$ identifies covariates with systematic directional contributions to residual confounding bias, while zero $\phi_j$ highlights either true irrelevance or contributions that cancel out on average. Hence, the causal interpretation of our \confoundingshap has a practical relevance for causal estimation.

\vspace{-0.3em}
\subsection*{\circnum{C} Estimating coalition values (with TabPFN)}
\label{sec:estimation}

\vspace{-0.5em}
In standard SHAP \citep{lundberg2017} explanations, the cooperative game is usually defined with respect to a fitted model. To evaluate a coalition $S$, the model is queried using the features in $S$, while features outside $S$ are treated as unavailable, for example by masking them or averaging over their conditional distribution. The coalition value $\nu(S)$ is then obtained from the resulting model output. Our target is more challenging because confounding bias is not the output of a single fitted model, so coalition values cannot be obtained by simply masking features in a fixed predictor. To see this, note that changing $S$ changes the conditioning set and therefore the contrasts that define $b_S$. Thus, for each coalition $S$, computing $\nu(S)$ may require fitting new coalition-specific models for the observational contrast and the projected causal contrast. Hence, a na\"ive implementation would refit these models from scratch for every coalition, which can quickly become computationally prohibitive for large or expensive regressors.\footnote{We address the separate question of how many coalitions need to be evaluated for Shapley estimation below, and provide a computational complexity analysis in Supplement~\ref{sec:complexity}.}

\vspace{-0.3em}
To make coalition-value evaluation scalable, we use TabPFN \citep{grinsztajn2026, hollmann2023, hollmann2025} as our regression backbone. TabPFN is well suited to this setting because each coalition evaluation can be viewed as a new tabular regression task with a different input subset $X_S$ and, depending on the contrast, a different regression target. Instead of  retraining from scratch for every such task, TabPFN uses a pretrained tabular foundation model with frozen parameters and produces predictions by conditioning on the training data provided in context. In our implementation, this allows us to evaluate the coalition-specific regressions needed for $\nu(S)$ \emph{without} updating model parameters for every coalition. Our estimation procedure has three sub-steps (see Panel~\circnum{C} in Fig.~\ref{fig:method}).

\vspace{-0.3em}
\textbf{\circled{1} Precompute the full outcome model.}
We first estimate the full outcome regression

\vspace{-1.5em}
\begin{equation}
m(x,a)
:=
\mathbb{E}[Y \mid X=x, A=a],
\label{eq:full-outcome-model}
\end{equation}

\vspace{-0.3em}
using the full covariate vector $X$. Here, the model $m$ is fitted once and is used to form the full-adjustment CATE estimates $\hat{\tau}_i := \hat{m}(x_i,1)-\hat{m}(x_i,0).$ The $\hat{\tau}_i$ (also called pseudo-outcomes) provide the reference causal signal that is projected onto covariate subsets, and the population average corresponds to $\bar{\tau}=\mathbb{E}[\tau(X)]$.

\vspace{-0.3em}
\textbf{\circled{2} Estimate coalition-specific quantities.}
For each coalition $S \subseteq [p]$, we estimate the quantities needed to evaluate $b_S$.\footnote{For global coalition values, the projected causal contrast does not need to be estimated separately for every coalition, since
$\mathbb{E}[g_S(X_S)]
=
\mathbb{E}[\mathbb{E}[\tau(X)\mid X_S]]
=
\mathbb{E}[\tau(X)]$.
Thus, the projection step is only required for local coalition values, improving efficiency when only population-level values are needed.} First, we estimate the coalition-specific outcome model

\vspace{-1.5em}
\begin{equation}
m_S(x_S,a)
:=
\mathbb{E}[Y \mid X_S=x_S, A=a],
\label{eq:coalition-outcome-model}
\end{equation}

\vspace{-0.3em}
which gives the observational contrast $\hat{\delta}_S(x_{i,S}) = \hat{m}_S(x_{i,S},1)-\hat{m}_S(x_{i,S},0).$ Second, we estimate the projection of the full-adjustment CATE onto the same subset, given by

\vspace{-1.5em}
\begin{equation}
g_S(x_S)
:=
\mathbb{E}[\tau(X) \mid X_S=x_S],
\label{eq:coalition-cate-projection}
\end{equation}

\vspace{-0.3em}
where we regress the pseudo-outcomes $\hat{\tau}_i$ on $X_{i,S}$. This yields $\hat{g}_S(x_{i,S})$ as our estimate of the projected causal contrast.

\textbf{\circled{3} Compute coalition values.}
Combining the above two estimates gives the plug-in estimate of the residual confounding bias for coalition $S$, $\hat{b}_S(x_{i,S}) = \hat{\delta}_S(x_{i,S}) - \hat{g}_S(x_{i,S})$.
The population-level coalition value simplifies to ${\nu}(S) = -\mathbb{E} \left[{\delta}_S(x_{S}) - {g}_S(x_{S}) \right] = -\mathbb{E}[{\delta}_S(x_{S}) - \bar{\tau}]$, where $\bar{\tau}=\mathbb{E}[\tau(X)]$ making the corresponding estimator $\hat{\nu}(S)$ computationally more efficient than its local counterpart.

\vspace{-0.3em}
The plug-in construction separates the full outcome model (\circled{1}) from the coalition-specific regressions (\circled{2}). The full outcome model is estimated once and used to compute pseudo-outcomes $\hat{\tau}_i=\hat m(x_i,1)-\hat m(x_i,0)$. For each queried coalition $S$, we then estimate two quantities on $X_S$: the observational contrast $\hat{\delta}_S$ and the projection $\hat g_S$ of the pseudo-outcomes.

\vspace{-0.3em}
We evaluate the coalition-specific regressions with TabPFN \citep{grinsztajn2026, hollmann2023, hollmann2025}. Each coalition defines an in-context regression task on $X_S$: one task for the outcome model entering $\hat{\delta}_S$, and one task for projecting the pseudo-outcomes $\hat{\tau}_i$ onto $X_S$. Thus, changing $S$ changes the regression task, \emph{but does not require retraining}. TabPFN is used only to obtain the coalition-value estimates $\hat{\nu}(S)$ efficiently (\circled{3}). The confounding-bias game itself remains unchanged and is the one defined in Section~\ref{sec:game}.

\textbf{Estimating Shapley values}
\label{sec:approximation}

\vspace{-0.3em}
After estimating coalition values, the remaining computational challenge is the Shapley aggregation itself. Exact Shapley values require evaluating $\nu$ across all $2^p$ coalitions, which becomes infeasible as the number of covariates grows.

\vspace{-0.3em}
We therefore use Shapley estimation: instead of evaluating all coalitions, the estimator queries at most $B \in \mathbb{N}$ coalitions and reconstructs approximate Shapley values from the resulting $\hat{\nu}(S)$ evaluations. Existing approximators provide three useful options for this setting \cite{kolpaczki2026}: Monte-Carlo estimators based on maximum sample reuse (MSR) \citep{kolpaczki2024,strumbelj2014,wang2023}, regression-based estimators based on weighted least-squares formulations \citep{fumagalli2026,fumagalli2026a,lundberg2017,musco2025,witter2025}, and hybrid estimators that combine regression-based proxy models with Monte-Carlo correction \citep{butler2025, witter2025}.

\vspace{-0.3em}
In our case, we use \textsc{RegressionMSR}~\citep{witter2025} for our main experiments in  Section~\ref{sec:experiments}. \textsc{RegressionMSR} fits a proxy $\tilde{\nu}$ to the queried coalition values, extracts Shapley values from this proxy, and applies an MSR correction to the residual game $\hat{\nu}-\tilde{\nu}$. We provide an ablation of this estimator choice in Supplement~\ref{sec:ablation} where we compare \textsc{RegressionMSR} to representatives of the other two classes: \textsc{MSR}~\citep{lundberg2017} as an MSR estimator and \textsc{KernelSHAP}~\citep{lundberg2017} as a weighted least-squares estimator. Importantly, all three estimators are consistent, i.e. recover the exact Shapley values in the limit $B \to 2^p$.

\begin{figure}[!t]
  \vspace*{-1cm}
  \centering

  \newcommand{\panelheight}{2.25cm}

  \begin{subfigure}[t]{0.23\textwidth}
    \centering
    \vspace{0pt}
    \includegraphics[width=\linewidth,height=\panelheight,keepaspectratio]{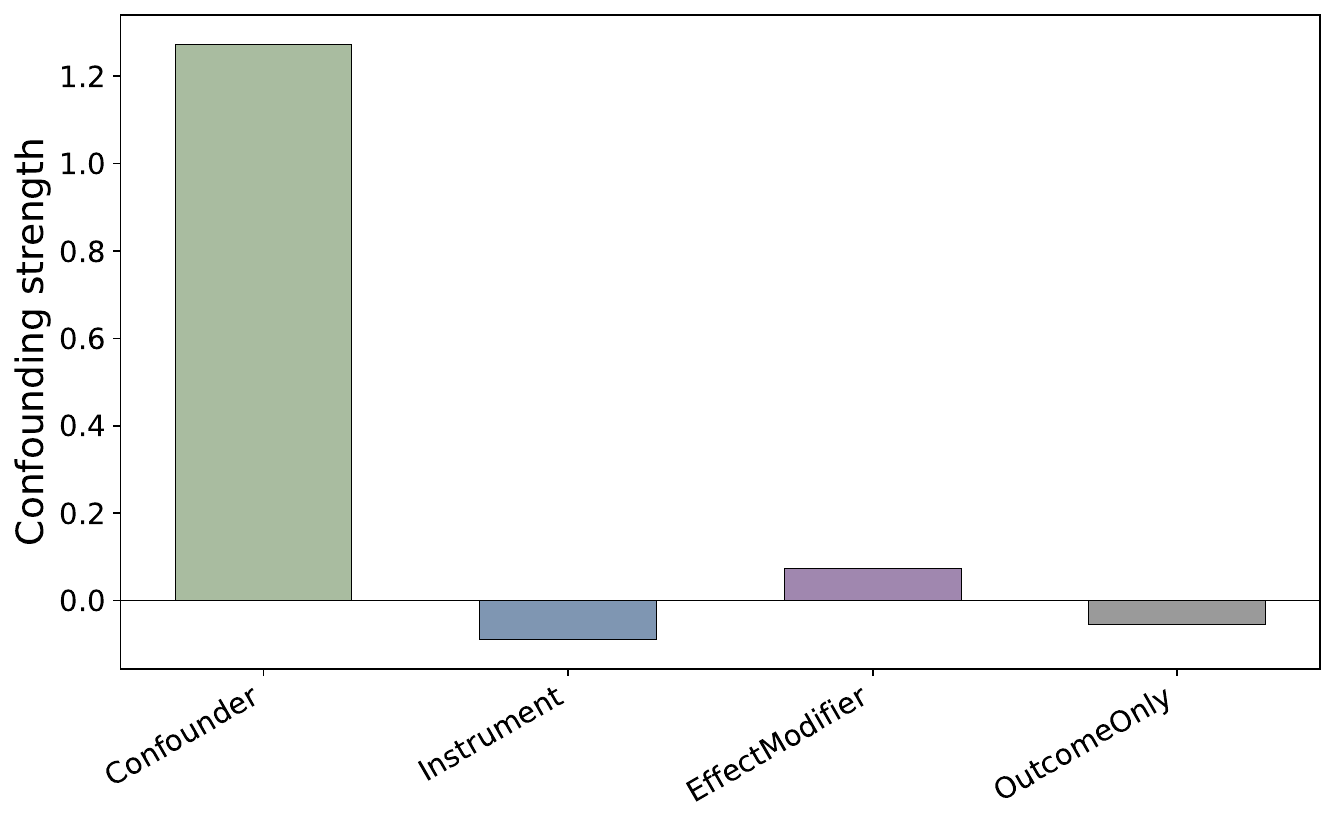}
    \caption{Global values}
    \label{fig:4cov_importance}
  \end{subfigure}
  \hfill
  \begin{subfigure}[t]{0.23\textwidth}
    \centering
    \vspace{0pt}
    \includegraphics[width=\linewidth,height=\panelheight,keepaspectratio]{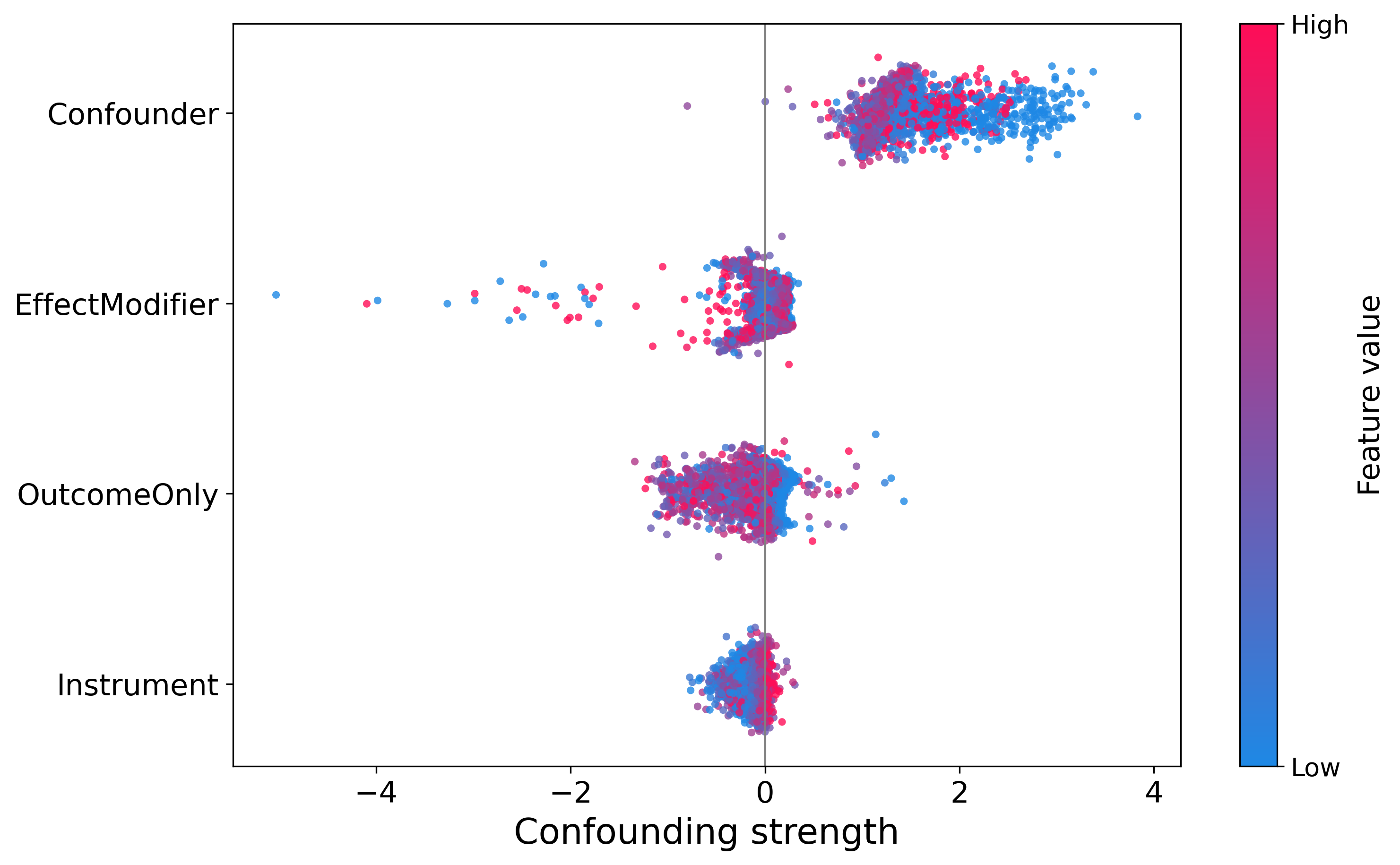}
    \caption{Local values}
    \label{fig:4cov_beeswarm}
  \setcounter{subfigure}{0}
  \end{subfigure}
  \hfill
  \begin{subfigure}[t]{0.23\textwidth}
    \centering
    \vspace{0pt}
    \includegraphics[width=\linewidth,height=\panelheight,keepaspectratio]{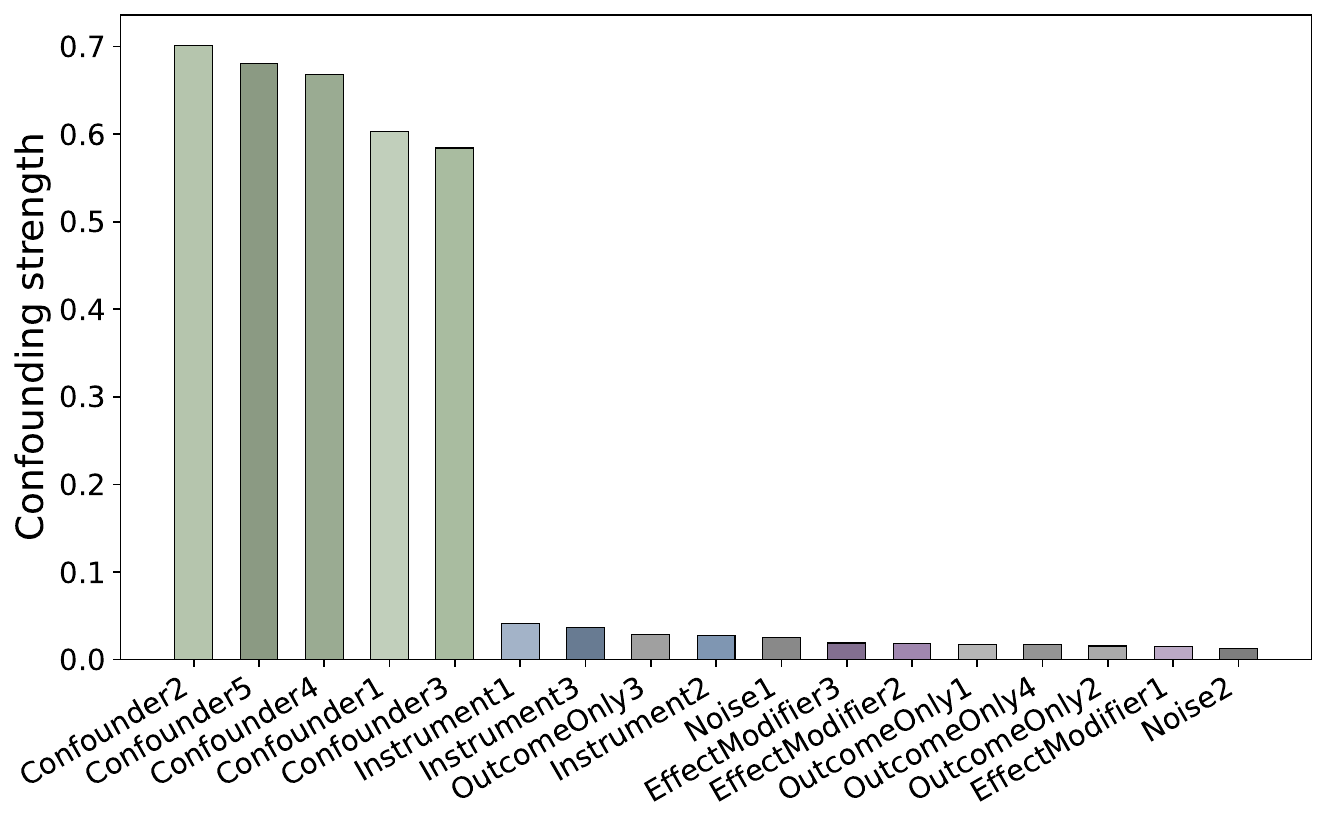}
    \caption{Avg. global values}
    \label{fig:synth17_importance}
  \end{subfigure}
  \hfill
  \begin{subfigure}[t]{0.23\textwidth}
    \centering
    \vspace{0pt}
    \includegraphics[width=\linewidth,height=\panelheight,keepaspectratio]{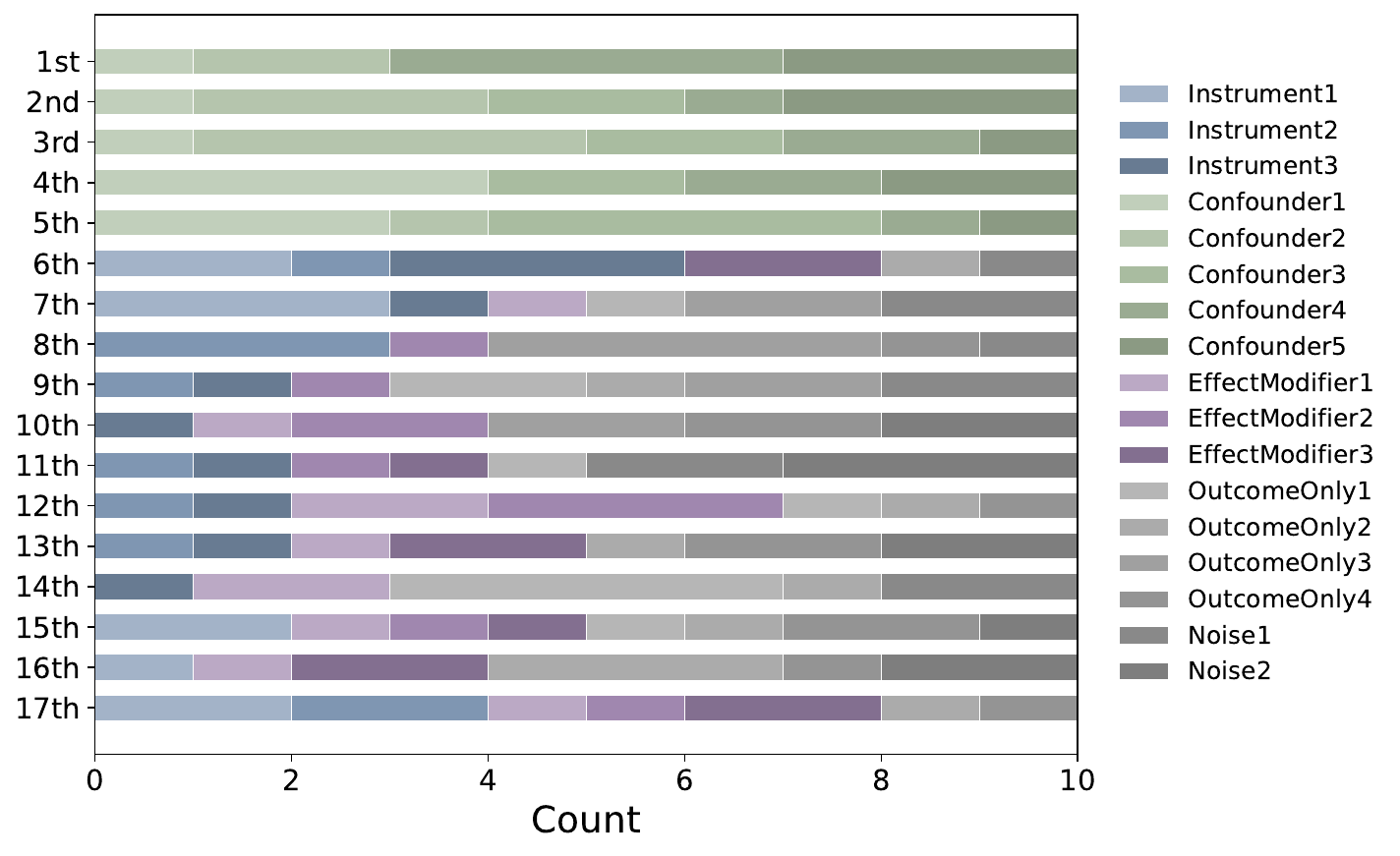}
    \caption{Rank stability}
    \label{fig:synth17_ranking}
  \end{subfigure}

  \vspace{0.05cm}
    \addtocounter{figure}{-1}
  \noindent
  \begin{tabular}{@{}p{0.48\textwidth}@{\hspace{0.04\textwidth}}p{0.48\textwidth}@{}}
    \captionof{figure}{\textbf{Low-dimensional covariate setting.} Synthetic dataset with four covariates where the confounder should be correctly recovered. $\Rightarrow$ \textit{The confounder has a large Shapley value; all others are close to zero.}}
    \label{fig:4cov_syn}
    &
    \captionof{figure}{\textbf{Medium-dimensional covariate setting.} Synthetic 17-covariate dataset with five confounders. $\Rightarrow$ \emph{The confounders have the largest Shapley values and are recovered in the top five ranks across seeds.}}
    \label{fig:synth17}
  \end{tabular}

  \vspace{-1cm}
\end{figure}

\section{Experiments}
\label{sec:experiments}

\vspace{-0.3em}
We evaluate whether \confoundingshap identifies covariates that contribute to residual confounding bias. Our experiments are designed for different objectives: \textbf{(1)}~We use synthetic data with known covariate roles to test whether our method recovers the \textbf{\confounder{true confounders}} and assigns a low confounding strength to \textbf{\iv{instruments}}, \textbf{\effmod{effect modifiers}}, and outcome predictors. \textbf{(2)}~We use the real-world ACTG experiment to test whether our method can detect that confounding strength disappears under random treatment assignment in randomized controlled trials. \textbf{(3)}~We apply our method to a real-world clinical study (i.e., the  SUPPORT right heart catheterization study \citep{connors1996}) to analyze whether \confoundingshap can identify variables that are clinically meaningful. We report additional experiments wrt. approximation quality, stability, ACIC data, and runtimes are reported in Supplement~\ref{sec:additional-experiments}. We report all implementation details in Supplement~\ref{sec:implementation-hardware}; details about datasets in Supplement~\ref{sec:datasets}; and runtimes in Supplement~\ref{sec:runtimes}.

\vspace{-0.3em}
\subsection{Synthetic data with known ground-truth confounders}
\label{sec:experiments-synthetic}

\vspace{-0.3em}
\textbf{Aim.}
Our first experiment with synthetic data is designed to benchmark whether \confoundingshap can successfully recover the ground-truth confounders. For this, we use a data-generating process (DGP) from prior work on CATE estimation \citep{curth2021}, which contains confounders but also instruments, effect modifiers, outcome predictors, as well as noise variables. We study different variants:

\vspace{-0.3em}
\textbf{(a) Low-dimensional covariate setting.}
Fig.~\ref{fig:4cov_syn} shows a simple setting with four covariates. Here, we can use the exact Shapley computation over all coalitions to show that our algorithm has the intended behavior and recovers the correct confounder. For the global attribution, the Shapley values concentrate correctly on the true confounder, while the Shapley value for the instrument, effect modifier, and outcome predictor is close to zero (Fig.~\ref{fig:4cov_syn}, left). For local attributions, the same behavior is seen where the contributions for the confounder are large and non-zero, while the contributions for the other covariates are close to zero. \textbf{$\Rightarrow$ Takeaway.}
This experiment confirms the effectiveness of \confoundingshap. \emph{Our methods assign the dominant attribution to the confounding variable as intended.}

\vspace{-0.3em}
\textbf{(b) Medium-dimensional covariate setting.}
We next evaluate a 17-covariate setting over 10 random seeds. This setting contains multiple confounders together with instruments, effect modifiers, outcome predictors, and noise variables. Here, we use the scalable algorithm from Section~\ref{sec:approximation}. We see that \confoundingshap assigns the largest importance to the true confounders while all other covariates receive close-to-zero values (Fig.~\ref{fig:synth17}, left). We further evaluate, over different runs, the rank stability of the covariates with the highest Shapley value (i.e., which our method identifies as most important confounders; Fig.~\ref{fig:synth17}, right). \textbf{$\Rightarrow$ Takeaway.} \emph{Our method is robust and, across repeated runs, consistently identifies the true confounders.}

\vspace{-0.3em}
\textbf{(c) High-dimensionality covariate setting and additional experiments.} We report additional synthetic experiments with covariate dimensions ranging from 25 to 200 in Supplement~\ref{sec:ablation}. We also confirm the rank stability when simulating multiple DGPs with different random seeds in Supplement~\ref{sec:stability-dgp}. Overall, our method is highly effective in identifying the correct confounders.

\vspace{-0.3em}
\subsection{Ground-truth confounders vs. randomized controlled trial setting.}
\label{sec:experiments-actg}

\begin{figure}[!t]
  \vspace*{-1cm}
  \centering

  \newcommand{\panelheight}{2.25cm}

  \begin{subfigure}[t]{0.23\textwidth}
    \centering
    \vspace{0pt}
    \includegraphics[width=\linewidth,height=\panelheight,keepaspectratio]{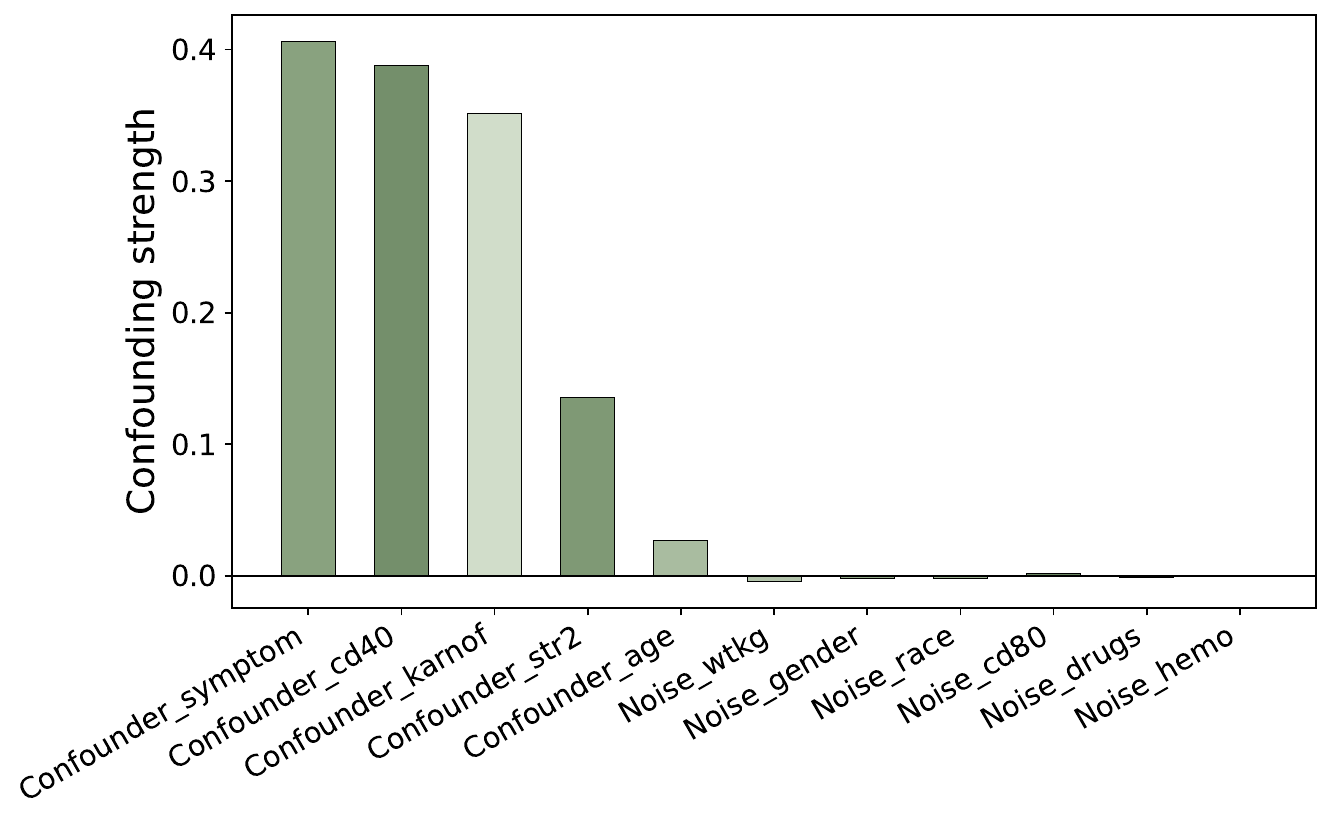}
    \caption{With confounding}
    \label{fig:actg_semisynth_importance_bar}
  \end{subfigure}
  \hfill
  \begin{subfigure}[t]{0.23\textwidth}
    \centering
    \vspace{0pt}
    \includegraphics[width=\linewidth,height=\panelheight,keepaspectratio]{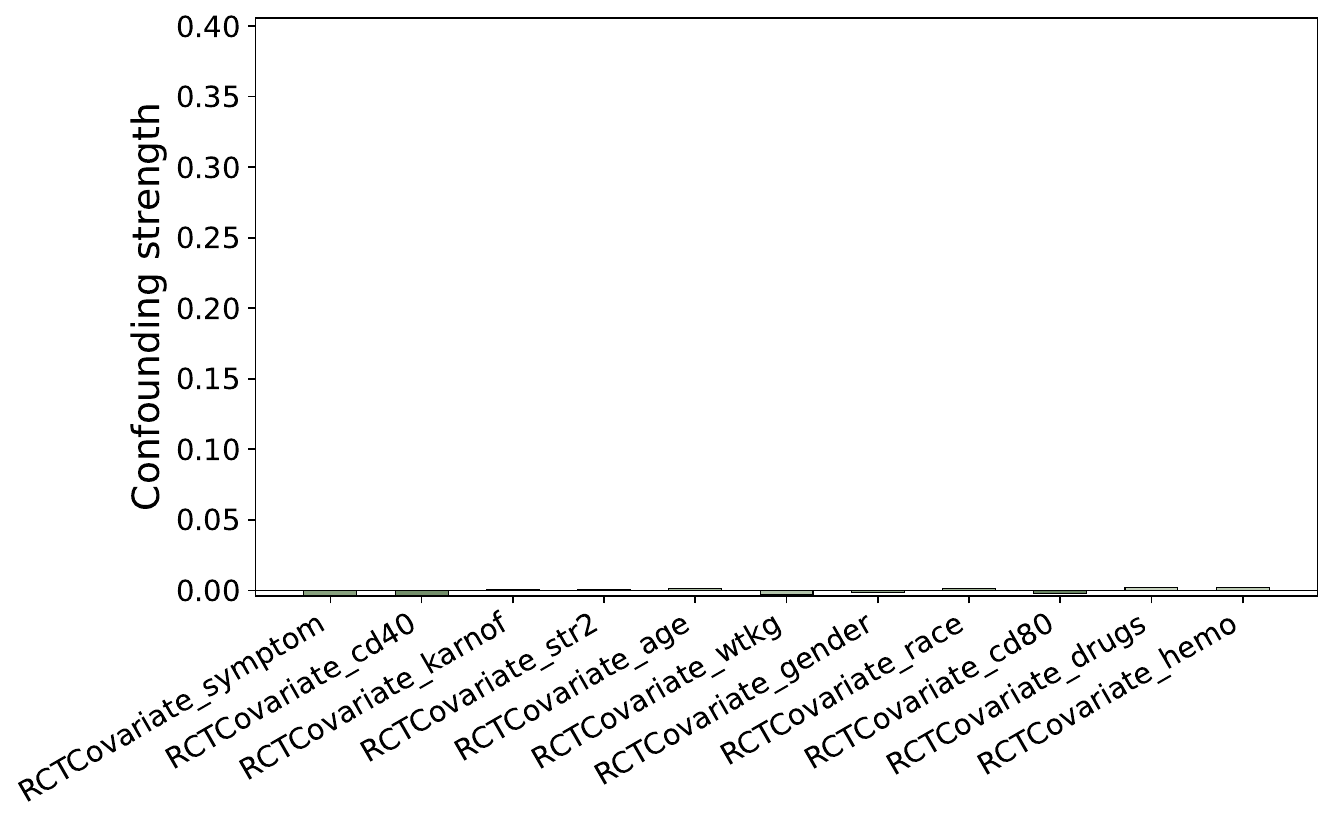}
    \caption{RCT}
    \label{fig:actg_rct_importance_bar}
    \setcounter{subfigure}{0}
  \end{subfigure}
  \hfill
  \begin{subfigure}[t]{0.23\textwidth}
    \centering
    \vspace{0pt}
    \includegraphics[width=\linewidth,height=\panelheight,keepaspectratio]{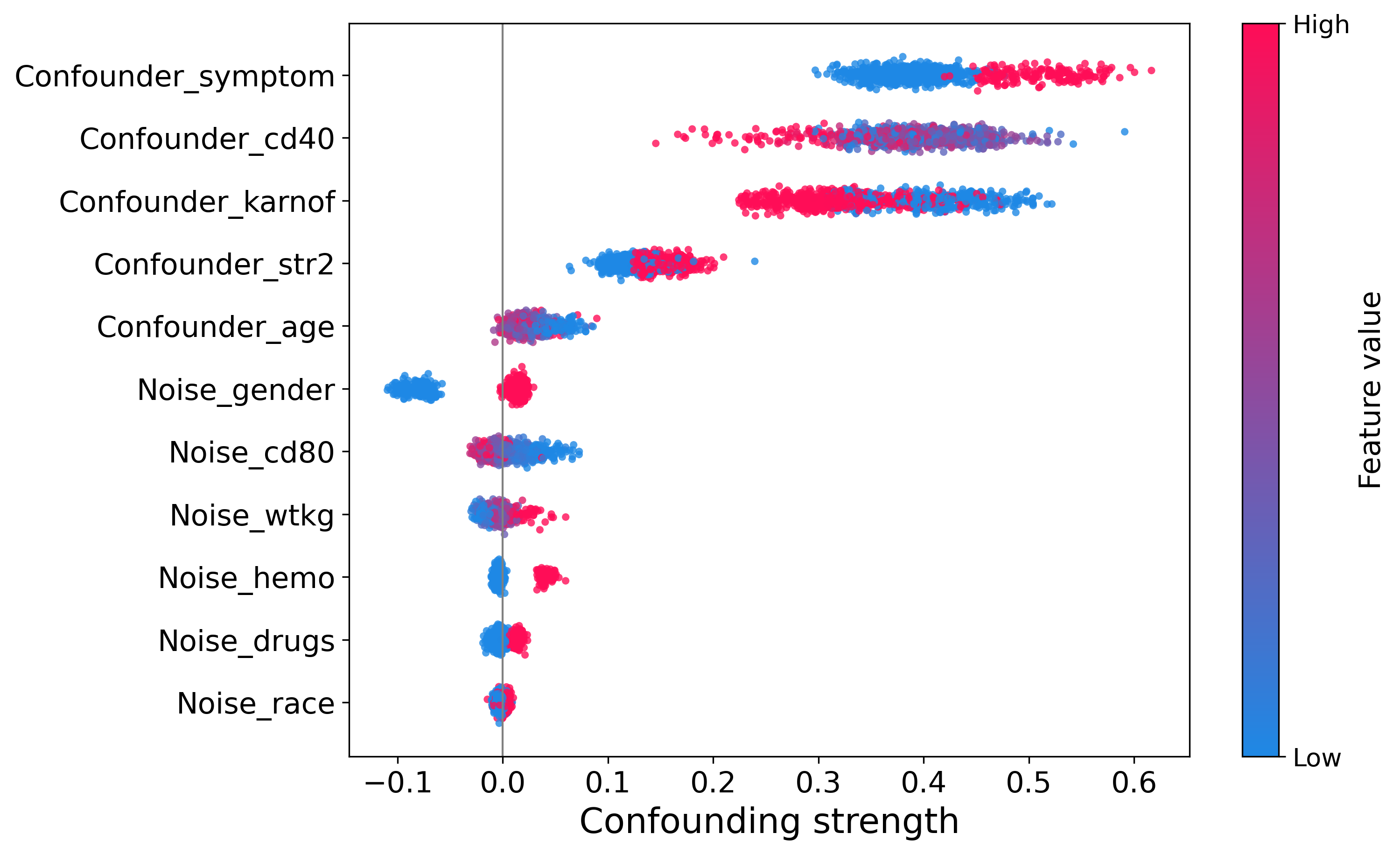}
    \caption{With confounding}
    \label{fig:actg_semisynth_beeswarm}
  \end{subfigure}
  \hfill
  \begin{subfigure}[t]{0.23\textwidth}
    \centering
    \vspace{0pt}
    \includegraphics[width=\linewidth,height=\panelheight,keepaspectratio]{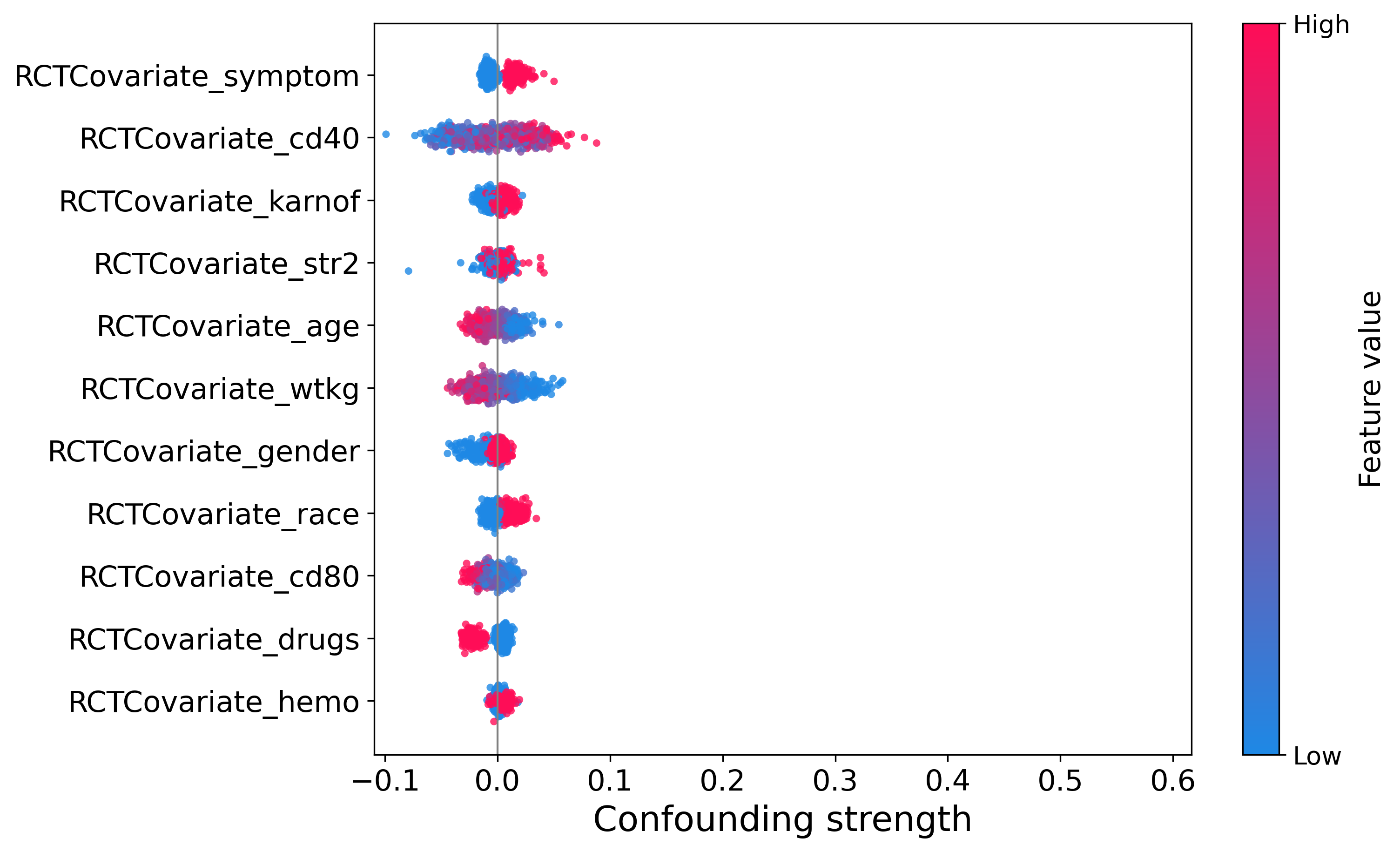}
    \caption{RCT}
    \label{fig:actg_rct_beeswarm}
  \end{subfigure}
    \addtocounter{figure}{-1}
  \vspace{0.05cm}

  \noindent
  \begin{tabular}{@{}p{0.48\textwidth}@{\hspace{0.04\textwidth}}p{0.48\textwidth}@{}}
  \vspace{-1em}
    \captionof{figure}{$\Rightarrow$ \emph{Global confounding strength disappears under randomly assigned treatment.}}
    \label{fig:actg_importance}
    &
    \vspace{-1em}
    \captionof{figure}{$\Rightarrow$ \emph{Local confounding strength disappears under randomly assigned treatment.}}
    \label{fig:actg_beeswarm}
  \end{tabular}
  \vspace{-0.8cm}
\end{figure}

\vspace{-0.5em}
\textbf{Aim.} The experiment benchmarks whether \confoundingshap successfully recovers ground-truth confounders from real-world covariates with artificially modified treatment and outcome, while assigning near-zero confounding strength to the same covariates when real-world treatment and outcome are used and treatment is randomly assigned.

\vspace{-0.3em}
\textbf{Randomly controlled trial setting.}
We use the real-world ACTG dataset to assess whether our method can detect the disappearance of confounding strength in a randomly controlled trial. We use the ACTG \citep{hammer1996} data for two experiments: (i)~we first artificially introduce confounding to the dataset by modifying treatment assignment and outcome and check whether our method can correctly identify the confounders; and (ii)~we apply our method to the original dataset including the real-world treatment and outcome, where, given the randomized nature of the treatment assignment, we should not see any confounders. Figures~\ref{fig:actg_importance} and~\ref{fig:actg_beeswarm} show that \confoundingshap successfully assigns the largest confounding strength to the confounders on the data with introduced confounding. For the same covariates under random treatment assignment, the confounding disappears with confounding strength values close-to-zero. \textbf{$\Rightarrow$ Takeaway.} \emph{Our method works as intended. \confoundingshap (i) recovers confounders when artificially introduced and (ii) recovers the disappearance of confounders when treatment is assigned at random for the same covariates.}

\vspace{-0.3cm}
\subsection{Real-world application to the SUPPORT right heart catheterization study}
\label{sec:experiments-support}
\begin{wrapfigure}{r}{0.3\linewidth}
  \centering
  \vspace{-0.7em}
  \includegraphics[width=\linewidth]{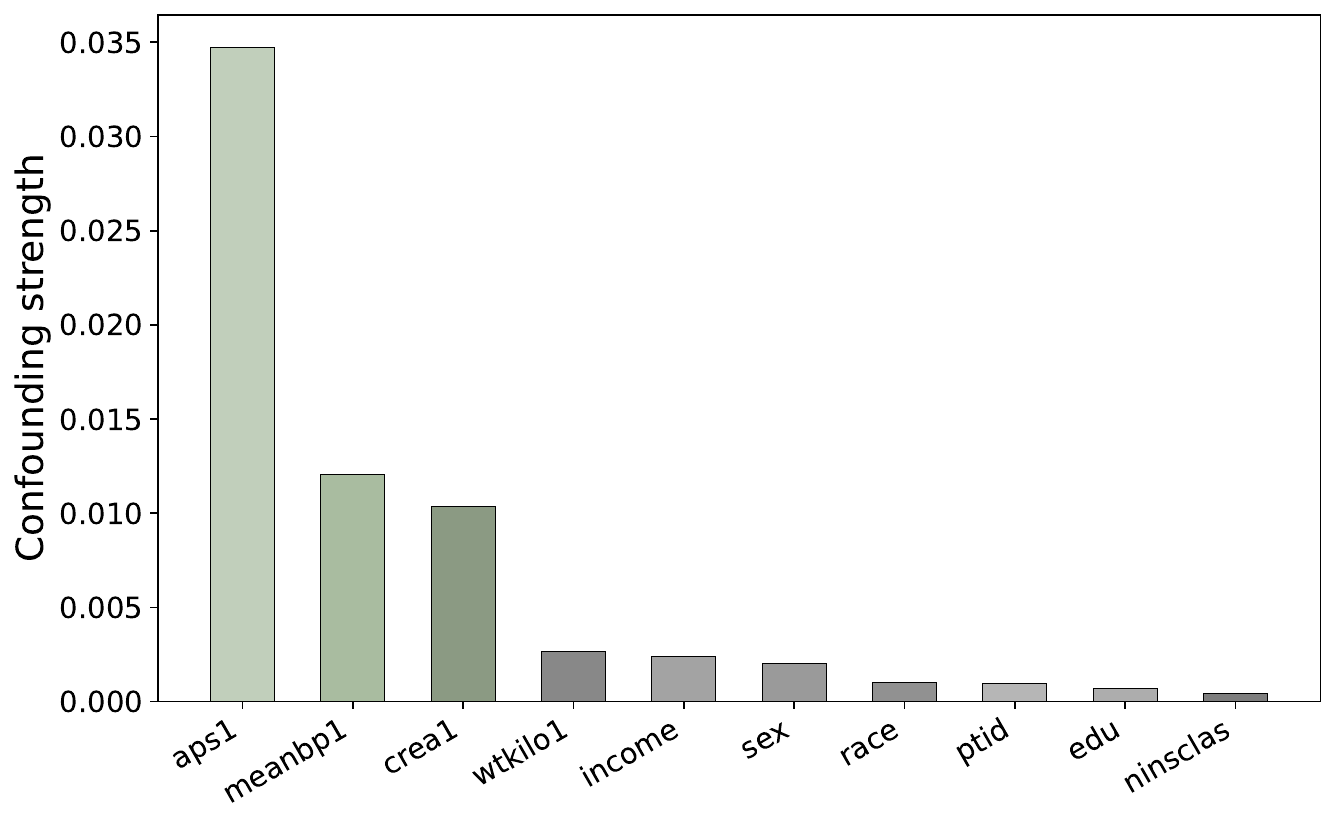}
  \vspace{-2em}
  \caption{\textbf{SUPPORT right heart catheterization study.} Confounding strength averaged over 10 runs. $\Rightarrow$ Several covariates are clinically plausible confounders.}
  \label{fig:support_realworld}
  \vspace{-1em}
\end{wrapfigure}

\vspace{-0.3em}
\textbf{Aim.} Finally, we demonstrate the clinical value of \confoundingshap by using the SUPPORT right heart catheterization study \citep{connors1996}, where the treatment is right-heart catheterization (RHC) within the first 24 hours in the intensive care unit (ICU). The outcome is 30-day mortality.

\vspace{-0.3em}
\textbf{Results.} Fig.~\ref{fig:support_realworld} shows that our method identifies several confounders that align with known confounders from clinical practice. For example, \texttt{aps1} (Acute Physiology Score) is a baseline severity-of-illness score summarizing acute physiological derangement, which influences both the likelihood of receiving RHC and the risk of mortality. We added the patient ID \texttt{ptid} as a refutation check: the patient ID is no confounder and is correctly attributed close-to-zero values. \textbf{$\Rightarrow$ Takeaway.} \emph{Our method identifies several variables that are plausible confounders in clinical practice, which confirms that our method can be highly informative.}

\textbf{Additional validations.} We provide additional validation experiments. We (i)~compare exact and estimated Shapley values in a medium-dimensional setting in Supplement~\ref{sec:additional-synthetic-results}, (ii)~study Shapley value estimation over increasing coalition budgets and covariate dimensions with different estimators in Supplement~\ref{sec:ablation}, (iii)~report stability over repeated synthetic DGP draws in Supplement~\ref{sec:stability-dgp}. (iv)~We evaluate on the \textbf{ACIC 2016 benchmark} \citep{dorie2019}: In this setting, removing the highest-ranked \confoundingshap covariates increases PEHE more than removing randomly selected or low-ranked covariates. \textbf{$\Rightarrow$ Takeaway.} \emph{Our method shows a strong and robust performance.}

\vspace{-0.5em}
\section{Discussion}
\label{sec:discussion}

\vspace{-0.5em}
\textbf{Clinical relevance.}
We follow the practical need to understand which measured covariates are confounders in observational data. These variables are clinically informative because they reveal which patient characteristics influence treatment decisions and cohort comparability, and thereby help in defining rigorous clinical questions from real-world medical data and support reliable observational trial evaluation through additional insight into relevant confounders \citep{chen2016a, feuerriegel2024, hemkens2018, hernan2016, textor2017}.

\vspace{-0.3em}
\textbf{Conclusion.} Despite the increasing use of XAI, methods for explaining causal machine learning are still scarce. We introduce a new XAI method to understand observed confounding, thereby enabling more transparent and reliable decision-making.

\section*{Acknowledgements}
This paper is supported by the DAAD program "Konrad Zuse Schools of Excellence in Artificial Intelligence", sponsored by the Federal Ministry of Education and Research.

\bibliography{literature}

\newpage
\appendix

\section{Extended related work}
\label{sec:extendedrelatedwork}

\textbf{Explainability.}
Explainability methods for machine learning provide the methodological backbone for our attribution construction. SHAP formalizes feature attributions as an additive explanation problem and assigns feature-level contributions to individual model predictions using Shapley values \citep{lundberg2017}. LIME explains individual predictions through an interpretable local surrogate model fitted around the instance of interest \citep{ribeiro2016}. SAGE is especially close in spirit because it also defines a Shapley-value game through a deliberately chosen value function \citep{covert2020}; however, its value function is based on predictive performance, whereas ours is based on a causal bias functional. Thus, our work builds on the game-theoretic attribution perspective of explainable machine learning, but changes the object being explained: rather than attributing to a prediction or predictive performance, we attribute to confounding.

\textbf{Explaining treatment assignment.}
A related line of applied work uses explainability for treatment assignment or propensity score modeling. In clinical applications, SHAP values have been used to interpret machine learning models for treatment allocation, for example, by identifying variables predictive of rescue therapy after aneurysmal subarachnoid hemorrhage and using these variables in subsequent propensity score matched analysis \citep{martini2022}. SHAP has also been proposed as a clinical face-validity check for propensity score models in frameworks for estimating patient-level treatment effects and learning treatment policies from observational data, where it can summarize the main predictors of an estimated treatment assignment mechanism \citep{gutman2025}. These workflows show that explainability can help inspect whether treatment assignment models reflect plausible clinical decision factors. Nevertheless, treatment assignment importance is not equivalent to confounding strength: an instrumental variable may strongly predict treatment while having no direct effect on the outcome, whereas confounding requires relevance to both treatment assignment and outcome.

\textbf{Explaining outcome prediction.}
In applied machine learning, SHAP is widely used to interpret outcome-prediction models. For example, SHAP has been used to explain early prediction of in-hospital mortality in sepsis, in-hospital mortality after myocardial infarction, clinical outcomes in HIV/TB co-infection, and ultrasound-radiomics XGBoost predictions of central cervical lymph-node metastasis in papillary thyroid carcinoma \citep{hu2022, shi2022, sun2025, tarabanis2023}. These studies illustrate the practical importance of interpretable prediction in medicine. However, outcome-prediction explanations answer a prognostic question: which covariates drive the predicted risk? This differs from the causal question of which covariates induce confounding bias in a treatment effect estimate. A strong prognostic variable need not be a strong confounder if it does not also affect treatment assignment.

\textbf{Treatment effect heterogeneity and CATE explanations.}
Explainability has also been studied in the context of treatment effect estimation and individualized decision making. A growing literature develops variable-importance measures for heterogeneous treatment effects. PermuCATE uses conditional permutation importance to assess variable importance for CATE estimation, while variable-importance measures for causal forests have been proposed to quantify how covariates contribute to treatment effect heterogeneity \citep{benard2025, paillard2025}. Related work develops treatment effect variable importance measures and inference procedures for local variable importance in heterogeneous treatment effects \citep{hines2025, morzywolek2025}. Several recent papers also combine SHAP or related explainability tools with CATE or individualized treatment rule modeling. For example, SHAP has been used to identify predictive biomarkers through CATE models, to support covariate ranking for effect modification in clinical drug development, and to interpret XGBoost-based individualized treatment rule models in precision medicine \citep{blumlein2022,sechidis2025a, svensson2025}. Here, the target is effect modification, individualized treatment benefit, or learned treatment rules. In contrast, our target is the bias induced by restricted adjustment, not heterogeneity of the causal effect itself.

\textbf{Causal Shapley explanations.}
Our work is also orthogonal to causal variants of Shapley explanations. Causal Shapley values incorporate causal knowledge into feature attributions for individual predictions, while asymmetric Shapley values restrict coalitions according to causal ordering information \citep{frye2020,heskes2020}. More recently, do-Shapley explanations use intervention-based reasoning to define causal attributions \citep{parafita2025, witter2026}. However, the target remains a model output, prediction, or intervention-based explanation of a causal estimand. Our contribution is different: we define the cooperative game directly on a confounding bias functional, so that the resulting Shapley values quantify contributions to confounding strength.

\textbf{Approximation of Shapley values and interactions.}
Exact Shapley value computation is generally exponential in the number of features, so a large literature studies scalable approximations. Existing methods can broadly be divided into model-specific and model-agnostic approaches. Model-specific methods exploit structural assumptions on the value function $\nu$, with tree-based games being the most prominent case and enabling polynomial- or linear-time extraction algorithms \citep{lundberg2020,nadel2026,yu2022,zern2023}. Efficient procedures have also been proposed for kernel models and $k$-nearest-neighbor predictors \citep{chau2022,wang2024}. Model-agnostic methods instead treat $\nu$ as a black box, typically using Monte Carlo estimation, regression-based estimation, or hybrid surrogate approaches \citep{castro2009,covert2020,fumagalli2026, fumagalli2026a, kolpaczki2024, lundberg2017, musco2025, strumbelj2014, wang2023} as follows: Monte Carlo methods estimate expected marginal contributions, while regression-based methods recover Shapley values as coefficients in a weighted regression problem. Recent work improves these estimators through stratified sampling, paired sampling, leverage-score designs, and more expressive polynomial regressions \citep{covert2021,fumagalli2026a,kolpaczki2024,musco2025}. Hybrid methods such as \textsc{ProxySPEX} and \textsc{RegressionMSR} learn an approximation $\hat{\nu}$ before extracting Shapley values, with \textsc{RegressionMSR} additionally accounting for the residual $\nu-\hat{\nu}$ \citep{butler2025,witter2025}. Related approximation ideas have also been extended from feature attributions to Shapley interactions, including Monte Carlo, KernelSHAP-style, tree-based, and graph-neural-network-specific methods \citep{fumagalli2023,fumagalli2024,kolpaczki2024a,muschalik2024a,muschalik2025}. 

\textbf{Unmeasured confounding and sensitivity analysis.}
Sensitivity analysis provides a complementary perspective on causal inference under unmeasured confounding. Rather than point-identifying a causal estimand under no unobserved confounding, it relaxes this assumption through sensitivity models and derives bounds for causal quantities \citep{cinelli2020, robins2000, rosenbaum1983, vanderweele2017}. A prominent line of work uses marginal sensitivity models, where hidden confounding is parameterized by bounding deviations between the observed and complete treatment assignment mechanisms, with recent extensions deriving sharp bounds and generalizing this perspective to continuous treatments and broader causal queries \citep{dorn2025, frauen2023, frauen2024, hess2026, melnychuk2023, tan2006}. Our work differs in the object being quantified. Sensitivity analysis asks how causal conclusions could change under hypothetical unmeasured confounding. We instead focus on measured covariates and attribute the bias that arises when adjustment is restricted to subsets of those covariates.

\textbf{TabPFN-based approximation.}
From a computational perspective, our work is related to prior-data fitted networks and tabular foundation models. TabPFN demonstrates that prior-data fitted networks can provide strong performance for small tabular prediction tasks by training on synthetic tabular data \citep{grinsztajn2026, hollmann2023, hollmann2025}. ExplainerPFN investigates whether meaningful Shapley-value estimates can be obtained in a zero-shot setting using a TabPFN-style model trained on synthetic structural causal models \citep{fonseca2026}. Our use of TabPFN is different: we use TabPFN-based estimation to make repeated evaluation of coalition-specific causal quantities scalable, while the Shapley game itself remains defined on confounding bias rather than predictive-model explanations.

\newpage
\section{Relations between SHAP for nuisance functions and \confoundingshap} \label{app:nuisance-shap-relation}

In the following, we briefly clarify the relation between \confoundingshap and standard SHAP explanations of the nuisance functions \(\pi(x)=\mathbb{P}(A=1\mid X=x)\) and \(\mu_a(x)=\mathbb{E}[Y\mid X=x,A=a]\), \(a\in\{0,1\}\). Although these nuisance functions enter the definition of confounding bias, their individual SHAP values are \textbf{not} sufficient to recover \confoundingshap values.

\textbf{Alternative expression for $b_S$.} Recall that, for an adjustment set \(S\subseteq[p]\), the residual confounding bias is

\vspace{-2em}
\begin{equation}
b_S(x_S)=\delta_S(x_S)-\tau_S(x_S),
\end{equation}

\vspace{-0.5em}
where \(\delta_S(x_S)\) is the observational treated-control contrast conditional on \(X_S=x_S\), and where \(\tau_S(x_S)=\mathbb{E}[\tau(X)\mid X_S=x_S]\) is the corresponding projected causal contrast. \confoundingshap then uses residual confounding bias $b_S$ to define the coalition values. These coalition values differ from standard coalition values for \(\pi\), \(\mu_0\), or \(\mu_1\), which explain treatment assignment or outcome prediction rather than residual bias. To see that, we make use of the following identity. Let

\vspace{-1em}
\begin{equation}
    e_S(x_S)=\mathbb{E}[\pi(X)\mid X_S=x_S].
\end{equation}

\vspace{-0.5em}
Then, under the full-adjustment identification assumptions, the residual bias can be written as

\vspace{-1em}
\begin{equation}
    b_S(x_S)
    =
    \frac{
    \operatorname{Cov}\!\left(\pi(X),\mu_1(X)\mid X_S=x_S\right)
    }
    {e_S(x_S)}
    +
    \frac{
    \operatorname{Cov}\!\left(\pi(X),\mu_0(X)\mid X_S=x_S\right)
    }
    {1-e_S(x_S)}.
\end{equation}

\vspace{-0.5em}
Thus, residual confounding is governed by the conditional alignment between treatment assignment and the potential-outcome regressions. This shows why \textbf{SHAP values for the nuisance functions alone cannot determine \confoundingshap}: ordinary nuisance SHAP values describe separate predictive relevance, whereas confounding depends on coalition-specific conditional covariance between treatment assignment and outcome levels. Hence, ordinary SHAP values are  neither sufficient nor necessary for \confoundingshap.

\textbf{Why not sufficient.} Consequently, large nuisance SHAP values are not sufficient for large \confoundingshap values. A variable may strongly predict treatment assignment but not outcomes, as in the case of an instrument; or it may strongly predict outcomes but not treatment assignment, as in the case of an outcome-only prognostic variable. Such variables can have large SHAP values for \(\pi\) or \(\mu_a\) while contributing little to residual confounding bias.

\textbf{Why not necessary.} Large nuisance SHAP values are also not necessary in general. For example, suppose \(X_1=U+\varepsilon_1\), \(X_2=U+\varepsilon_2\), and \(X_3=U\), with mutually independent noise terms. Let treatment depend only on \(X_1\), so that \(\pi(X)\) is a function of \(X_1\), and let the outcome regressions depend only on \(X_2\). Then, under standard model-output SHAP, \(X_3\) may receive zero attribution for all nuisance functions \(\pi,\mu_0,\mu_1\). Nevertheless, adjusting for \(X_3\) can remove the association between \(X_1\) and \(X_2\), and therefore substantially reduce residual confounding. Hence, \(X_3\) can have a large \confoundingshap value despite small or zero nuisance SHAP values. This illustrates that \confoundingshap measures adjustment value, and not direct predictive relevance.

\textbf{Absence of monotonicity.} There is therefore \textbf{no general monotonicity relation} of the form

\vspace{-1em}
\begin{equation}
    |\phi_j^{\mathrm{conf}}|
    \text{ large}
    \Rightarrow
    |\phi_j^{\pi}|,\ |\phi_j^{\mu_0}|,\ |\phi_j^{\mu_1}|
    \text{ large},
\end{equation}

\vspace{-0.5em}
nor the reverse implication. Shapley monotonicity applies when comparing marginal contributions within the same cooperative game. Here, however, nuisance SHAP and \confoundingshap are Shapley values of different games with different value functions.

\textbf{When can monotonicity hold?} Only under restrictive orthogonal-additive settings does a product-like relation emerge. For instance, if the covariates are independent, the nuisance functions are additive, and the denominators \(e_S(x_S)\) are treated as approximately constant, then the contribution of covariate \(j\) to residual confounding is proportional to a treatment-relevance term times an outcome-relevance term. Outside such special cases, correlated covariates, proxies, nonlinearities, interactions, and cancellations break any monotone relationship between nuisance SHAP values and \confoundingshap.

\textbf{Summary.} In summary, SHAP values for \(\pi,\mu_0,\mu_1\) can be useful diagnostics for understanding which variables are predictive of treatment assignment or outcomes. They are, however, neither sufficient nor necessary for \confoundingshap. \confoundingshap targets the coalition-specific residual-bias functional and therefore captures how covariates reduce treatment--outcome imbalance under adjustment.

\newpage
\section{Cancellation example} \label{app:cancel}
We give a simple population example showing that $b_\emptyset=0$ need not imply that all partially adjusted biases are zero. Let $X_1,X_2\in\{0,1\}$ be independent Bernoulli random variables with $\Pr(X_1=x_1,X_2=x_2)=1/4$ for all $(x_1,x_2)$. Let treatment be generated as
\begin{equation}
    A \mid X_1=x_1,X_2=x_2 \sim
    \mathrm{Bernoulli}(\pi_{x_1x_2}),
\end{equation}
with
\begin{equation}
\begin{array}{c c c c}
X_1 & X_2 & \pi_{x_1x_2} & m(x_1,x_2) \\ \hline
0 & 0 & 1/3  & 0 \\
0 & 1 & 3/10 & -3/2 \\
1 & 0 & 1/10 & -4/3 \\
1 & 1 & 9/10 & -7/6 .
\end{array}
\end{equation}
Let the potential outcomes be
\begin{equation}
    Y(0)=m(X_1,X_2)+\varepsilon,
    \qquad
    Y(1)=m(X_1,X_2)+\varepsilon,
    \qquad
    \varepsilon\sim\mathcal{N}(0,\sigma^2),
\end{equation}
with $\varepsilon$ independent of $(X_1,X_2,A)$ conditional on $(X_1,X_2)$.
Then, $Y$ is continuous, and the true treatment effect is zero:
\begin{equation}
    \tau(X)=Y(1)-Y(0)=0.
\end{equation}
Hence, for every subset $S\subseteq\{1,2\}$, we have
\begin{equation}
    \tau_S(x_S)=\mathbb{E}[\tau(X)\mid X_S=x_S]=0,
\end{equation}
and therefore
\begin{equation}
    b_S(x_S)=\delta_S(x_S).
\end{equation}

First, consider the empty adjustment set. The treated mean is
\begin{equation}
\mathbb{E}[Y\mid A=1]
=
\frac{
    \sum_{x_1,x_2}\Pr(X_1=x_1,X_2=x_2)\pi_{x_1x_2}m(x_1,x_2)
}{
    \sum_{x_1,x_2}\Pr(X_1=x_1,X_2=x_2)\pi_{x_1x_2}
}
=
\frac{-49/120}{49/120}
=
-1.
\end{equation}
The untreated mean is
\begin{equation}
\mathbb{E}[Y\mid A=0]
=
\frac{
    \sum_{x_1,x_2}\Pr(X_1=x_1,X_2=x_2)(1-\pi_{x_1x_2})m(x_1,x_2)
}{
    \sum_{x_1,x_2}\Pr(X_1=x_1,X_2=x_2)(1-\pi_{x_1x_2})
}
=
\frac{-71/120}{71/120}
=
-1.
\end{equation}
Thus,
\begin{equation}
    b_\emptyset
    =
    \delta_\emptyset
    =
    \mathbb{E}[Y\mid A=1]-\mathbb{E}[Y\mid A=0]
    =
    0.
\end{equation}

Now consider the adjustment for $X_1$ only. For $X_1=0$, we arrive at
\begin{equation}
\begin{aligned}
b_{\{1\}}(0)
&=
\frac{\pi_{00}m_{00}+\pi_{01}m_{01}}
     {\pi_{00}+\pi_{01}}
-
\frac{(1-\pi_{00})m_{00}+(1-\pi_{01})m_{01}}
     {(1-\pi_{00})+(1-\pi_{01})}  \\
&=
-\frac{27}{38}+\frac{63}{82}
=
\frac{45}{779}
>
0.
\end{aligned}
\end{equation}
For $X_1=1$, we yield
\begin{equation}
\begin{aligned}
b_{\{1\}}(1)
&=
\frac{\pi_{10}m_{10}+\pi_{11}m_{11}}
     {\pi_{10}+\pi_{11}}
-
\frac{(1-\pi_{10})m_{10}+(1-\pi_{11})m_{11}}
     {(1-\pi_{10})+(1-\pi_{11})}  \\
&=
-\frac{71}{60}+\frac{79}{60}
=
\frac{2}{15}
>
0.
\end{aligned}
\end{equation}
Hence, the population-level partially adjusted bias is also positive:
\begin{equation}
    \mathbb{E}[b_{\{1\}}(X_1)]
    =
    \frac{1}{2}\left(\frac{45}{779}+\frac{2}{15}\right)
    =
    \frac{2233}{23370}
    >
    0.
\end{equation}

Similarly, for adjustment for $X_2$ only, we obtain
\begin{equation}
\begin{aligned}
b_{\{2\}}(0)
&=
\frac{\pi_{00}m_{00}+\pi_{10}m_{10}}
     {\pi_{00}+\pi_{10}}
-
\frac{(1-\pi_{00})m_{00}+(1-\pi_{10})m_{10}}
     {(1-\pi_{00})+(1-\pi_{10})}  \\
&=
-\frac{4}{13}+\frac{36}{47}
=
\frac{280}{611}
>
0,
\end{aligned}
\end{equation}
and
\begin{equation}
\begin{aligned}
b_{\{2\}}(1)
&=
\frac{\pi_{01}m_{01}+\pi_{11}m_{11}}
     {\pi_{01}+\pi_{11}}
-
\frac{(1-\pi_{01})m_{01}+(1-\pi_{11})m_{11}}
     {(1-\pi_{01})+(1-\pi_{11})}  \\
&=
-\frac{5}{4}+\frac{35}{24}
=
\frac{5}{24}
>
0.
\end{aligned}
\end{equation}
Thus, we have
\begin{equation}
    \mathbb{E}[b_{\{2\}}(X_2)]
    =
    \frac{1}{2}\left(\frac{280}{611}+\frac{5}{24}\right)
    =
    \frac{9775}{29328}
    >
    0.
\end{equation}

Finally, under full adjustment, it follows that
\begin{equation}
\begin{aligned}
\delta_{\{1,2\}}(x_1,x_2)
&=
\mathbb{E}[Y\mid A=1,X_1=x_1,X_2=x_2]
-
\mathbb{E}[Y\mid A=0,X_1=x_1,X_2=x_2] \\
&=
m(x_1,x_2)-m(x_1,x_2)
=
0.
\end{aligned}
\end{equation}
Since $\tau_{\{1,2\}}(x_1,x_2)=0$, we have
\begin{equation}
    b_{\{1,2\}}(x_1,x_2)=0
    \qquad
    \text{for all } (x_1,x_2).
\end{equation}

Therefore, this data-generating process satisfies
\begin{equation}
    b_\emptyset=0,
    \qquad
    b_{\{1,2\}}(x_1,x_2)=0,
\end{equation}
but
\begin{equation}
    b_{\{1\}}(0)>0,\quad b_{\{1\}}(1)>0,
    \qquad
    b_{\{2\}}(0)>0,\quad b_{\{2\}}(1)>0.
\end{equation}
Thus, the unadjusted residual bias is zero because of cancellation in the crude comparison, even though conditioning on either single covariate reveals positive residual confounding bias.

\newpage
\section{Implementation details}
\label{sec:implementation-hardware}

\confoundingshap is implemented in Python using \texttt{numpy} and
\texttt{pandas} for data handling, \texttt{TabPFN} \citep{grinsztajn2026, hollmann2023, hollmann2025} for regression nuisance estimation, and \texttt{shapiq} \citep{muschalik2024} for exact and approximate Shapley-value computation. Experiment settings are stored in configuration files, including dataset parameters, random seeds, coalition budgets, approximation methods, and plotting options.

For each dataset, we first estimate the full-adjustment outcome model with a TabPFN-based S-learner on the augmented design matrix $(X,A)$. This yields plug-in pseudo-outcomes
\begin{equation}
\hat\tau_i = \hat\mu_1(X_i)-\hat\mu_0(X_i).
\end{equation}
For each coalition $S$, we then estimate two coalition-specific quantities: the observational treated-control contrast $\hat\delta_S$ and the projected CATE $\hat g_S$. The signed coalition value passed to the Shapley routine is
\begin{equation}
\hat\nu(S)
=
-\frac{1}{n}\sum_{i=1}^n
\left[
\hat\delta_S(X_{i,S})-\hat g_S(X_{i,S})
\right].
\end{equation}
All main experiments use this signed value function. Absolute and squared value modes are implemented as alternatives but are not used for the reported main results.

Low-dimensional synthetic experiments use exact Shapley computation by enumerating all $2^p$ coalitions. Higher-dimensional experiments use budgeted approximations through \texttt{shapiq}, primarily \textsc{RegressionMSR}, with coalition budgets specified per experiment. We use first-order Shapley values for all main attributions. Local Shapley values are computed only for exact runs, where all coalition values are available.

All nuisance regressions use \texttt{TabPFNRegressor}. To reduce the cost of repeated coalition evaluation, most synthetic and ACTG experiments use a single TabPFN estimator per regression call; the SUPPORT/RHC experiment uses the TabPFN package default of eight estimators. We disable the regression target power transform via
\texttt{REGRESSION\_Y\_PREPROCESS\_TRANSFORMS=(None,)} to avoid numerical overflow during inverse target transformation. Other TabPFN settings are left at their package defaults unless specified in the experiment configuration.

Although early TabPFN descriptions are often associated with stricter ``native'' context limits, such as roughly 100 features and 1000 samples, our implementation uses the current \texttt{TabPFNRegressor} interface. In the version used for the experiments, the built-in pretraining-limit checks allow up to 10,000 samples and 500 input columns by default. We do not bypass these limits by silently truncating rows or covariates: the global S-learner is fit on the full augmented matrix $(X,A)$, and each coalition-specific nuisance model is fit on $(X_S,A)$ or $X_S$ for the queried coalition. Thus, the largest reported fits, including SUPPORT/RHC with 5,735 observations and the high-dimensional synthetic runs with up to 200 covariates (201 columns after adding treatment), remain within the current TabPFN interface limits. To accommodate larger limits, a different model than TabPFN-2.5 is needed, such as a later version. Alternatively, one can adapt the code, such as through subsampling, feature screening/grouping, an alternative nuisance learner, or setting \texttt{ignore\_pretraining\_limits=True}; none of these overrides is used in the reported main results.

We use the \texttt{TabPFNRegressor.fit} routine for in-context learning, which does not retrain TabPFN's pretrained weights, but sets the labeled in-context reference set used for prediction. We use TabPFN as a plug-in nuisance estimator as follows: for each coalition, the corresponding design matrix defines the context, and the resulting nuisance predictions are averaged to form the empirical coalition value. A cross-fitted variant could instead build the TabPFN context on training folds and evaluate coalition values on held-out folds; this changes only the finite-sample nuisance-estimation scheme, not the \confoundingshap game itself.

All reported runs were executed on a GPU cluster. The hardware consisted of two NVIDIA H200 NVL GPUs with approximately 144GB of memory each, NVIDIA driver version 580.95.05, and CUDA version 13.0. Runtime measurements in Section~\ref{sec:runtimes} report end-to-end wall-clock times under this cluster environment, including coalition-value evaluation and Shapley reconstruction.

\newpage
\section{Ablation studies}
\label{sec:ablation}

This section evaluates the budgeted Shapley approximators used to scale \confoundingshap beyond exact enumeration. Since the synthetic benchmarks provide ground-truth covariate roles, we can directly assess whether approximation methods concentrate attribution on the true confounders. We report two metrics. First, the \emph{confounder Shapley mass} measures the fraction of absolute attribution assigned to true confounders, i.e.,
\begin{equation}
M(\hat\phi; \mathcal{C})
:=
\frac{\sum_{j \in \mathcal{C}} |\hat\phi_j|}
     {\sum_{j=1}^{p} |\hat\phi_j|}
\in [0,1],
\label{eq:confounder-mass}
\end{equation}
where $\mathcal{C} \subseteq [p]$ denotes the set of true confounders. Second, we report \emph{confounder recovery}, defined as the fraction of true confounders recovered among the top-$|\mathcal{C}|$ variables ranked by $|\hat\phi_j|$. Let
\begin{equation}
\operatorname{Top}_{|\mathcal{C}|}(\hat\phi)
\end{equation}
denote the set of $|\mathcal{C}|$ covariates with largest absolute attribution $|\hat\phi_j|$. Confounder recovery is then given by
\begin{equation}
R(\hat\phi; \mathcal{C})
:=
\frac{
\left|
\mathcal{C}
\cap
\operatorname{Top}_{|\mathcal{C}|}(\hat\phi)
\right|
}{
|\mathcal{C}|
}
\in [0,1].
\label{eq:confounder-recovery}
\end{equation}
For example, if there are ten true confounders and nine of them appear among the top ten ranked covariates, then $R=0.9$. Higher values indicate that the approximation preserves the confounder ranking induced by the exact Shapley game.

\begin{figure}[htbp]
  \centering
  \begin{subfigure}[t]{0.48\linewidth}
    \centering
    \vspace{0pt}
    \includegraphics[width=\linewidth]{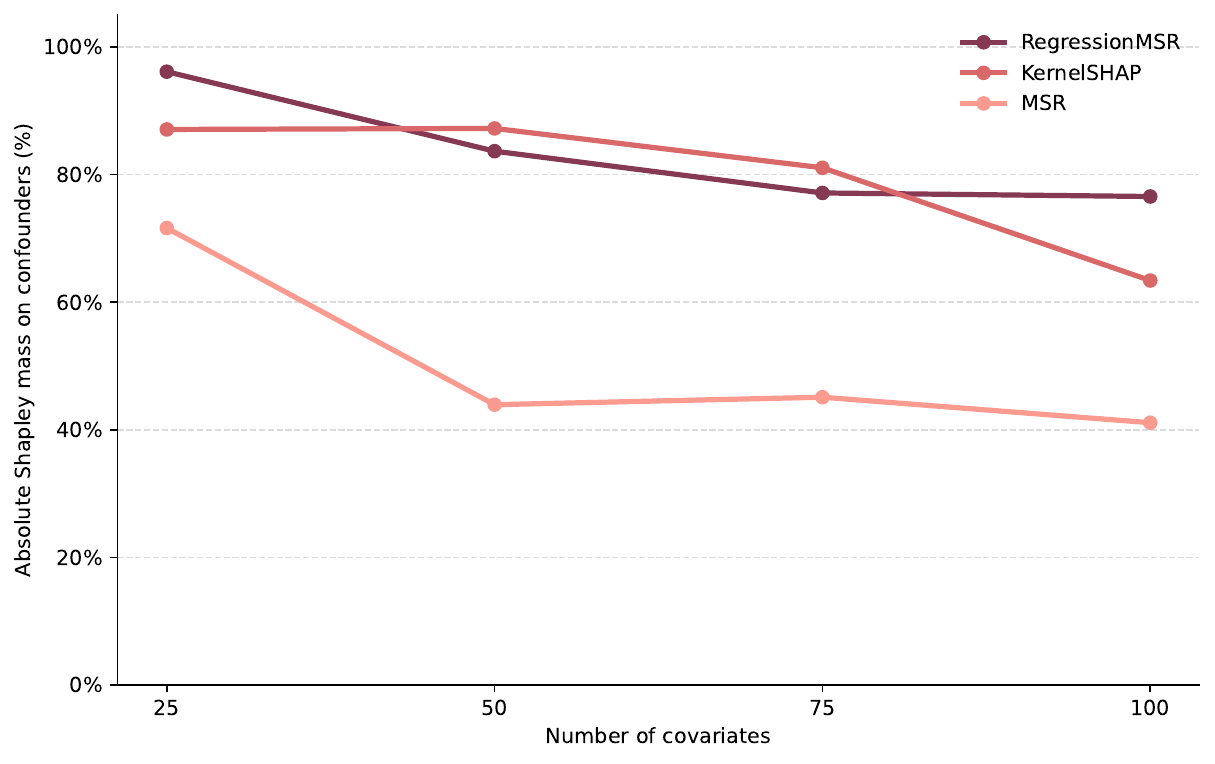}
    \caption{Confounder Shapley mass.}
    \label{fig:shapleymass_budget1024}
  \end{subfigure}
  \hspace{0.02\linewidth}
  \begin{subfigure}[t]{0.48\linewidth}
    \centering
    \vspace{0pt}
    \includegraphics[width=\linewidth]{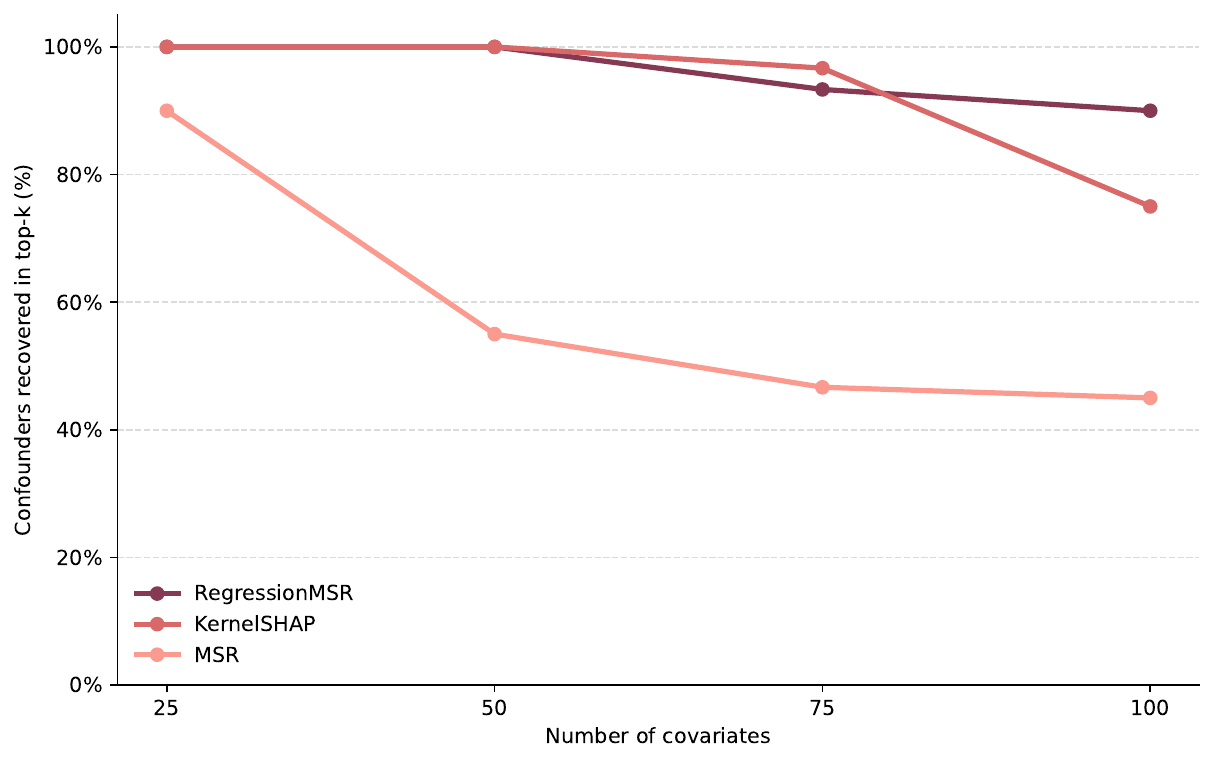}
    \caption{Confounder recovery.}
    \label{fig:confounderrecovery_budget1024}
  \end{subfigure}
  \vspace{-0.2cm}
  \caption{Approximation quality at fixed budget $B=1024$ across increasing numbers of covariates.}
  \label{fig:ablation_budget1024_dimensionality}
\end{figure}

All three approximators recover the exact Shapley values in the limit $B \to 2^p$, but they differ substantially at finite coalition budgets. At a fixed budget of $B=1024$, Fig.~\ref{fig:ablation_budget1024_dimensionality} shows that \textsc{RegressionMSR} and \textsc{KernelSHAP} assign substantially more attribution mass to true confounders than \textsc{MSR}, and the difference becomes more pronounced as the number of covariates increases. The same pattern is observed for confounder recovery: proxy- and kernel-based approximations preserve the confounder ranking more reliably at matched budget.

\vspace{0.5cm}

\begin{figure}[htbp]
  \centering

  \begin{subfigure}[t]{0.4\linewidth}
    \centering
    \includegraphics[width=\linewidth]{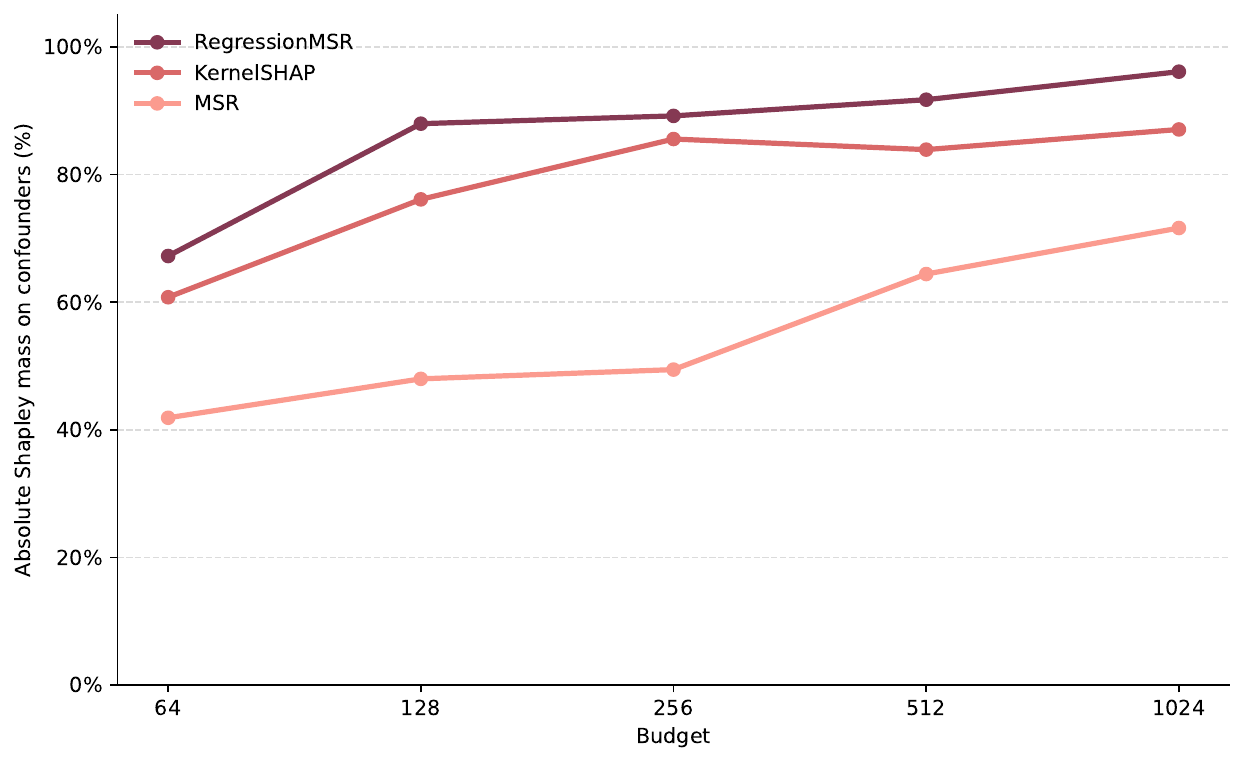}
    \caption{25 covariates}
    \label{fig:shapleymass_25}
  \end{subfigure}
  \hfill
  \begin{subfigure}[t]{0.4\linewidth}
    \centering
    \includegraphics[width=\linewidth]{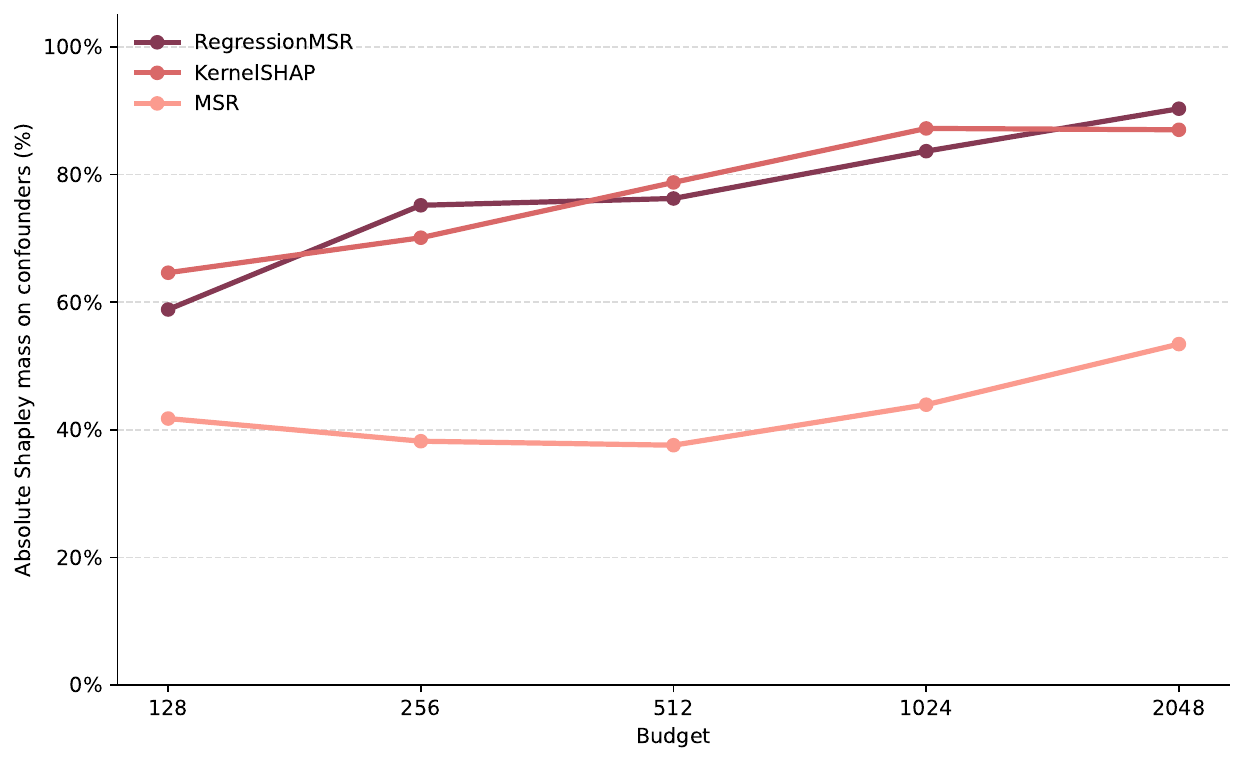}
    \caption{50 covariates}
    \label{fig:shapleymass_50}
  \end{subfigure}

  \vspace{0.2cm}

  \begin{subfigure}[t]{0.4\linewidth}
    \centering
    \includegraphics[width=\linewidth]{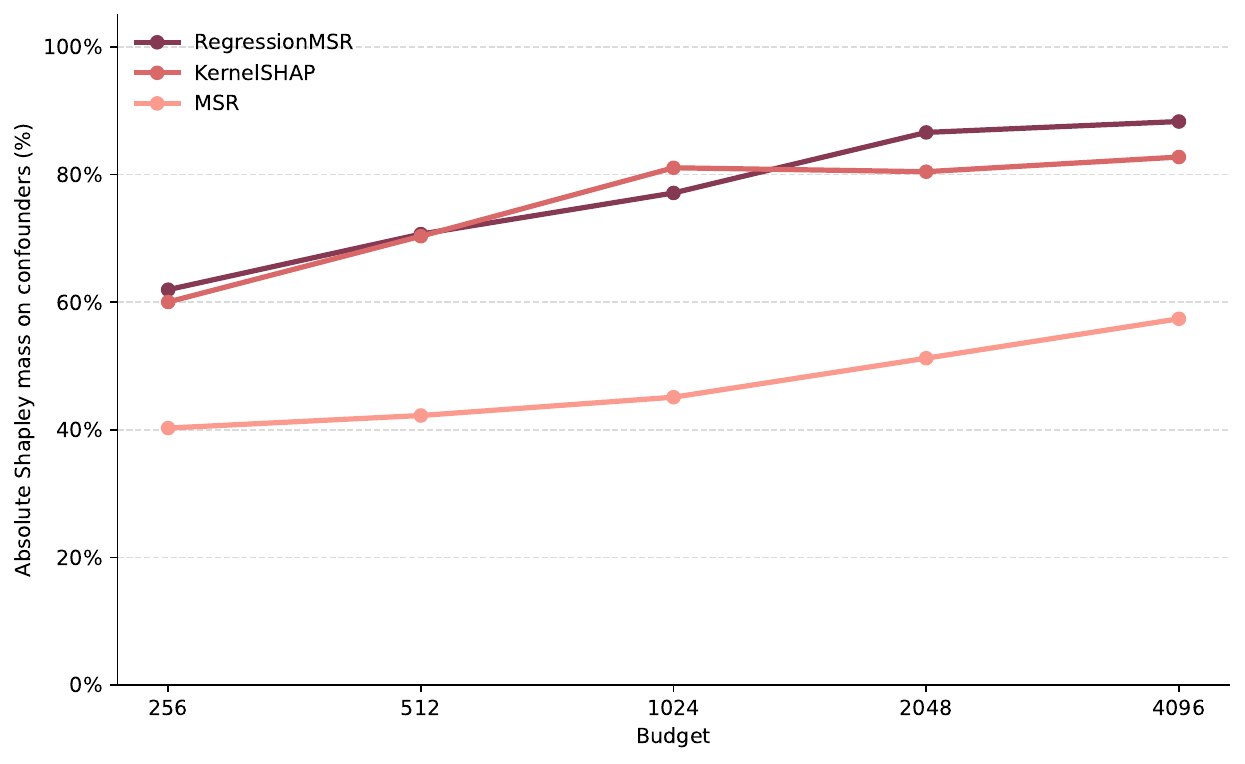}
    \caption{75 covariates}
    \label{fig:shapleymass_75}
  \end{subfigure}
  \hfill
  \begin{subfigure}[t]{0.4\linewidth}
    \centering
    \includegraphics[width=\linewidth]{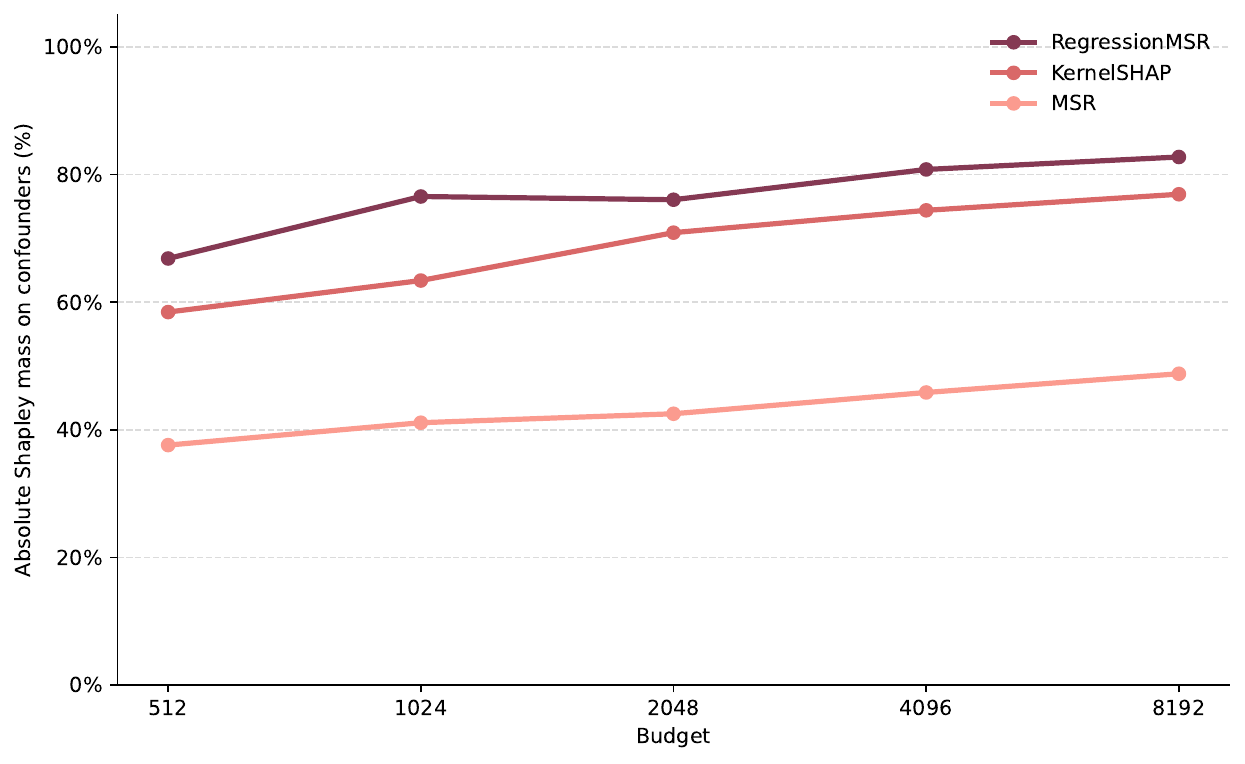}
    \caption{100 covariates}
    \label{fig:shapleymass_100}
  \end{subfigure}

  \caption{Absolute Shapley mass assigned to true confounders as a function of coalition budget. Each setting contains 40\% confounders.}
  \label{fig:absolute_shapley_mass}
\end{figure}

\begin{figure}[htbp]
  \centering

  \begin{subfigure}[t]{0.4\linewidth}
    \centering
    \includegraphics[width=\linewidth]{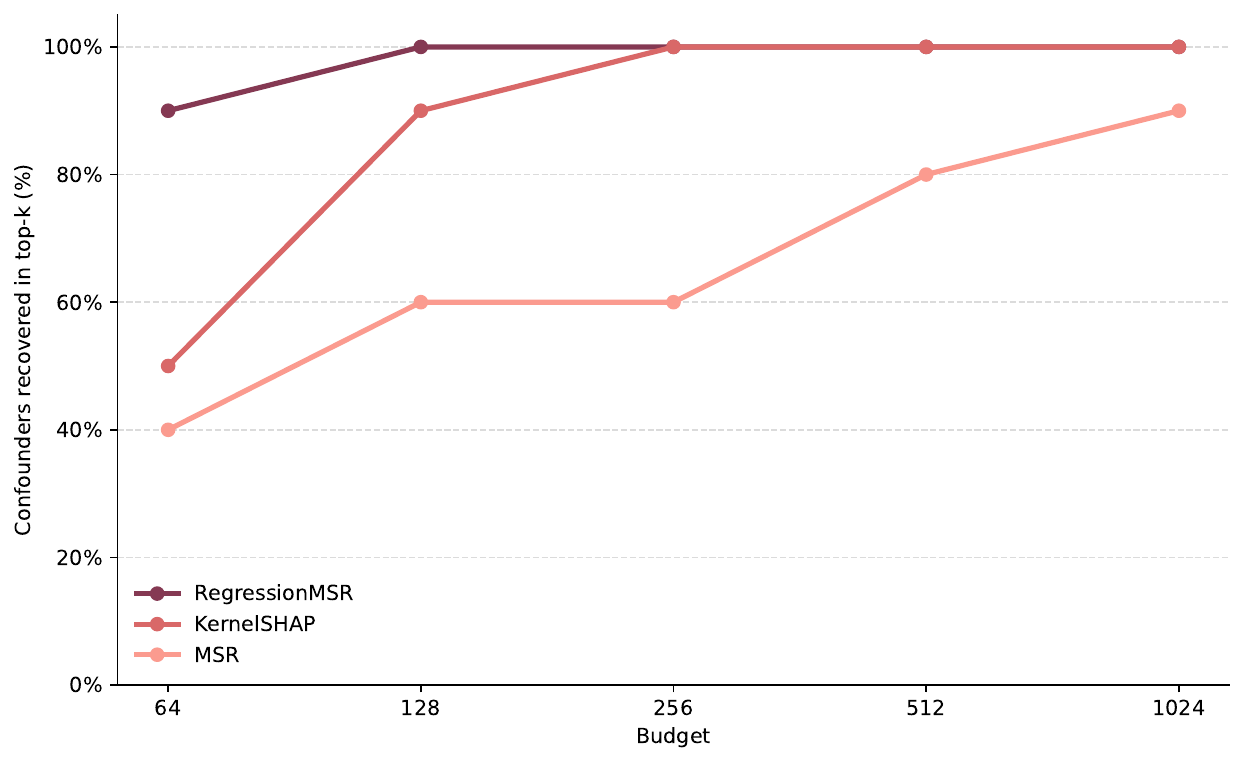}
    \caption{25 covariates}
    \label{fig:confounderrecovery_25}
  \end{subfigure}
  \hfill
  \begin{subfigure}[t]{0.4\linewidth}
    \centering
    \includegraphics[width=\linewidth]{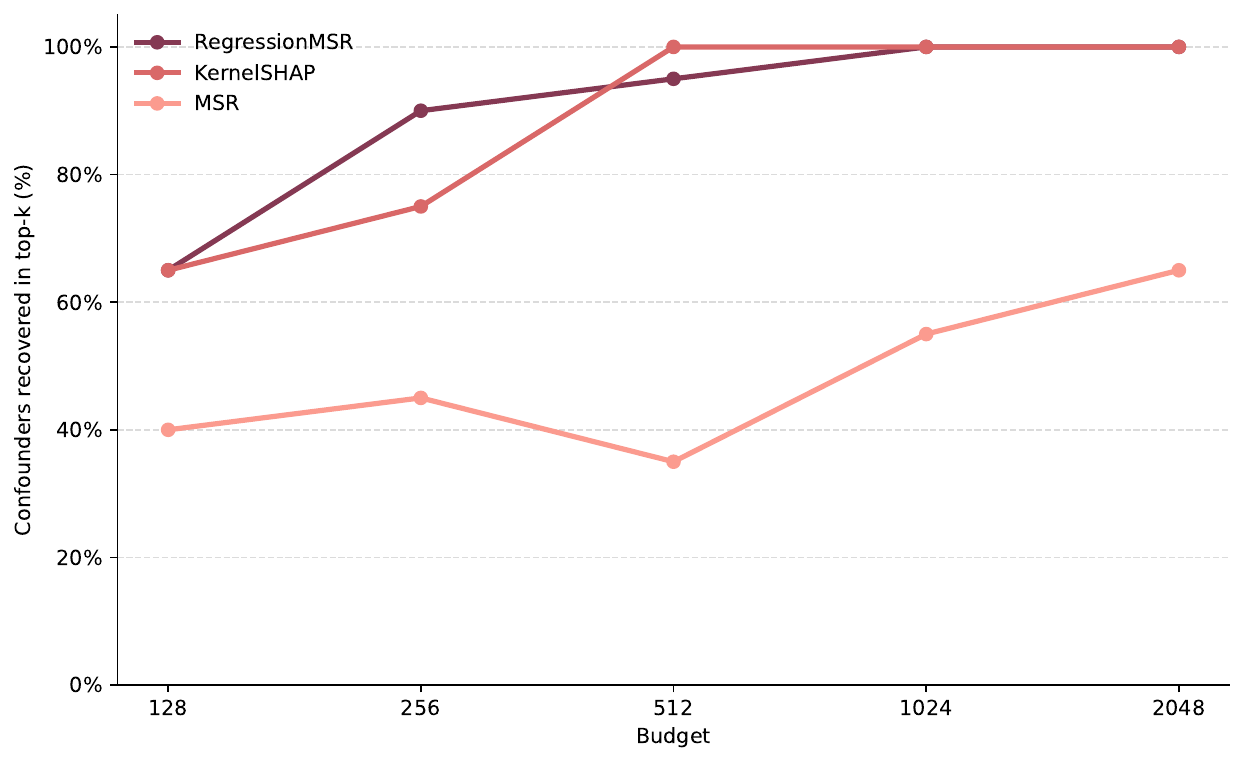}
    \caption{50 covariates}
    \label{fig:confounderrecovery_50}
  \end{subfigure}

  \vspace{0.2cm}

  \begin{subfigure}[t]{0.4\linewidth}
    \centering
    \includegraphics[width=\linewidth]{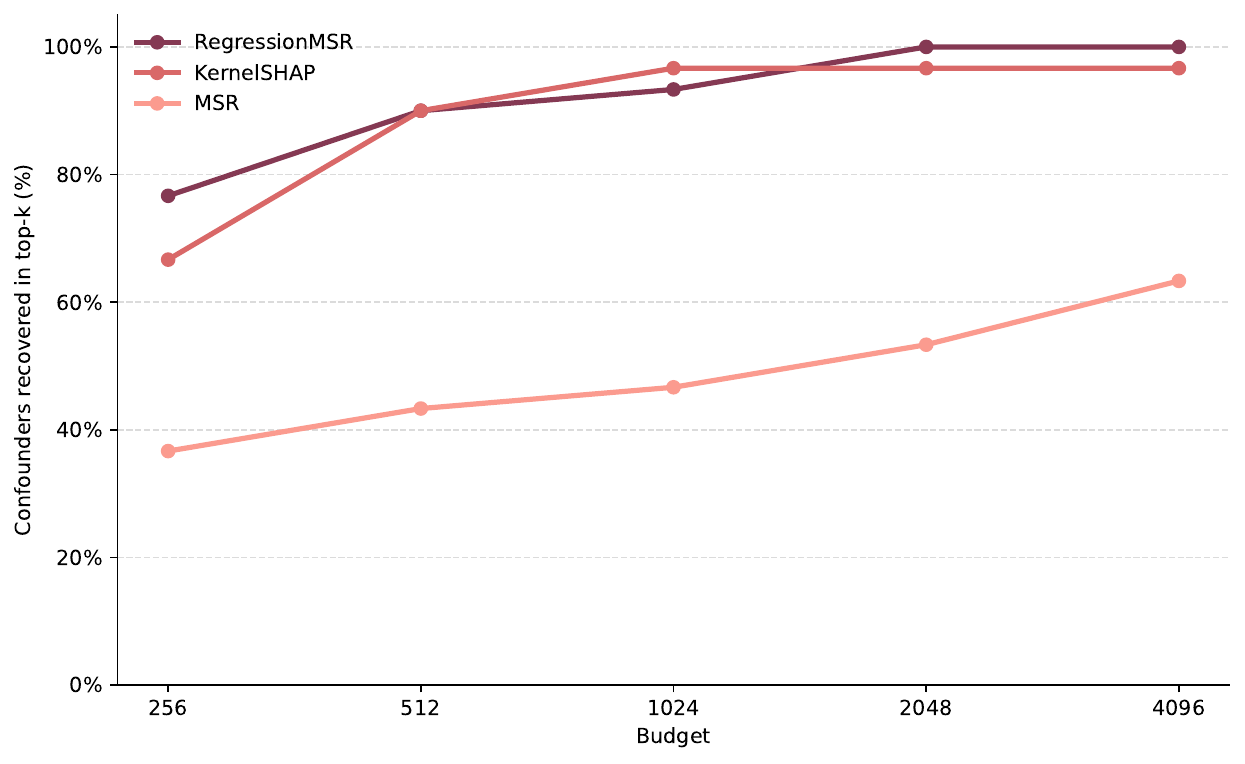}
    \caption{75 covariates}
    \label{fig:confounderrecovery_75}
  \end{subfigure}
  \hfill
  \begin{subfigure}[t]{0.4\linewidth}
    \centering
    \includegraphics[width=\linewidth]{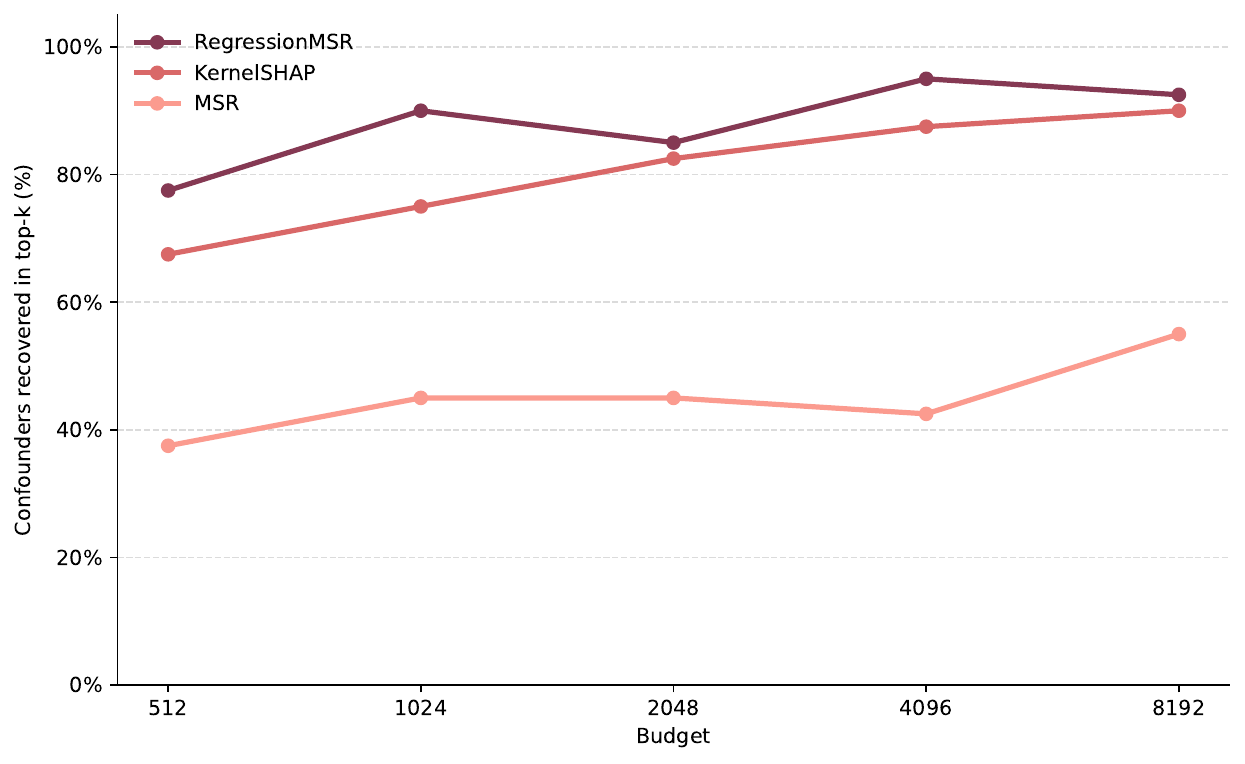}
    \caption{100 covariates}
    \label{fig:confounderrecovery_100}
  \end{subfigure}

  \caption{Fraction of true confounders recovered among the top 40\% of
  covariates as a function of coalition budget.}
  \label{fig:confounder_recovery_budget}
\end{figure}

We next vary the coalition budget for synthetic problems across 25, 50, 75, and 100 covariates, each containing 40\% true confounders. Figures~\ref{fig:absolute_shapley_mass} and~\ref{fig:confounder_recovery_budget} show that increasing the budget improves both attribution mass and confounder recovery. This confirms the expected accuracy-compute tradeoff: larger budgets provide more information about the set function and yield more reliable rankings, especially in higher-dimensional settings.

\vspace{0.5cm}

\vspace{0.5cm}

\begin{figure}[htbp]
  \centering
  \begin{subfigure}[t]{0.48\linewidth}
    \centering
    \vspace{0pt}
    \includegraphics[width=\linewidth]{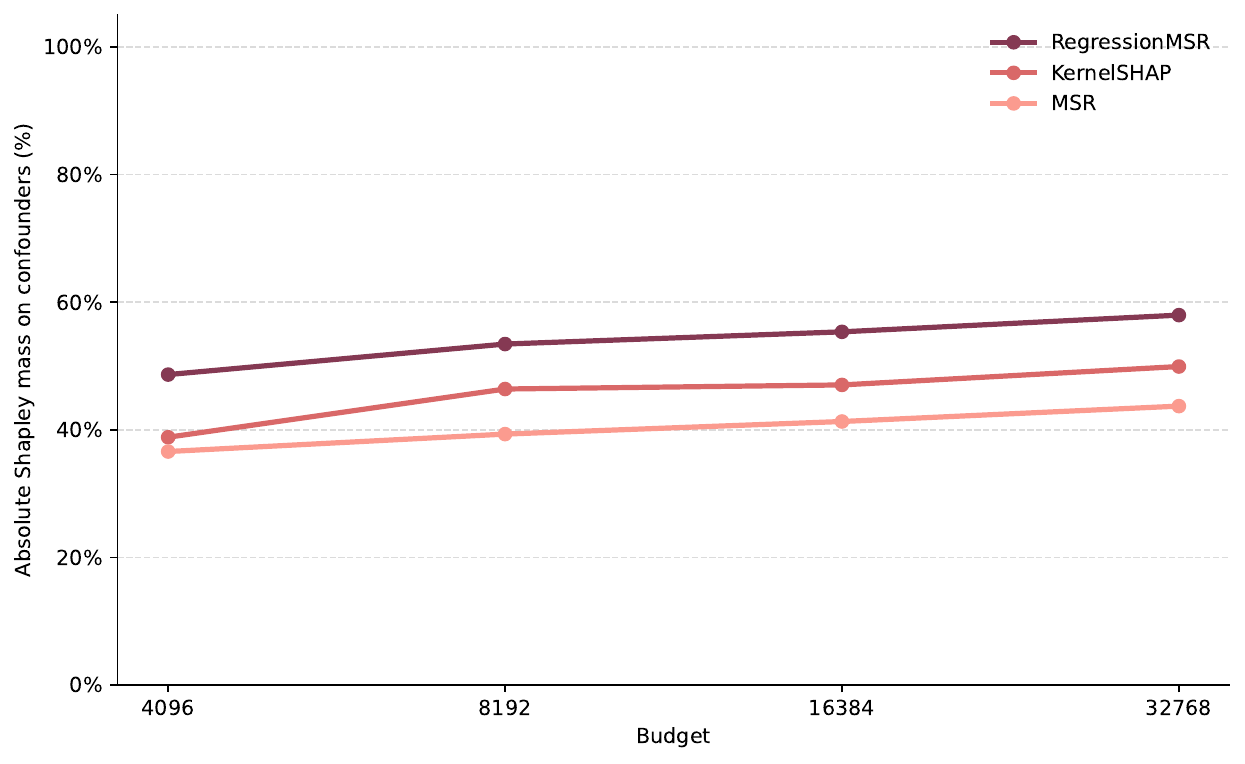}
    \caption{Confounder Shapley mass.}
    \label{fig:shapleymass_200}
  \end{subfigure}
  \hspace{0.02\linewidth}
  \begin{subfigure}[t]{0.48\linewidth}
    \centering
    \vspace{0pt}
    \includegraphics[width=\linewidth]{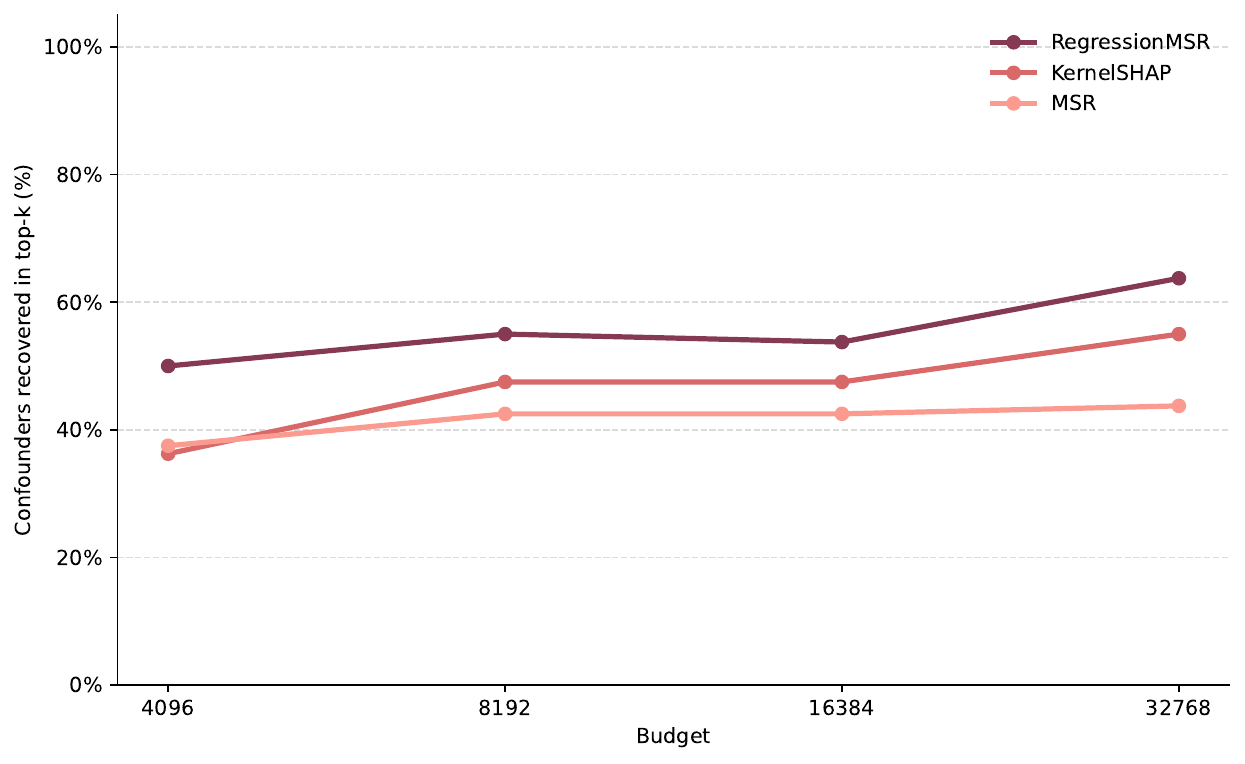}
    \caption{Confounder recovery.}
    \label{fig:confounderrecovery_200}
  \end{subfigure}
  \vspace{-0.2cm}
  \caption{Approximation quality for 200 covariates across increasing budgets.}
  \label{fig:ablation_200covariates}
\end{figure}

Finally, we consider a 200-covariate setting. Fig.~\ref{fig:ablation_200covariates} shows that approximation quality continues to improve with budget, but the absolute values remain lower than in smaller problems. This highlights the main computational limitation of the current implementation: very high-dimensional settings require large coalition budgets to obtain stable rankings, and these budgets can lead to long runtimes; see Section~\ref{sec:runtimes}.

\vspace{0.5cm}

\clearpage
\newpage
\section{Additional experiments}
\label{sec:additional-experiments}

\subsection{Confounders can cancel when explaining a CATE estimator}
\label{sec:cancellingconfounder}

We construct a synthetic observational dataset with \(n=5000\) independent observations and four covariates with distinct causal roles: an instrument \(Z\), a confounder \(C\), an effect modifier \(M\), and an outcome-only prognostic variable \(O\). All covariates are sampled independently as
\begin{equation}
Z_i, C_i, M_i, O_i \sim \mathcal{N}(0,1).
\end{equation}
Treatment assignment is observational and follows
\begin{equation}
\Pr(A_i=1 \mid X_i)
=
\operatorname{logit}^{-1}(3C_i+Z_i),
\end{equation}
so both the confounder \(C_i\) and the instrument \(Z_i\) affect treatment uptake.

The potential outcome means share the same prognostic component
\begin{equation}
h(C_i,O_i)=2C_i+O_i+0.5O_i^2.
\end{equation}
The control mean is
\begin{equation}
\mu_0(X_i)=h(C_i,O_i),
\end{equation}
and the treated mean is
\begin{equation}
\mu_1(X_i)=h(C_i,O_i)+\tau(X_i),
\end{equation}
where the treatment effect is
\begin{equation}
\tau(X_i)=1+3M_i^2.
\end{equation}
Observed potential outcomes are generated as
\begin{equation}
Y_i(a)=\mu_a(X_i)+\epsilon_i,
\qquad
\epsilon_i \sim \mathcal{N}(0,0.5^2),
\end{equation}
and the observed outcome is
\begin{equation}
Y_i=(1-A_i)Y_i(0)+A_iY_i(1).
\end{equation}

This construction makes \(C_i\) a genuine confounder: it affects treatment assignment, and it also affects the outcome level through the shared prognostic term \(h(C_i,O_i)\). However, \(C_i\) does not modify the treatment effect, because it enters the treated and untreated potential outcome means identically. Consequently, in the conditional treatment contrast, the shared prognostic component cancels out:
\begin{equation}
\mu_1(X_i)-\mu_0(X_i)
=
\{h(C_i,O_i)+\tau(X_i)\}-h(C_i,O_i)
=
\tau(X_i)
=
1+3M_i^2.
\end{equation}
Thus, the true CATE depends only on the effect modifier \(M_i\), not on the confounder \(C_i\). The confounder is therefore essential for adjustment in the observational treatment-outcome relationship, but it is not a source of treatment effect heterogeneity.

For the CATE baseline, we fit a doubly robust learner \citep{kennedy2023} using \texttt{causalml.inference.meta.BaseDRRegressor} \citep{chen2020}. The propensity model, treated and control outcome models, and final treatment effect model are implemented with gradient-boosted trees from \texttt{xgboost} \citep{chen2016}. We then apply SHAP to the fitted CATE predictions \(\hat\tau(x)\), so the resulting feature importance reflects variation in the estimated treatment effect function. This is the natural target for explaining CATE heterogeneity, but it is not the same estimand as attributing confounding bias.

The experiment is designed so that CATE-based SHAP and \confoundingshap answer different questions. Since the true treatment effect is
\(\tau(X_i)=1+3M_i^2\), SHAP on a CATE function should primarily attribute variation to the effect modifier \(M_i\). The confounder \(C_i\) can have little or no CATE-based SHAP importance because its direct outcome contribution is shared across both potential outcomes and cancels in the treatment contrast. In contrast, \confoundingshap targets confounding in the observational treatment contrast. Since \(C_i\) drives both treatment assignment and the outcome level, it receives the dominant \confoundingshap attribution. This illustrates the central distinction: CATE-based SHAP explains heterogeneity in the fitted treatment effect, whereas \confoundingshap attributes confounding bias induced by variables that jointly affect treatment assignment and outcomes.

\begin{figure}[htbp]
  \centering
  \begin{subfigure}[t]{0.48\linewidth}
    \centering
    \vspace{0pt}
    \includegraphics[width=\linewidth]{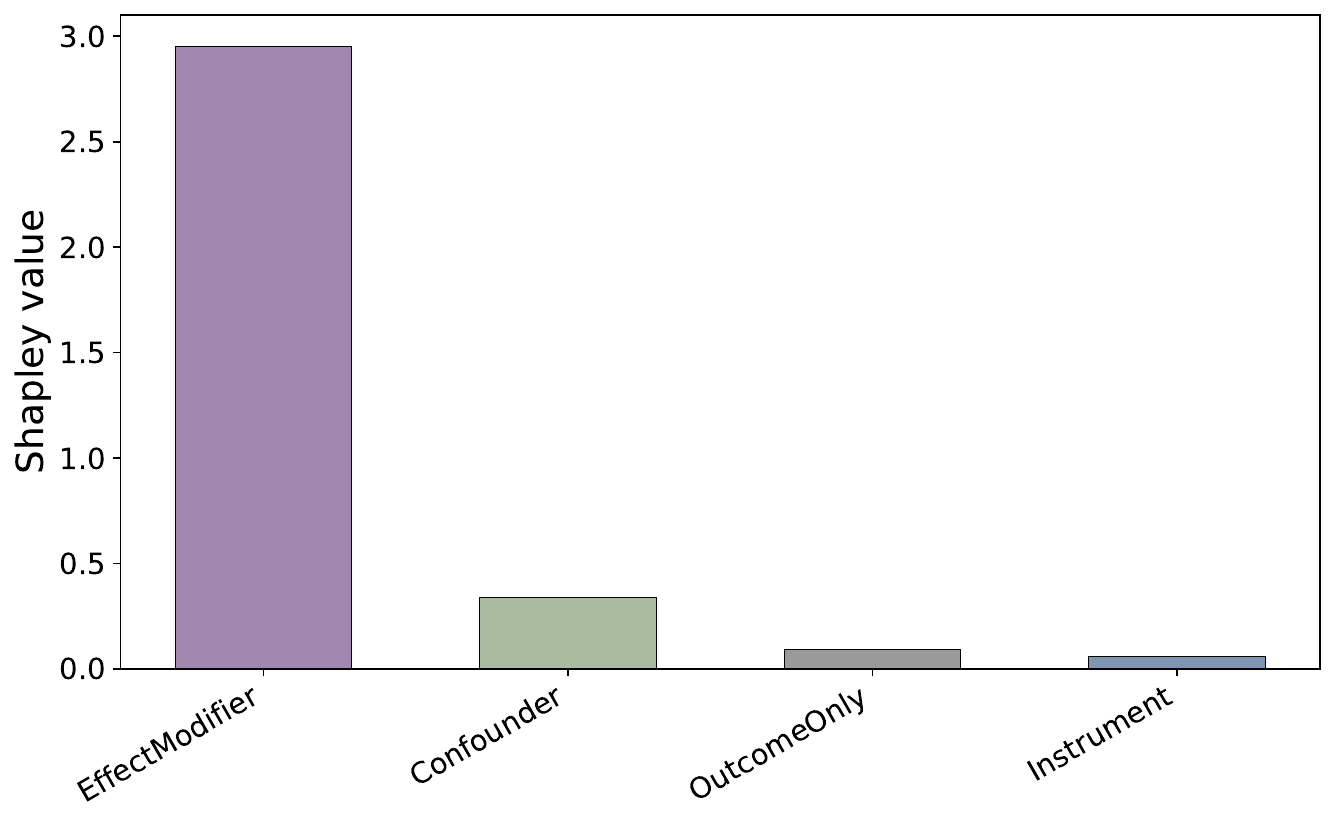}
    \caption{SHAP on a DR-learner}
    \label{fig:drlearner}
  \end{subfigure}
  \hspace{0.02\linewidth}
  \begin{subfigure}[t]{0.48\linewidth}
    \centering
    \vspace{0pt}
    \includegraphics[width=\linewidth]{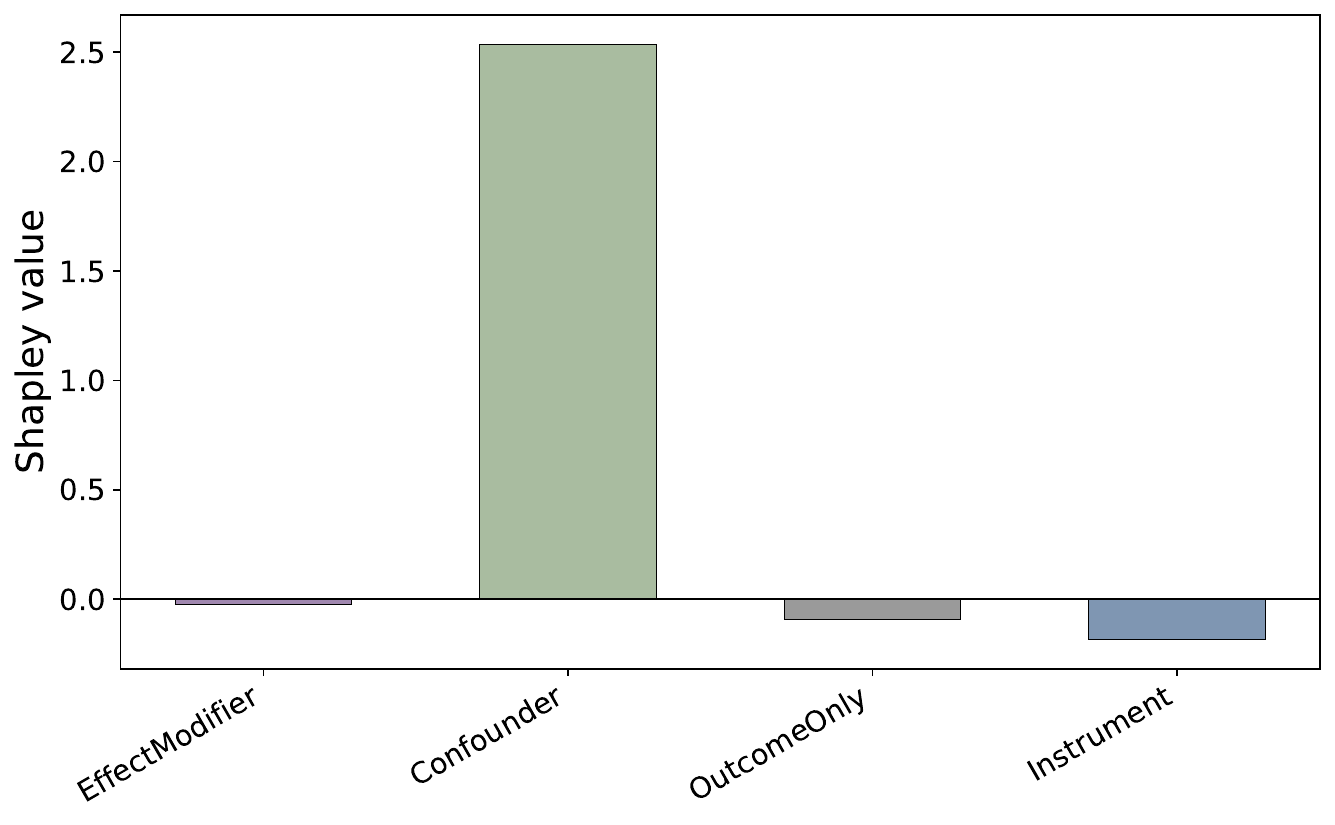}
    \caption{\confoundingshap}
    \label{fig:11cov_proxy_confounding}
  \end{subfigure}
  \vspace{-0.2cm}
  \caption{CATE explanation versus confounding bias attribution on a synthetic dataset with 4 covariates.}
  \label{fig:11cov_drlearner_vs_confoundingshap}
\end{figure}
\vspace{0.5cm}

\subsection{ACIC 2016 data challenge}
\label{sec:acic-pehe}

We also evaluate whether the ranking induced by \confoundingshap is meaningful for downstream CATE estimation. For this purpose, we use the ACIC 2016 data challenge benchmark \citep{dorie2019}, where covariates are real but treatment assignments and potential outcomes are simulated, making ground-truth treatment effects available. This allows us to assess whether covariates ranked highly by \confoundingshap are important for estimating individual treatment effects.

We rank covariates by their absolute \confoundingshap values and compare three feature-removal strategies. First, we remove the top-$k$ ranked covariates. Second, we remove $k$ randomly selected covariates. Third, we remove the lowest-ranked $k$ covariates. After removal, we refit the same CATE learner used in Section~\ref{sec:cancellingconfounder} and evaluate the precision in estimation of heterogeneous effects (PEHE),
\begin{equation}
\operatorname{PEHE}
=
\sqrt{
\frac{1}{n}
\sum_{i=1}^{n}
\left(
\hat{\tau}(x_i) - \tau(x_i)
\right)^2
},
\label{eq:pehe}
\end{equation}
where $\hat{\tau}(x_i)$ is the estimated CATE and $\tau(x_i)$ is the known ground-truth treatment effect in the ACIC benchmark.

\begin{figure}[htbp]
  \centering
  \includegraphics[width=0.7\linewidth]{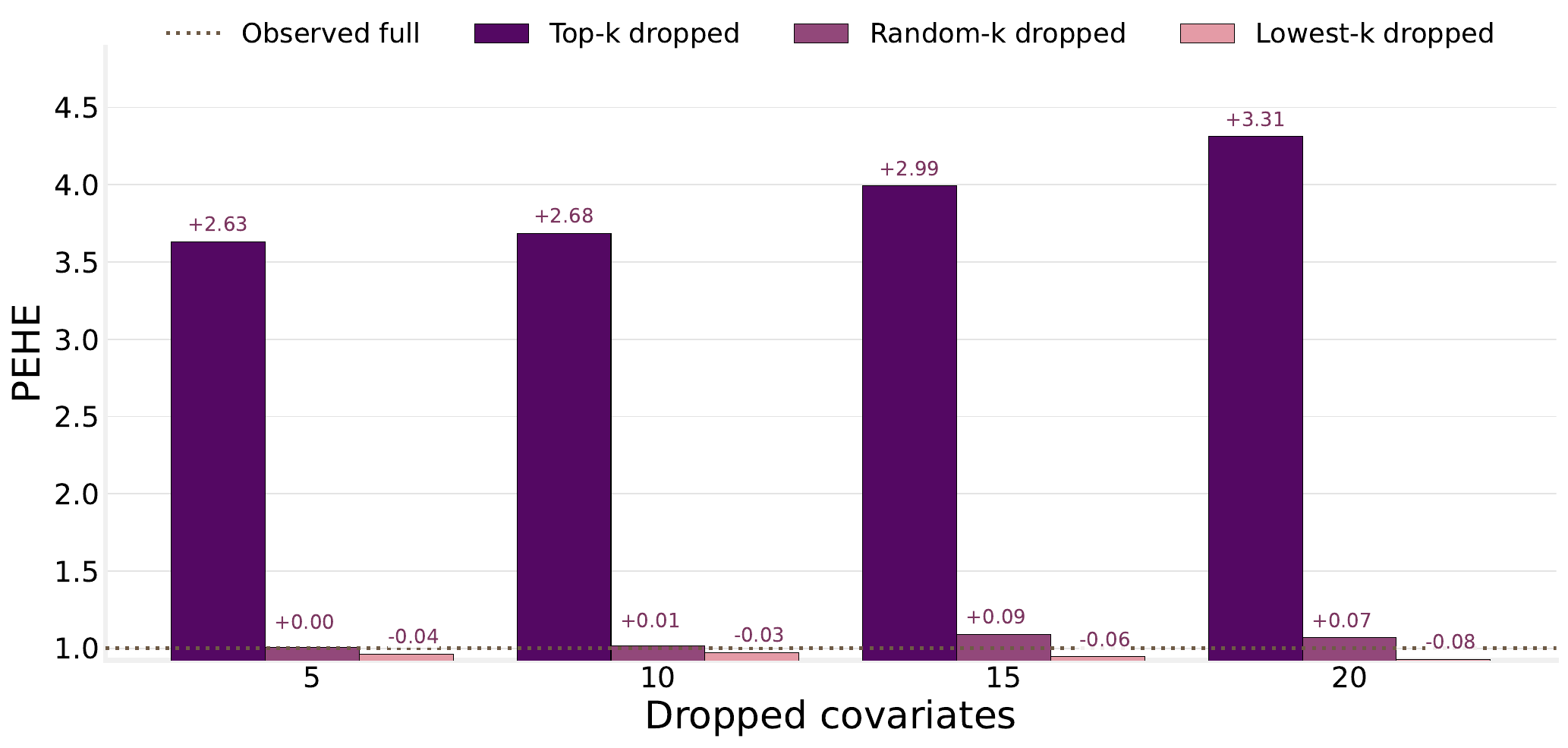}
  \caption{PEHE after removing top-ranked, random, or lowest-ranked covariates
  on the ACIC 2016 benchmark.}
  \label{fig:pehe}
\end{figure}

If \confoundingshap correctly identifies covariates that are important for adjustment, then removing the top-ranked variables should make CATE estimation substantially harder and increase PEHE. In contrast, removing the lowest-ranked variables should have little effect, while random removal should lie between these two extremes because it sometimes removes important adjustment variables and sometimes removes irrelevant ones. Fig.~\ref{fig:pehe} shows exactly this pattern: dropping the top-ranked \confoundingshap covariates leads to a much larger PEHE increase than dropping randomly selected or lowest-ranked covariates.

This experiment should be interpreted as a downstream validation rather than a standalone benchmark of CATE learners. The magnitude of the PEHE increase also depends on the learner used after feature removal, since different CATE estimators may be more or less robust to omitted adjustment variables. Here, the purpose is to test whether the \confoundingshap ranking identifies variables whose removal harms downstream treatment effect estimation under a fixed learner.

\vspace{0.5cm}

We use the same DR-learner setup as in Section~\ref{sec:cancellingconfounder}, with \texttt{xgboost} models for the propensity, outcome, and final treatment effect stages. Propensity scores are clipped before fitting, and PEHE is evaluated on a held-out test set using the ACIC ground-truth treatment effects.

\FloatBarrier

\subsection{Additional synthetic results}
\label{sec:additional-synthetic-results}

\begin{figure}[htbp]
  \centering
  \begin{subfigure}[t]{0.48\linewidth}
    \centering
    \vspace{0pt}
    \includegraphics[width=\linewidth]{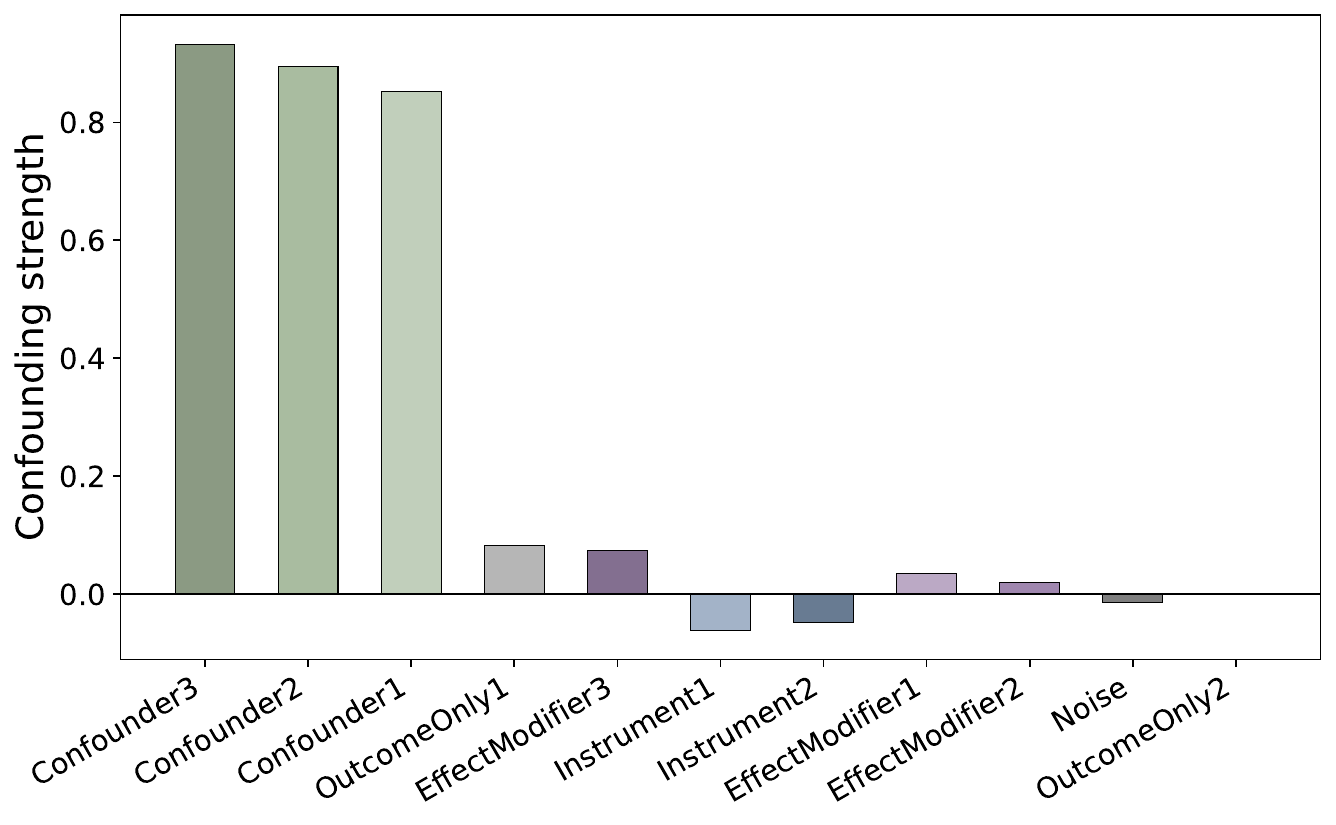}
    \caption{Exact computation}
    \label{fig:11cov_exact}
  \end{subfigure}
  \hspace{0.02\linewidth}
  \begin{subfigure}[t]{0.48\linewidth}
    \centering
    \vspace{0pt}
    \includegraphics[width=\linewidth]{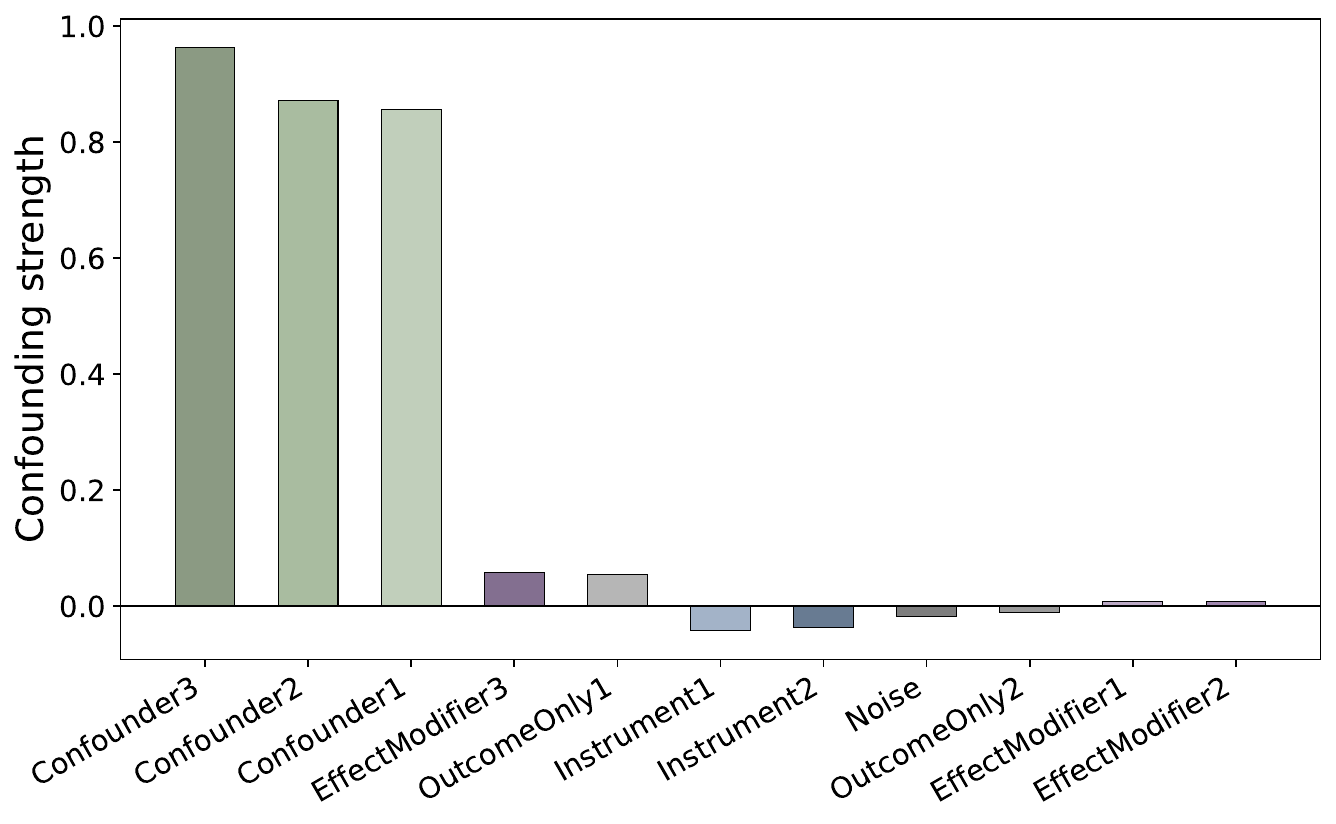}
    \caption{RegressionMSR approximation}
    \label{fig:11cov_proxy}
  \end{subfigure}
  \vspace{-0.2cm}
  \caption{Global Shapley values for synthetic datasets with 11 covariates:
  exact versus approximate computation.}
  \label{fig:11cov_exact_vs_approx}
\end{figure}

\begin{figure}[htbp]
  \centering
  \includegraphics[width=0.4\linewidth]{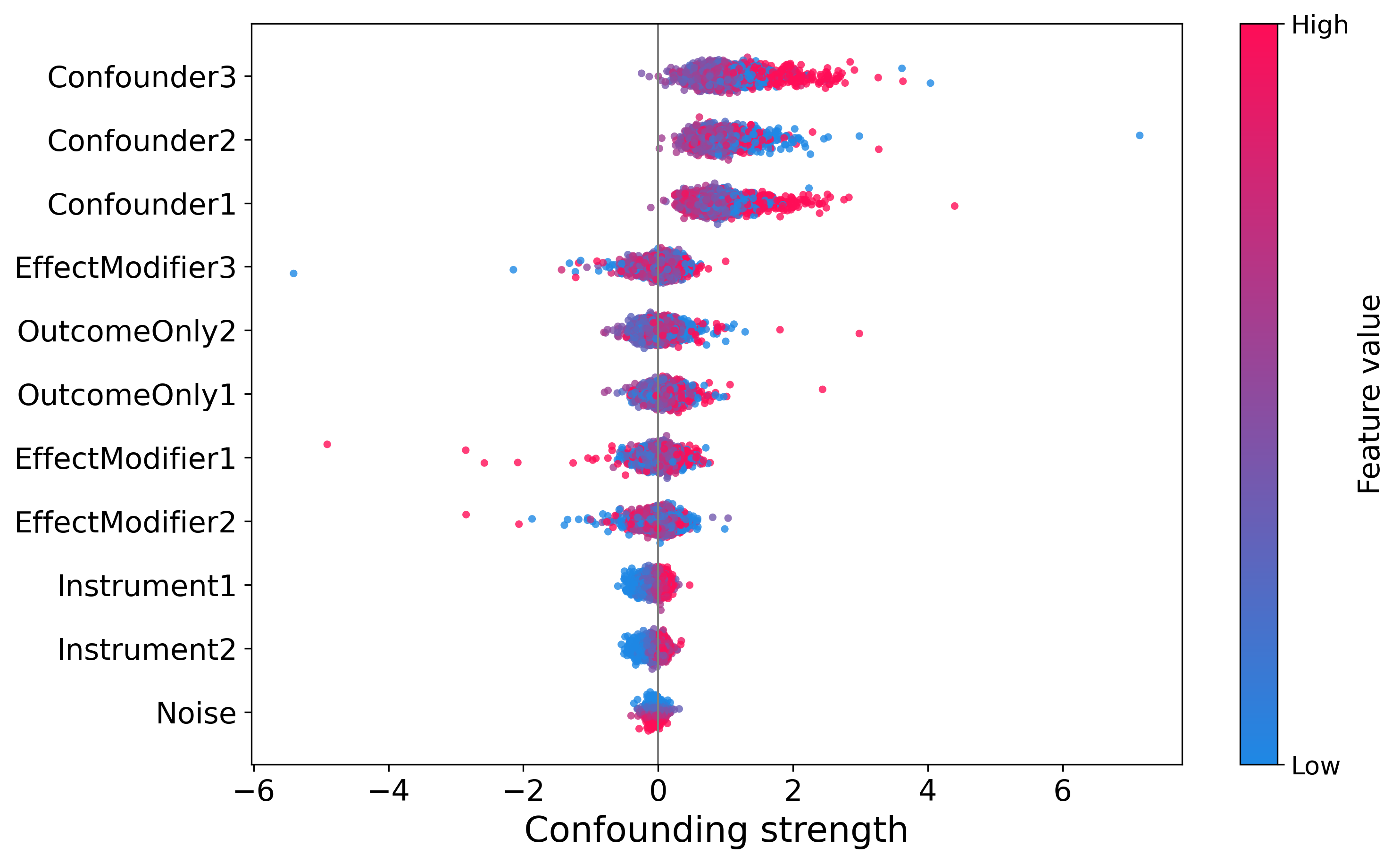}
  \caption{Local Shapley values for the 11-covariate synthetic dataset with
  exact computation.}
  \label{fig:synth_11cov_beeswarm}
\end{figure}

\begin{figure}[htbp]
  \centering
  \includegraphics[width=0.4\linewidth]{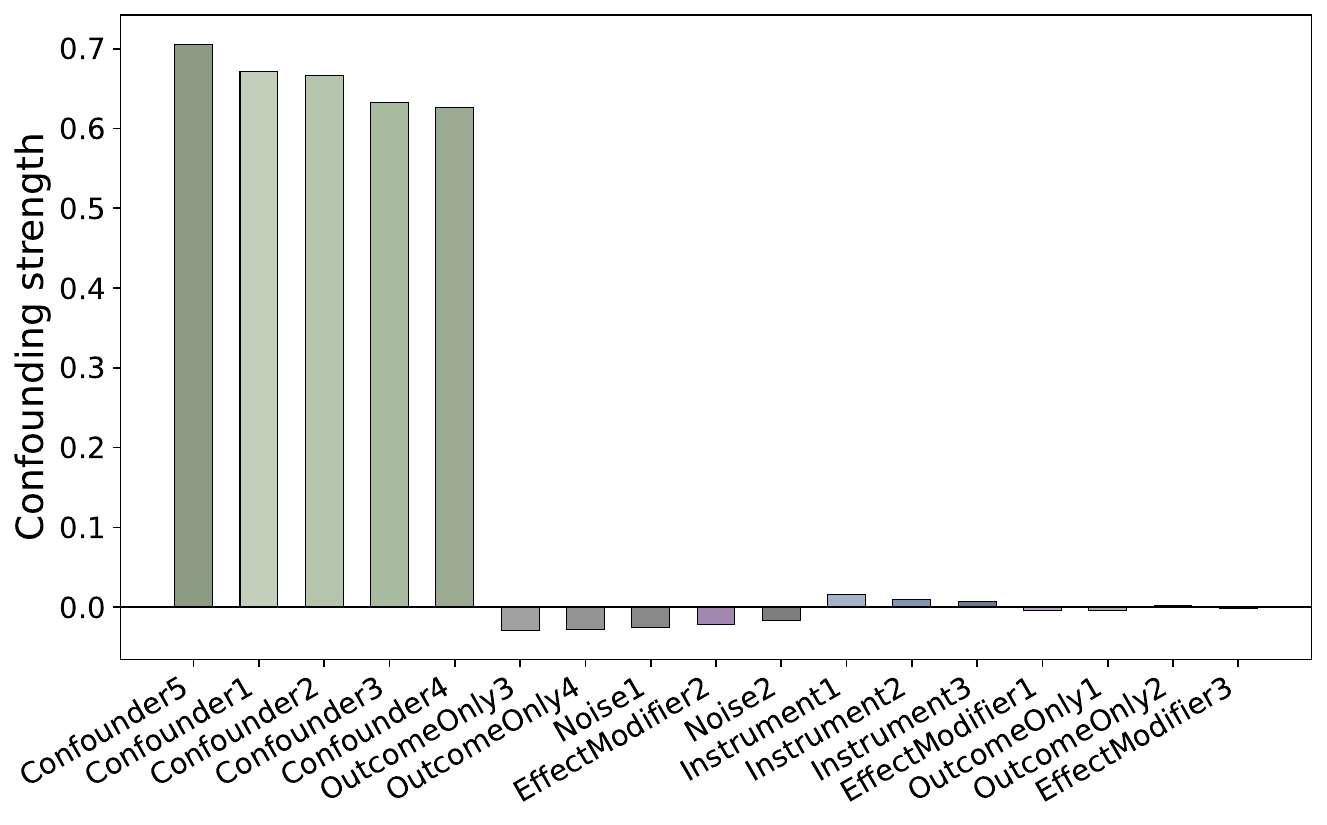}
  \caption{Global Shapley values for the 17-covariate synthetic dataset with
  RegressionMSR approximation.}
  \label{fig:synth_17cov_importancebar_proxy}
\end{figure}

We further compare exact and approximate computation in the 11-covariate synthetic setting. Fig.~\ref{fig:11cov_exact_vs_approx} shows that the budgeted approximation preserves the main qualitative ranking: true confounders receive the largest attributions, while non-confounding variables remain small. Fig.~\ref{fig:synth_11cov_beeswarm} reports the corresponding local explanations, showing that confounder contributions are visible at the individual level as well. Fig.~\ref{fig:synth_17cov_importancebar_proxy} provides an additional 17-covariate result, again showing that attribution is concentrated on the true confounders.

\vspace{0.5cm}

\vspace{0.5cm}

\vspace{0.5cm}

\subsection{Stability over approximation seeds}
\label{sec:stability-approximation}

\begin{figure}[htbp]
  \centering
  \begin{subfigure}[t]{0.48\linewidth}
    \centering
    \vspace{0pt}
    \includegraphics[width=\linewidth]{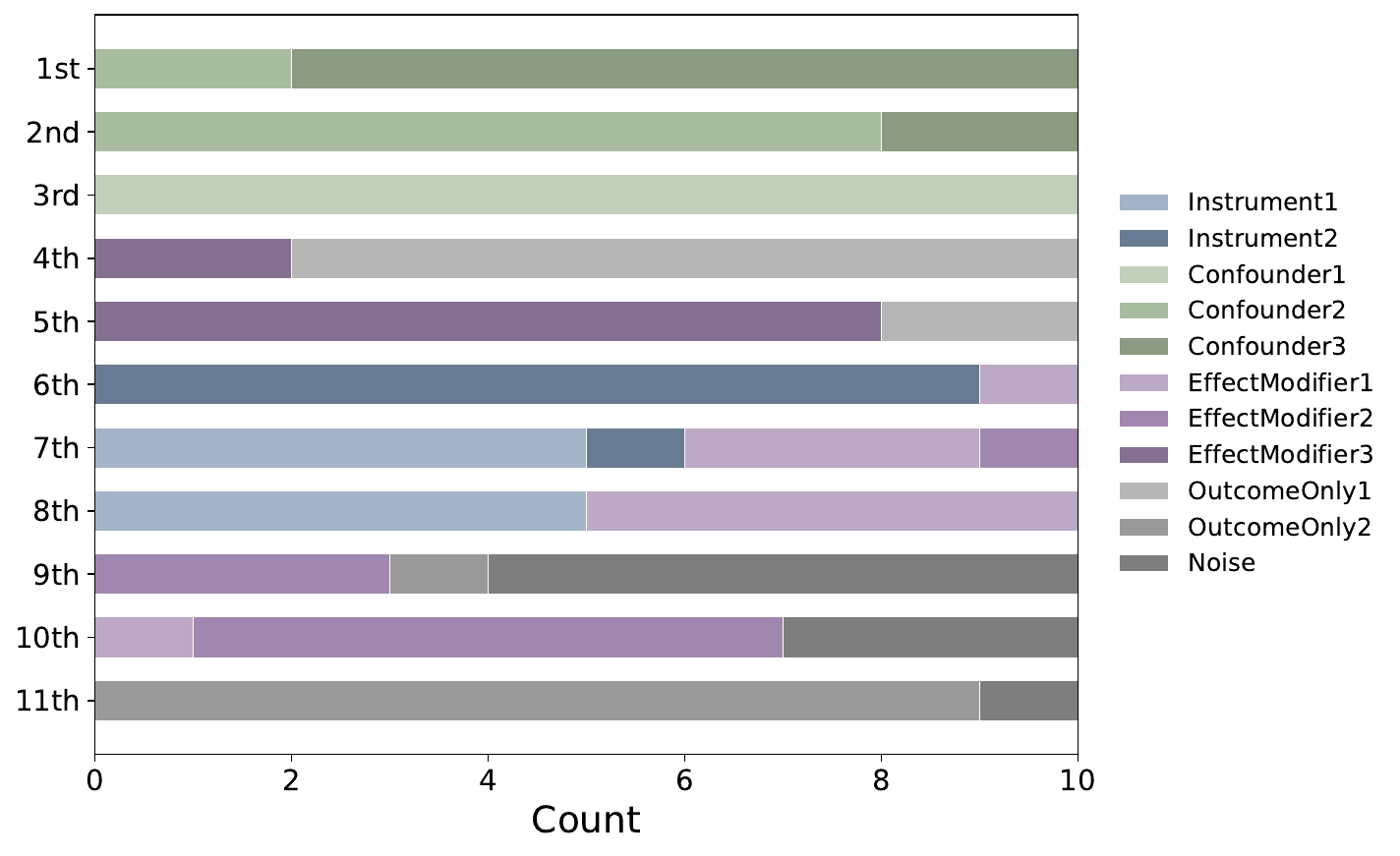}
    \caption{Rank stability}
    \label{fig:11cov_proxy_ranking}
  \end{subfigure}
  \hspace{0.02\linewidth}
  \begin{subfigure}[t]{0.48\linewidth}
    \centering
    \vspace{0pt}
    \includegraphics[width=\linewidth]{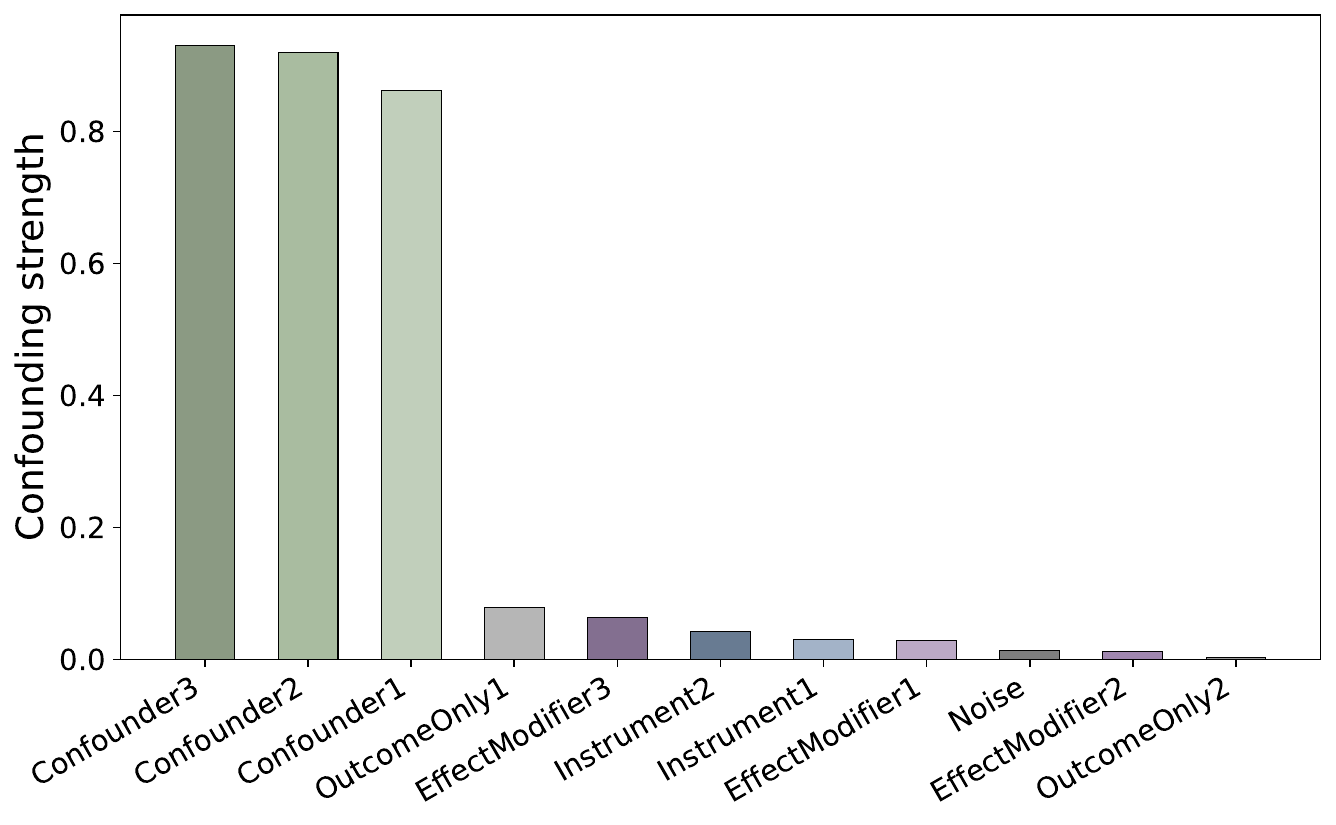}
    \caption{Average attribution}
    \label{fig:11cov_stability_proxy_importancebar}
  \end{subfigure}
  \vspace{-0.2cm}
  \caption{Stability of approximate \confoundingshap values over 10 runs in the
  11-covariate synthetic setting.}
  \label{fig:11cov_stability_proxy}
\end{figure}

We examine whether the approximate rankings are stable across repeated runs. The rank-stability plot in Fig.~\ref{fig:11cov_stability_proxy} records how often each covariate appears at each rank across 10 runs, while the bar plot reports the average absolute attribution. The true confounders are consistently among the highest ranks, indicating that the approximation does not merely assign importance to arbitrary correlated features.

\vspace{0.5cm}

\subsection{Stability over repeated synthetic datasets}
\label{sec:stability-dgp}

\begin{figure}[htbp]
  \centering
  \begin{subfigure}[t]{0.48\linewidth}
    \centering
    \vspace{0pt}
    \includegraphics[width=\linewidth]{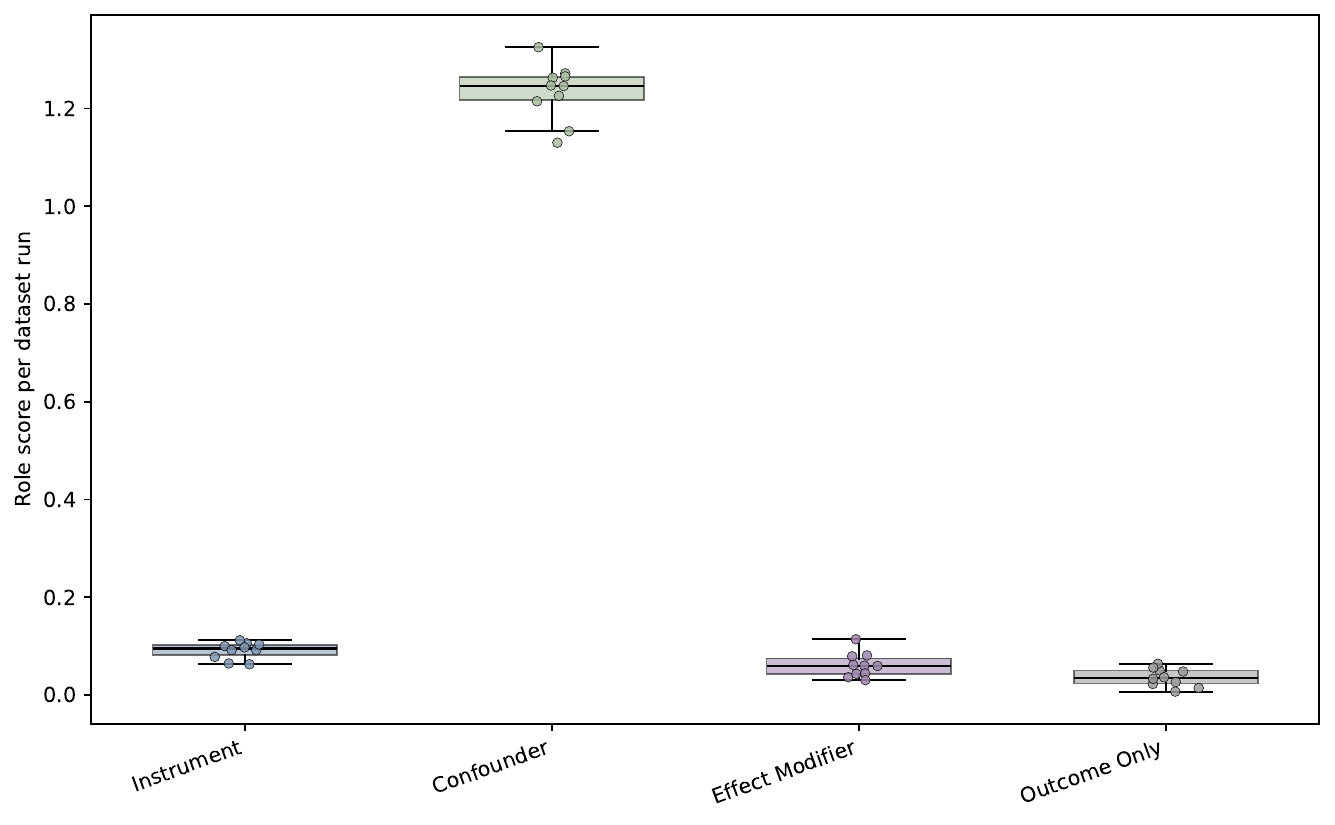}
    \caption{Role scores over datasets}
    \label{fig:4cov_datasetstability_exact_boxplot}
  \end{subfigure}
  \hspace{0.02\linewidth}
  \begin{subfigure}[t]{0.48\linewidth}
    \centering
    \vspace{0pt}
    \includegraphics[width=\linewidth]{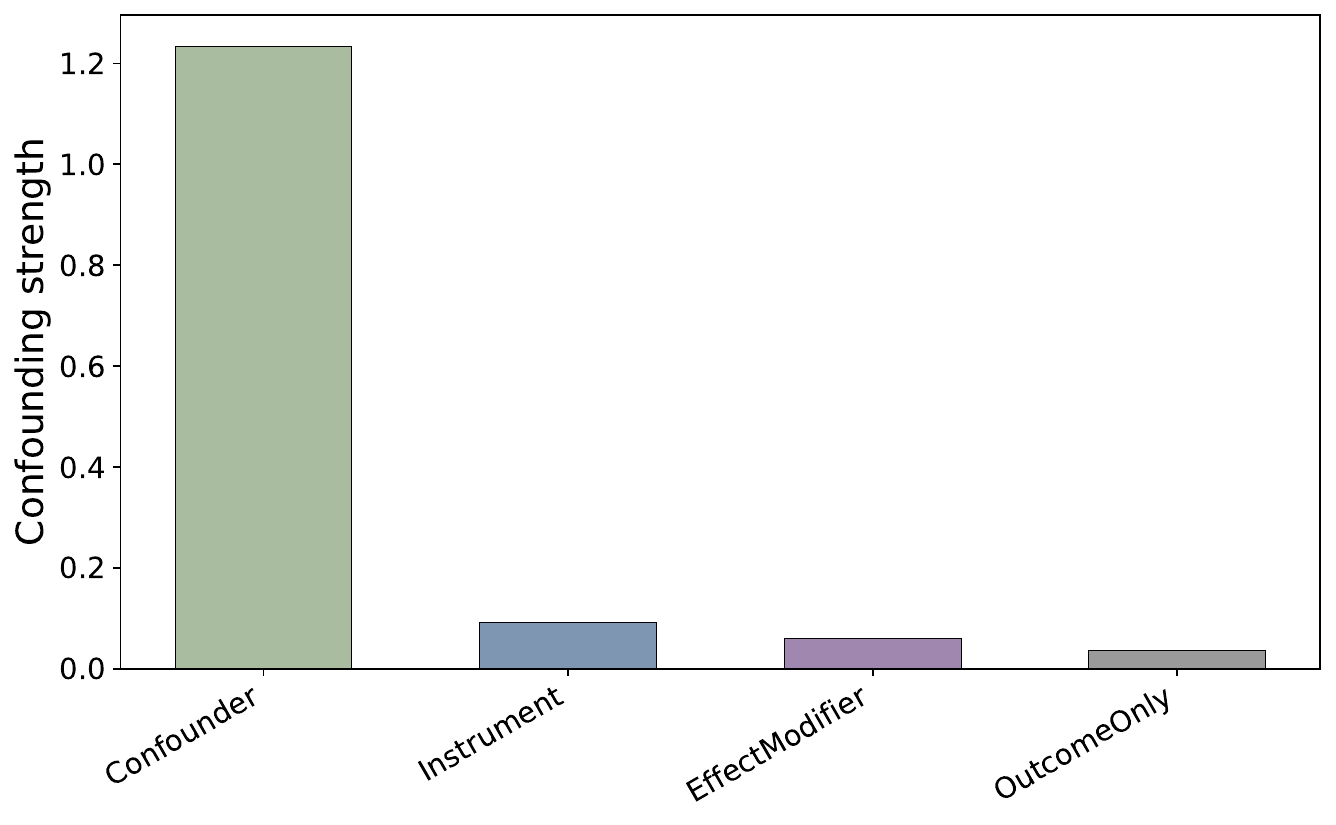}
    \caption{Average attribution}
    \label{fig:4cov_datasetstability_exact_importancebar}
  \end{subfigure}
  \vspace{-0.2cm}
  \caption{Dataset-level stability for the four-covariate synthetic setting with exact computation.}
  \label{fig:4cov_datasetstability_exact}
\end{figure}

\begin{figure}[htbp]
  \centering
  \begin{subfigure}[t]{0.48\linewidth}
    \centering
    \vspace{0pt}
    \includegraphics[width=\linewidth]{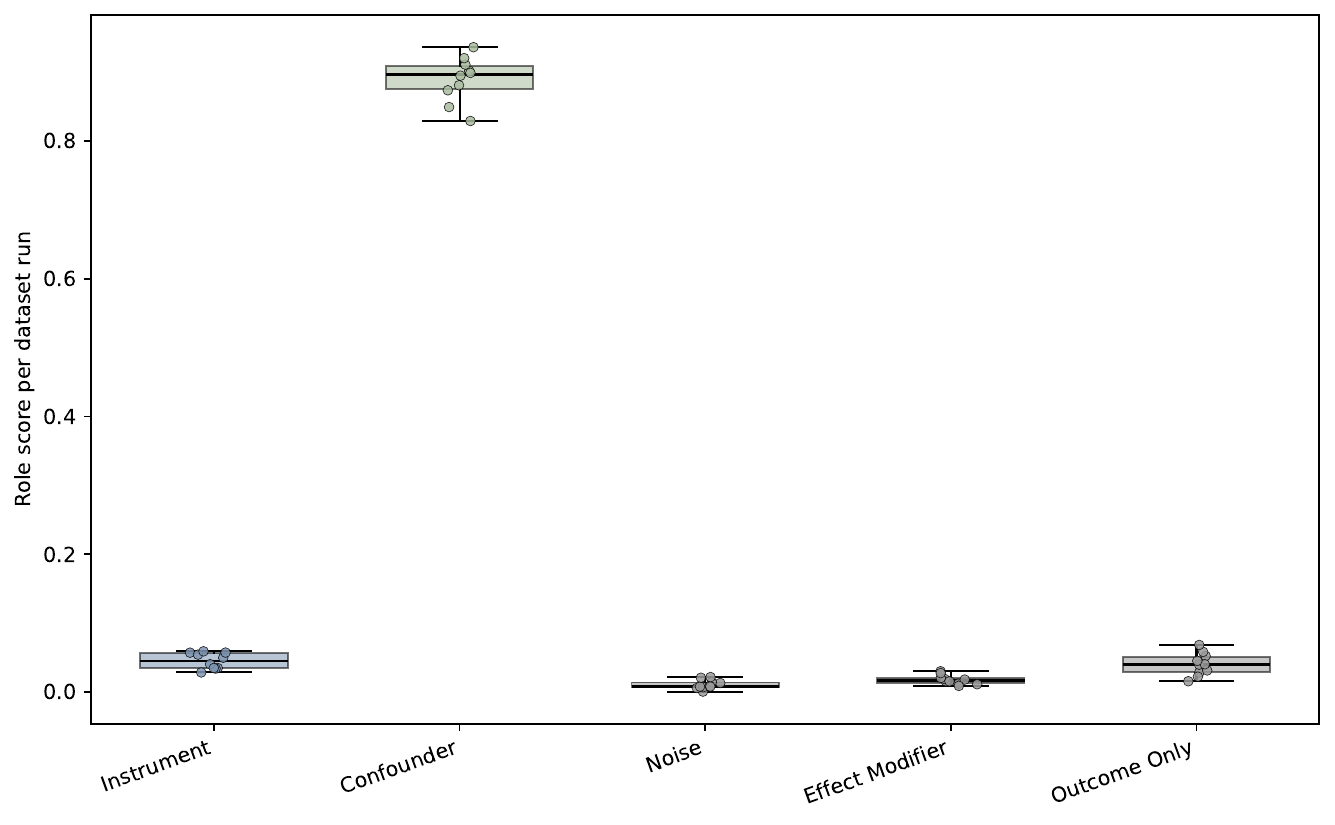}
    \caption{Role scores over datasets}
    \label{fig:11cov_datasetstability_exact_boxplot}
  \end{subfigure}
  \hspace{0.02\linewidth}
  \begin{subfigure}[t]{0.48\linewidth}
    \centering
    \vspace{0pt}
    \includegraphics[width=\linewidth]{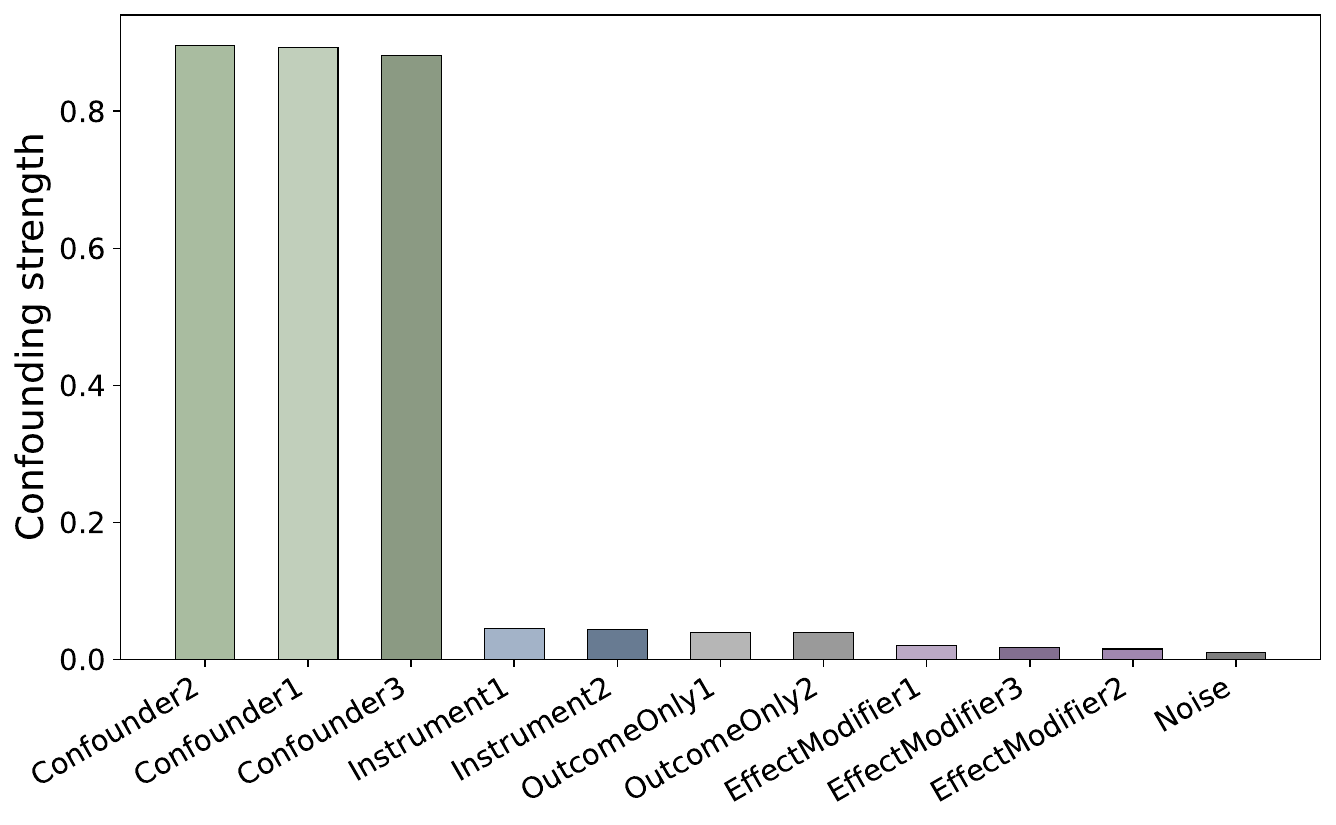}
    \caption{Average attribution}
    \label{fig:11cov_datasetstability_exact_importancebar}
  \end{subfigure}
  \vspace{-0.2cm}
  \caption{Dataset-level stability for the 11-covariate synthetic setting with exact computation.}
  \label{fig:11cov_datasetstability_exact}
\end{figure}

\begin{figure}[htbp]
  \centering
  \begin{subfigure}[t]{0.48\linewidth}
    \centering
    \vspace{0pt}
    \includegraphics[width=\linewidth]{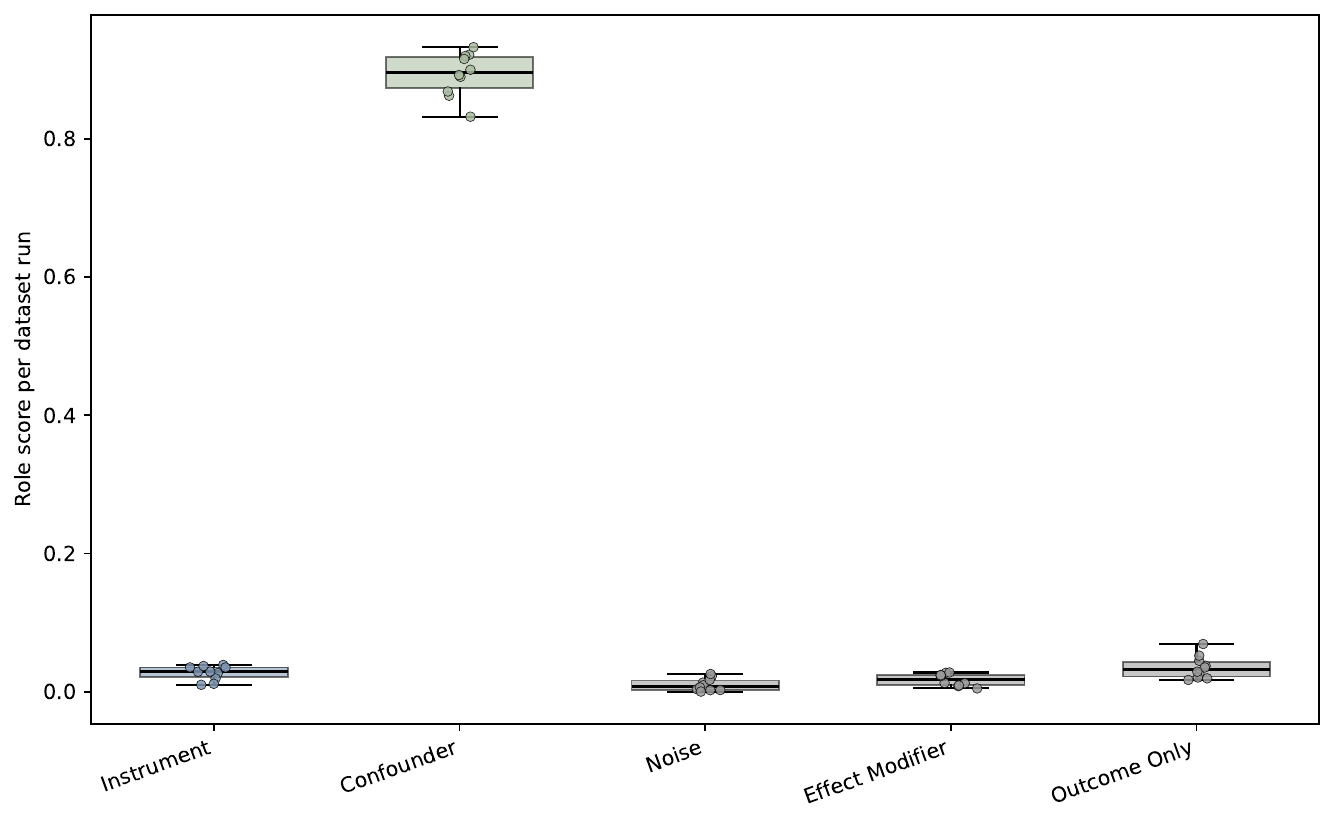}
    \caption{Role scores over datasets}
    \label{fig:11cov_datasetstability_proxy_boxplot}
  \end{subfigure}
  \hspace{0.02\linewidth}
  \begin{subfigure}[t]{0.48\linewidth}
    \centering
    \vspace{0pt}
    \includegraphics[width=\linewidth]{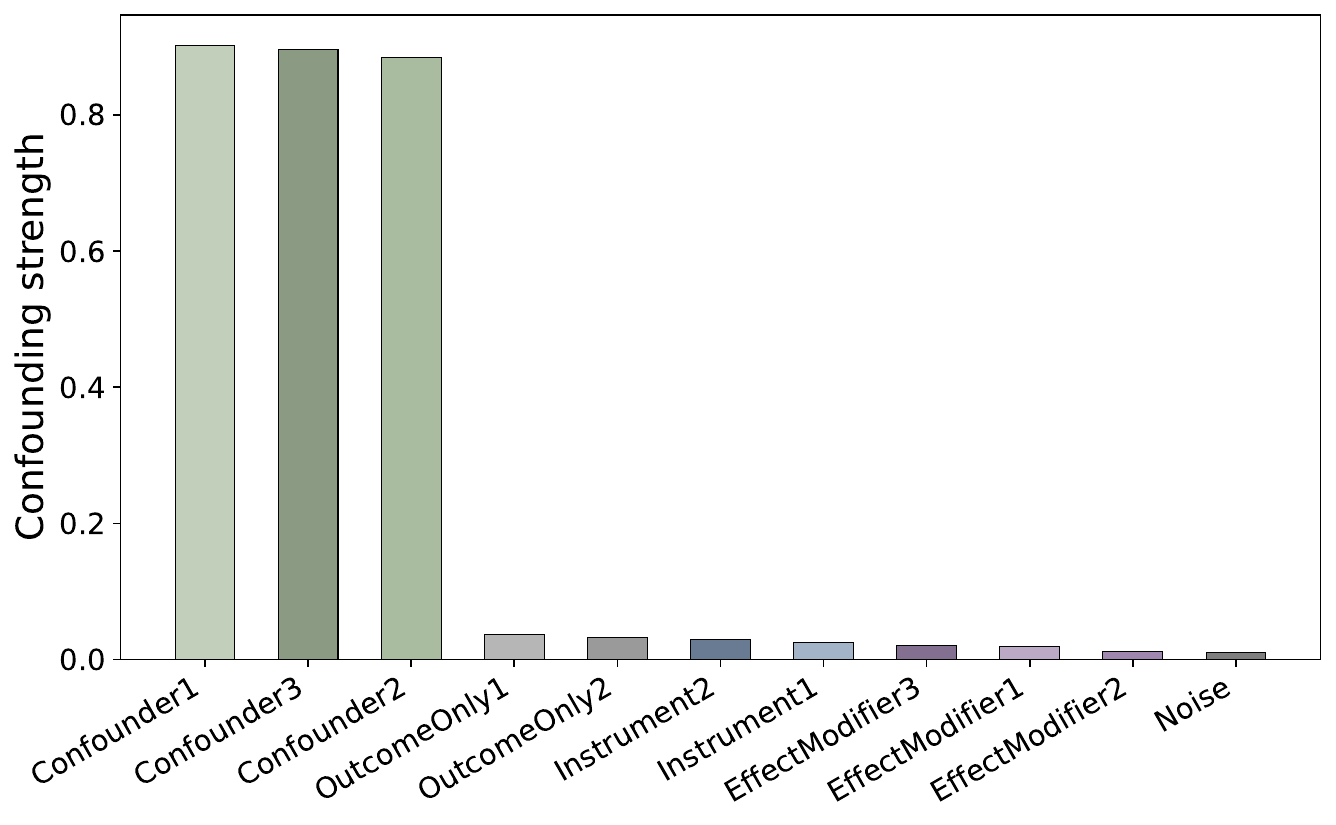}
    \caption{Average attribution}
    \label{fig:11cov_datasetstability_proxy_importancebar}
  \end{subfigure}
  \vspace{-0.2cm}
  \caption{Dataset-level stability for the 11-covariate synthetic setting with RegressionMSR approximation.}
  \label{fig:11cov_datasetstability_proxy}
\end{figure}

We additionally test stability across independently generated synthetic datasets. This evaluates whether the method recovers the intended covariate roles across different samples from the same data-generating process, rather than relying on a favorable random draw. Figures~\ref{fig:4cov_datasetstability_exact}, \ref{fig:11cov_datasetstability_exact}, and \ref{fig:11cov_datasetstability_proxy} show that true confounders remain the dominant contributors across repeated datasets for both exact and approximate computation.

\vspace{0.5cm}

\vspace{0.5cm}

\vspace{0.5cm}

\FloatBarrier

\clearpage
\section{Runtimes}
\label{sec:runtimes}

Table~\ref{tab:runtimes} reports wall-clock runtimes for the main experiments and ablations. The timings reflect end-to-end coalition evaluation and Shapley reconstruction under the reported budgets. They should be interpreted together with the budget ablations in Section~\ref{sec:ablation}: larger coalition budgets improve confounder recovery but can become expensive in high-dimensional settings.

\input{tab/runtimes}

\newpage
\section{Datasets}
\label{sec:datasets}

We evaluate \confoundingshap using synthetic, semi-synthetic, and real-world datasets. The synthetic and semi-synthetic experiments are designed so that the confounding structure is known by construction. The SUPPORT/RHC benchmark is a less controlled observational stress test in which the exact covariate-level confounding ground truth is unavailable.

\paragraph{Synthetic datasets.} Our controlled experiments use a synthetic data-generating process based on the simulation design of \citep{curth2021}. Covariates are drawn independently from a standard normal distribution and partitioned into interpretable roles: instruments, confounders, effect modifiers, outcome-only variables, and noise variables. We use several benchmark dimensions with known role composition. The 4-covariate setting contains one instrument, one confounder, one effect modifier, and one outcome-only variable. The 11-covariate setting contains two instruments, three confounders, three effect modifiers, two outcome-only variables, and one noise variable. The 17-covariate setting contains three instruments, five confounders, three effect modifiers, four outcome-only variables, and two noise variables. In the high-dimensional 100-covariate setting, we use 15 instruments, 40 confounders, 15 effect modifiers, 15 outcome-only variables, and 15 noise variables.

Let
\begin{equation}
X = (X_Z, X_C, X_\tau, X_O, X_N),
\qquad
X_j \sim \mathcal{N}(0,1),
\end{equation}
where $X_Z$ denotes instruments, $X_C$ confounders, $X_\tau$ effect modifiers, $X_O$ outcome-only variables, and $X_N$ noise variables. The untreated outcome surface follows the sum-of-squares structure in \citep{curth2021}, i.e.,
\begin{equation}
\mu_0(X) = \sum_{j \in C \cup O} X_j^2,
\end{equation}
and the treated outcome surface is
\begin{equation}
\mu_1(X) = \mu_0(X) + \sum_{j \in \tau} X_j^2.
\end{equation}
Thus, the conditional average treatment effect is
\begin{equation}
\tau(X) = \mu_1(X) - \mu_0(X) = \sum_{j \in \tau} X_j^2,
\end{equation}
so only the effect-modifier block changes the individual treatment effect.

Treatment assignment is generated from a logistic propensity score. We define the confounding score
\begin{equation}
m_C(X) = \frac{1}{|C|}\sum_{j \in C} X_j^2,
\qquad
\omega = Q_{0.5}\{m_C(X)\},
\end{equation}
where $\omega$ is the sample median of $m_C(X)$. To include variables that affect treatment but not outcome, we add an explicit instrument term
\begin{equation}
z(X) = \frac{1}{|Z|}\sum_{j \in Z} X_j.
\end{equation}
The propensity score is
\begin{equation}
e(X)
=
\mathbb{P}(A=1 \mid X)
=
\operatorname{logit}^{-1}
\left(\xi\{m_C(X)-\omega\}+\gamma_Z z(X)\right),
\end{equation}
with $\xi=3$ and $\gamma_Z=1$ in the main experiments. The treatment is sampled via
\begin{equation}
A \sim \operatorname{Bernoulli}(e(X)).
\end{equation}
Finally, with $\varepsilon \sim \mathcal{N}(0,\sigma^2)$, potential and factual
outcomes are generated as
\begin{equation}
Y(0)=\mu_0(X)+\varepsilon,\qquad
Y(1)=\mu_1(X)+\varepsilon,\qquad
Y = A Y(1) + (1-A)Y(0).
\end{equation}
This construction separates variables that predict treatment only, variables that truly confound treatment effect estimation, variables that modify treatment effects, variables that predict outcomes only, and irrelevant noise variables.

\paragraph{ACTG 175 semi-synthetic dataset.} We construct a semi-synthetic benchmark from ACTG 175, a randomized clinical trial comparing nucleoside monotherapy and combination therapy in HIV-infected adults with baseline CD4 counts between 200 and 500 cells per cubic millimeter \citep{hammer1996}. We restrict the data to the zidovudine-only control arm and the zidovudine-plus-didanosine treatment arm. The retained baseline covariates are age, weight, hemophilia status, history of intravenous drug use, Karnofsky score, race, gender, symptomatic status, prior antiretroviral experience, baseline CD4 count, and baseline CD8 count.

To obtain a known confounding structure while preserving realistic covariate distributions, we replace the randomized treatment assignment with a semi-synthetic mechanism. We designate clinically plausible variables as confounders, namely, baseline severity and treatment-history variables: age, Karnofsky score, symptomatic status, prior antiretroviral experience, and baseline CD4 count. These variables determine both the synthetic propensity score and the synthetic potential outcome model. The remaining variables are retained in the feature matrix but excluded from the assignment and outcome mechanisms. They therefore act as non-confounding covariates in the semi-synthetic benchmark. Continuous covariates are standardized before generating treatment and outcomes.

Let $X_C$ denote the designated confounder block and $X_N$ the remaining retained baseline covariates. The treatment assignment is generated by
\begin{equation}
e(X_C)
=
\mathbb{P}(A=1 \mid X_C)
=
\operatorname{logit}^{-1}(\alpha_0 + X_C^\top \alpha),
\qquad
A \sim \operatorname{Bernoulli}(e(X_C)).
\end{equation}
In the continuous outcome experiments, potential outcomes are generated by
\begin{equation}
\mu_0(X_C) = \beta_0 + X_C^\top \beta,
\qquad
\mu_1(X_C) = \mu_0(X_C) + \tau,
\end{equation}
and
\begin{equation}
Y(0)=\mu_0(X_C)+\varepsilon,\qquad
Y(1)=\mu_1(X_C)+\varepsilon,\qquad
\varepsilon \sim \mathcal{N}(0,\sigma^2),
\end{equation}
with factual outcome
\begin{equation}
Y = A Y(1) + (1-A)Y(0).
\end{equation}
The non-confounding ACTG covariates $X_N$ remain in the observed feature matrix but do not enter $e(X_C)$ or $\mu_a(X_C)$. This produces realistic empirical covariate correlations while keeping the true confounding roles known by construction.

In the reported ACTG 175 semi-synthetic continuous outcome experiment, we use the following parameter values:
\begin{equation}
\alpha_0=-0.2,\qquad
\alpha=(0.2,-0.8,0.9,0.5,-0.9),
\end{equation}
\begin{equation}
\beta_0=0,\qquad
\beta=(0.1,-0.5,0.5,0.25,-0.55),
\qquad
\tau=0.45,\qquad
\sigma=0.35.
\end{equation}
The coefficients are ordered according to the confounder block (age, Karnofsky score, symptomatic status, prior antiretroviral experience, baseline CD4 count). We use random seed 42 and retain all available observations after filtering to the two treatment arms and dropping missing values. Both the real data and the semi-synthetic data contain $n=1054$ observations.

\paragraph{ACIC 2016 benchmark.} For the ACIC experiment, we use the ACIC Challenge 2016 data distributed with the \texttt{causallib} repository \cite{shimoni2019} (\texttt{BiomedSciAI/causallib}, path \texttt{causallib/datasets/data/acic\_challenge\_2016}). We use \texttt{x.csv} for the covariates and the first simulation instance, \texttt{zymu\_1.csv}, for treatment assignments, factual outcomes, and potential-outcome means. The files are cached locally under \texttt{data/acic\_challenge\_2016}. Categorical variables are one-hot encoded and missing numeric values are median-imputed, yielding \(n=4802\) observations and 82 encoded covariates. The reported run uses random seed 42, a RegressionMSR coalition budget of 2048, PEHE test fraction 0.3, and feature-drop sizes \(\{5,10,15,20\}\).

\paragraph{SUPPORT/RHC observational dataset.}
The SUPPORT/RHC benchmark is derived from the observational study of right-heart catheterization in critically ill patients \citep{connors1996}. Here,  treatment corresponds to receiving a right-heart catheterization during the first 24 hours of ICU care, while the outcome is 30-day mortality. We use the following clinical variables as confounders: acute physiology score, mean blood pressure, and serum creatinine. Additional demographic, socioeconomic, and administrative variables are retained (with only small confounding strength). Unlike the synthetic and ACTG semi-synthetic benchmarks, SUPPORT/RHC does not provide exact feature-role ground-truth. We therefore use it as a qualitative observational benchmark to assess whether \confoundingshap assigns low attribution to background descriptors and higher attribution to clinically plausible adjustment variables.

\newpage
\section{Computational complexity}
\label{sec:complexity}

We measure computational cost in terms of coalition-value evaluations. A coalition-value evaluation estimates the nuisance quantities needed to compute
$\hat{\nu}(S)$ for one coalition $S$. With TabPFN, this is implemented through a constant number of in-context conditioning steps on a frozen pretrained model, rather than through gradient-based refitting. We therefore omit constant factors and focus on the number of queried coalitions.

\textbf{Exact computation.}
Exact Shapley computation evaluates every coalition, i.e.,
\begin{equation}
C_{\mathrm{exact}}
=
1 + 2^p
=
\mathcal{O}(2^p),
\label{eq:complexity-exact}
\end{equation}
where the additive constant corresponds to the full-covariate outcome model used to form pseudo-outcomes.

\textbf{Approximate computation.}
With coalition budget $B$, only $B$ coalitions are queried, i.e.,
\begin{equation}
C_{\mathrm{approx}}
=
1 + B
=
\mathcal{O}(B).
\label{eq:complexity-approx}
\end{equation}
This expression counts coalition evaluations; the actual wall-clock time also depends on sample size, covariate dimension, hardware, and the chosen Shapley reconstruction method. Empirical runtimes in Section~\ref{sec:runtimes} show the practical cost of this approximation for the datasets considered in our
experiments.

\section{Limitations}
\label{sec:limitations}

\confoundingshap attributes observed confounding bias relative to the available full covariate set. The interpretation is therefore conditional on the measured variables: if important confounders are unobserved, the attributions should be understood relative to the observed reference set rather than as ground-truth causal bias. As with other plug-in causal estimation procedures, attribution quality can also be affected by limited overlap, finite-sample variability, and nuisance-estimation error. In this sense, \confoundingshap is complementary to sensitivity analysis, causal diagrams, domain knowledge, and robustness checks.

A second limitation is computational. Although TabPFN-based coalition evaluation avoids gradient-based refitting, reliable approximation in high-dimensional settings still requires sufficiently large coalition budgets. The 200-covariate ablations show that attribution quality improves with increasing budget, but such runs can be computationally expensive in practice.

\section{Broader impacts}
\label{sec:broader-impacts}

\confoundingshap is intended to improve transparency in causal machine learning by showing which measured covariates contribute most to observed confounding. This is particularly relevant in clinical and other observational studies, where treatment groups often differ systematically in baseline characteristics and where transparent adjustment decisions are important. By identifying measured variables that drive treatment--control incomparability, \confoundingshap may support the construction of minimal or parsimonious adjustment sets, improve communication between methodological and domain experts, and inform the design of future observational studies or trials by highlighting variables that should be measured, stratified on, or checked for balance. The main risk is overinterpretation \citep{martens2025}: \confoundingshap values should not be read as automatically certifying that an adjustment set is sufficient or that a causal estimate is unbiased. Rather, the values provide diagnostic evidence about measured confounding and should be interpreted together with domain knowledge, causal reasoning, sensitivity analyses, and robustness checks.

\end{document}

%% file: tab/runtimes.tex
\begin{table}[htbp]
  \centering
  \small
  \begin{tabular}{llrrrll}
    \hline
    Dataset & Method & Covariates & Budget & Samples & Runtime in s & Runtime in h:m:s \\
    \hline
    Synthetic & Exact & 4 & -- & 5000 & 179.34 & 0h 2m 59.34s \\
    Synthetic & Exact & 11 & -- & 5000 & 27004.25 & 7h 30m 4.25s \\
    Synthetic & RegressionMSR$^\dagger$ & 11 & 128 & 5000 & 1631.46 & 0h 27m 11.46s \\
    Synthetic & RegressionMSR & 17 & 128 & 5000 & 2430.24 & 0h 40m 30.24s \\
    \hline
    ACTG RCT & Exact & 11 & -- & 1054 & 28171.89 & 7h 49m 31.89s \\
    ACTG semisynth & Exact & 11 & -- & 1054 & 20646.45 & 5h 44m 6.45s \\
    SUPPORT & RegressionMSR & 10 & 256 & 5735 & 5883.33 & 1h 38m 3.33s \\
    \hline
    \multicolumn{7}{p{.9\textwidth}}{\footnotesize $^\dagger$ MSR and KernelSHAP reused the precomputed coalition values from RegressionMSR and are therefore not reported separately.}
  \end{tabular}
  \caption{Reported runtimes}
  \label{tab:runtimes}
\end{table}